\pdfoutput=1

\documentclass[11pt]{article}

\usepackage{CJKutf8}

\usepackage[dvipsnames]{xcolor}

\usepackage{ACL2023}

\usepackage{times}
\usepackage{latexsym}

\usepackage[T1]{fontenc}

\usepackage[utf8]{inputenc}

\usepackage{microtype}

\usepackage{inconsolata}

\usepackage{algorithm}
\usepackage{algpseudocode}
\usepackage{amsmath}
\usepackage{amssymb}
\usepackage{graphicx}
\usepackage{multirow}
\usepackage{pifont}
\usepackage{rotating}
\usepackage{scalefnt}
\usepackage{tablefootnote}

\usepackage{CJKutf8}

\newcommand{\cmark}{\ding{51}}
\newcommand{\xmark}{\ding{55}}

\title{$M^3$ Scaling Law: Optimizing Multi-Epoch, Multi-Lingual, and\\Multi-Stage Training for Low-Resource Language Models}

\author{Kosuke Akimoto \and Taiki Miyagawa \and Masafumi Oyamada \\
NEC Corporation\\
\texttt{\{kosuke\_a, miyagawataik, oyamada\}@nec.com}
}

\begin{document}
\maketitle

\begin{abstract}
In this paper, we study a fundamental design problem in pretraining Large Language Models (LLMs) for low-resource language regimes.
Existing works adopt multi-epoch, multi-lingual, and multi-stage training to utilize the limited target-language corpus efficiently, but no prior scaling law can compare recipes spanning these approaches under the same compute budget $C$ and target-language corpus size $D_T$, leaving the optimal training setup unclear.
To address this gap, we propose the $M^3$ Scaling Law, a unified predictive model parameterized by the model scale, the number of target-corpus epochs $k$, the average target-language ratio $r$, and the final-stage target-language ratio $r_f$, which places monolingual single-stage, multi-lingual single-stage, and multi-lingual multi-stage recipes on a single target-language loss surface.
Across three language pairs, it extrapolates to unseen hyperparameter regions more accurately than existing scaling laws.
Using $M^3$ as a surrogate objective, we derive two practical guidelines for low-resource LLM pretraining: (i) as $D_T$ decreases, the optimal recipe shifts directly from monolingual single-stage to multi-lingual two-stage training at a compute-budget-dependent threshold, with multi-lingual single-stage never optimal in our experimental grid; and (ii) the optimal number of epochs collapses onto a single curve in the scarcity variable $D_T/D^*(C)$, where $D^*(C) \propto C^{\alpha/(\alpha+\beta)}$ is the monolingual compute-optimal corpus size.
\end{abstract}

\section{Introduction}

\begin{figure}[t]
  \includegraphics[width=\columnwidth]{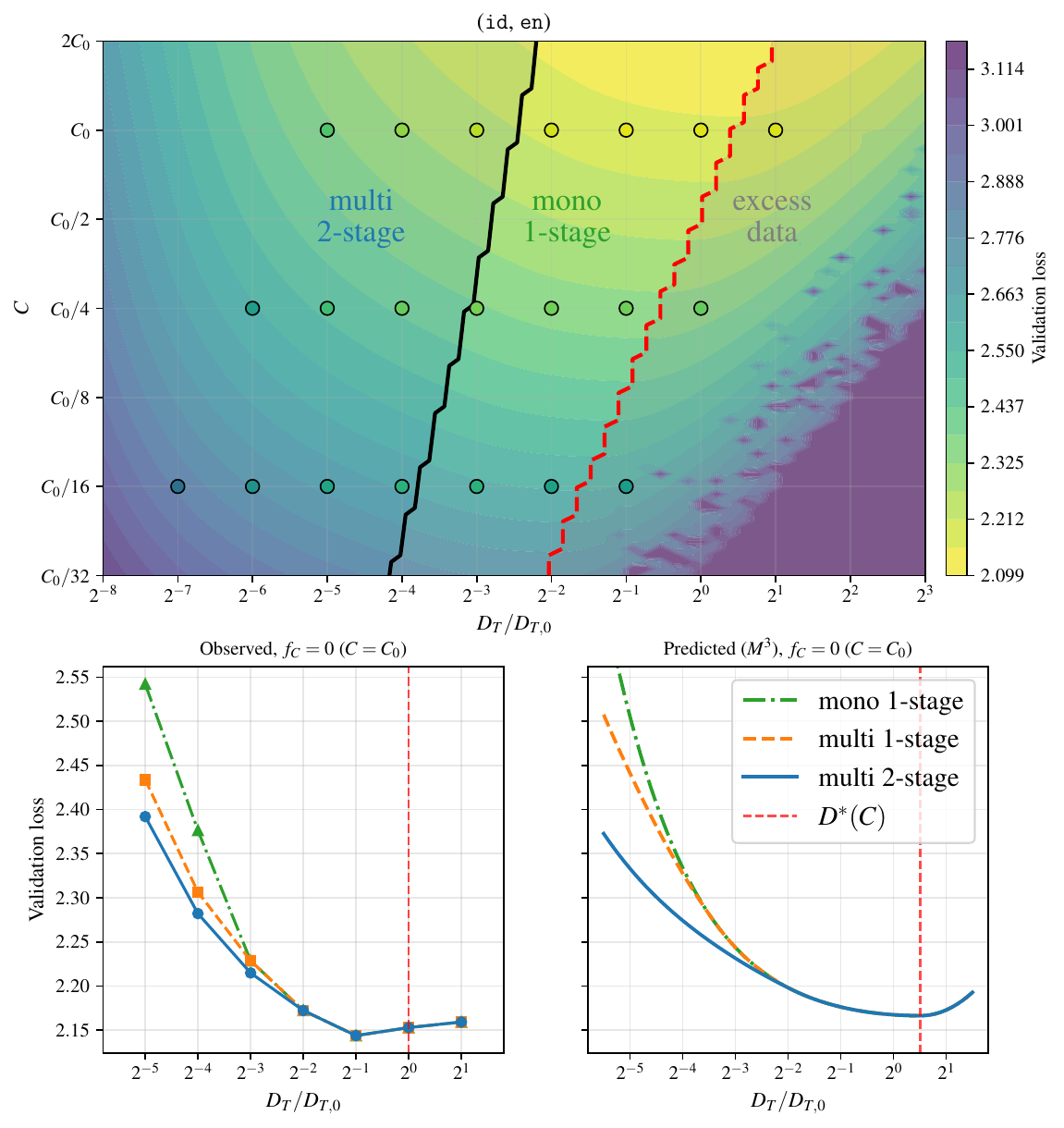}
  \caption{\textbf{Top:} Predicted minimum validation loss landscape over $(D_T, C)$ for the (Indonesian, English) pair, with three regime regions separated by the compute-optimal data size $D^*(C)$ (red dashed) and the approach boundary (black solid) at which the optimal training approach switches from mono 1-stage to multi 2-stage as $D_T$ decreases. Circles show observed best loss at each experimental $(D_T, C)$ point, colored by the same scale as the contours.
  \textbf{Bottom left:} Observed minimum validation loss for each training approach at $C = C_0$.
  \textbf{Bottom right:} Corresponding predictions by the $M^3$ Scaling Law.}
  \label{fig:main_total_graph}
\end{figure}

\begin{table*}[t]
  \centering
  \scalebox{0.68}{
    \begin{tabular}{c||cccccc|cc}
      & \begin{tabular}{c}model \\ scale \end{tabular} & \#epochs & \begin{tabular}{c} 1st \\ \#token \end{tabular} & \begin{tabular}{c} 1st lang. \\ ratio \end{tabular} & \begin{tabular}{c} 2nd \\ \#token \end{tabular} & \begin{tabular}{c} 2nd lang. \\ ratio \end{tabular} & \begin{tabular}{c} single- \\ stage \end{tabular} & \begin{tabular}{c} two- \\ stage \end{tabular} \\
      \hline\hline

      \begin{tabular}{c}
        \cite{kaplan2020scaling} \\ \cite{hoffmann2022training}
      \end{tabular} & \textcolor{ForestGreen}{$\bigstar$} & \textcolor{red}{\xmark} & \textcolor{ForestGreen}{$\bigstar$} & \textcolor{red}{\xmark} & -- & -- & \textcolor{ForestGreen}{$\bigstar$} & \textcolor{red}{\xmark} \\

      \cite{muennighoff2024scaling} & \textcolor{ForestGreen}{$\bigstar$} & \textcolor{ForestGreen}{$\bigstar$}$^*$ & \textcolor{ForestGreen}{$\bigstar$} & \textcolor{ForestGreen}{\cmark}$^*$ & -- & -- & \textcolor{ForestGreen}{$\bigstar$} & \textcolor{red}{\xmark} \\

      \cite{chang2023multilinguality} & \textcolor{ForestGreen}{\cmark} & \textcolor{red}{\xmark}$^\diamond$ & \textcolor{ForestGreen}{\cmark} & \textcolor{ForestGreen}{\cmark} & -- & -- & \textcolor{ForestGreen}{\cmark} & \textcolor{red}{\xmark} \\

      \cite{faysse2024croissantllm} & \textcolor{ForestGreen}{$\bigstar$} & \textcolor{red}{\xmark} & \textcolor{red}{\xmark} & \textcolor{ForestGreen}{$\bigstar$} & -- & -- & \textcolor{ForestGreen}{$\bigstar$} & \textcolor{red}{\xmark} \\

      \begin{tabular}{c}
        \cite{he2024scaling} \\ \cite{shukor2026scaling}$^\S$
      \end{tabular} & \textcolor{ForestGreen}{$\bigstar$} & \textcolor{red}{\xmark} & \textcolor{ForestGreen}{$\bigstar$} & \textcolor{ForestGreen}{$\bigstar$} & -- & -- & \textcolor{ForestGreen}{$\bigstar$} & \textcolor{red}{\xmark} \\

      \cite{longpre2025atlas} & \textcolor{ForestGreen}{$\bigstar$} & \textcolor{ForestGreen}{$\bigstar$} & \textcolor{ForestGreen}{$\bigstar$} & \textcolor{ForestGreen}{$\bigstar$} & -- & -- & \textcolor{ForestGreen}{$\bigstar$} & \textcolor{red}{\xmark}$^\dagger$ \\

      \cite{sedova2026scalinglawsmixturepretraining} & \textcolor{ForestGreen}{$\bigstar$} & \textcolor{ForestGreen}{$\bigstar$} & \textcolor{ForestGreen}{$\bigstar$} & \textcolor{ForestGreen}{$\bigstar$} & -- & -- & \textcolor{ForestGreen}{$\bigstar$} & \textcolor{red}{\xmark} \\

      \cite{ibrahim2024simple} & \textcolor{ForestGreen}{\cmark} & \textcolor{red}{\xmark} & -- & -- & \textcolor{ForestGreen}{\cmark} & \textcolor{ForestGreen}{\cmark} & \textcolor{red}{\xmark}$^\bigtriangleup$ & \textcolor{ForestGreen}{\cmark} \\

      \cite{que2024d}$^\S$ & \textcolor{ForestGreen}{$\bigstar$} & \textcolor{red}{\xmark} & -- & -- & \textcolor{ForestGreen}{$\bigstar$} & \textcolor{ForestGreen}{$\bigstar$} & \textcolor{red}{\xmark} & \textcolor{ForestGreen}{$\bigstar$} \\

      \cite{zhang2024scaling}$^\S$ & \textcolor{ForestGreen}{$\bigstar$}$^\|$ & \textcolor{red}{\xmark} & \textcolor{ForestGreen}{$\bigstar$}$^\|$ & -- & \textcolor{ForestGreen}{$\bigstar$} & \textcolor{red}{\xmark} & \textcolor{red}{\xmark} & \textcolor{ForestGreen}{$\bigstar$} \\

      \begin{tabular}{c}
        \cite{liew2026reusingovertrainedlanguagemodels}$^\S$ \\ \cite{seto2026optimal}$^\S$
      \end{tabular} & \textcolor{ForestGreen}{$\bigstar$} & \textcolor{red}{\xmark} & \textcolor{ForestGreen}{$\bigstar$} & -- & \textcolor{ForestGreen}{$\bigstar$} & \textcolor{red}{\xmark} & \textcolor{red}{\xmark} & \textcolor{ForestGreen}{$\bigstar$} \\

      \cite{goffinet2025ptpp} & \textcolor{ForestGreen}{$\bigstar$} & \textcolor{red}{\xmark} & \textcolor{ForestGreen}{$\bigstar$} & -- & \textcolor{ForestGreen}{$\bigstar$} & \textcolor{ForestGreen}{$\bigstar$} & \textcolor{red}{\xmark} & \textcolor{ForestGreen}{$\bigstar$} \\

      \cite{baek2026finetuner}$^\S$ & \textcolor{ForestGreen}{\cmark} & \textcolor{ForestGreen}{\cmark} & \textcolor{ForestGreen}{$\bigstar$} & \textcolor{ForestGreen}{$\bigstar$} & \textcolor{red}{\xmark}$^\sharp$ & \textcolor{red}{\xmark} & \textcolor{ForestGreen}{$\bigstar$} & \textcolor{ForestGreen}{$\bigstar$} \\

      \cite{wang2025learning}$^\S$ & \textcolor{ForestGreen}{$\bigstar$}$^\|$ & \textcolor{red}{\xmark} & \textcolor{ForestGreen}{$\bigstar$} & \textcolor{red}{\xmark} & \textcolor{ForestGreen}{$\bigstar$} & \textcolor{ForestGreen}{$\bigstar$}$^\|$ & \textcolor{ForestGreen}{$\bigstar$} & \textcolor{ForestGreen}{$\bigstar$} \\

      Ours & \textcolor{ForestGreen}{$\bigstar$} & \textcolor{ForestGreen}{$\bigstar$} & \textcolor{ForestGreen}{$\bigstar$}$^\ddagger$ & \textcolor{ForestGreen}{$\bigstar$}$^\ddagger$ & \textcolor{ForestGreen}{$\bigstar$}$^\ddagger$ & \textcolor{ForestGreen}{$\bigstar$} & \textcolor{ForestGreen}{$\bigstar$} & \textcolor{ForestGreen}{$\bigstar$} \\

      \hline
    \end{tabular}
  }

  \footnotesize{\textcolor{ForestGreen}{$\bigstar$}: covered by a proposed scaling law. $^\S$Focusing on adaptation to non-language domains. $^*$Epochs and language ratio impacts studied independently. $^\diamond$Only one number of epochs per corpus budget tested. $^\bigtriangleup$Only tested one baseline single-stage setup mixing all English and German corpus. $^\dagger$Only monolingual single-epoch continual pretraining from a fixed multi-lingual base model; second-stage hyperparameters are not explicitly explored. $^\sharp$Finetuning length is implicitly determined by early stopping rather than explicitly analyzed. $^\|$Only one of these parameters is covered at a time. $^\ddagger$These variables are not directly included in the proposed $M^3$ Scaling Law, but are reflected through the average language ratio and total number of tokens.}
  \caption{Scope of hyperparameters explored in existing works and our study.}
  \label{table:scope_of_hyperparameters}
\end{table*}

In this paper, we study a fundamental design problem in pretraining Large Language Models (LLMs) for low-resource language regimes.
Given a fixed compute budget $C$ and a fixed amount of available target-language corpus $D_T$, the practitioner faces a training-recipe design space whose axes include the model scale together with the hyperparameters of three well-known approaches: \textbf{multi-epoch training} \cite{muennighoff2024scaling} repeats the target corpus $k$ times; \textbf{multi-lingual training} \cite{le2023bloom} mixes high-resource data with target-language ratio $r$; and \textbf{multi-stage training} \cite{ibrahim2024simple} varies $r$ across stages.

Existing scaling laws cannot rank competing recipes under the same $(C, D_T)$.
As shown in Table~\ref{table:scope_of_hyperparameters}, prior laws cover only restricted slices of this design space:
single-stage laws \cite{hoffmann2022training,longpre2025atlas,sedova2026scalinglawsmixturepretraining} do not extend to multi-stage training, continual pretraining laws \cite{que2024d,goffinet2025ptpp,zhang2024scaling} lack a common objective with single-stage recipes, and even laws nominally spanning both settings \cite{baek2026finetuner,wang2025learning} leave orthogonal axes uncovered such as second-stage length or multi-epoch training.
As a result, no current law can compare, under the same $(C, D_T)$, recipes that differ along multiple axes at once, such as whether one should repeat the target corpus, mix high-resource data throughout training, or use high-resource data early and concentrate target-language data in the final stage.

We address this gap with the \textbf{$M^3$ Scaling Law}, a unified predictive model for target-language validation loss parameterized by $(M, D_T, k, r, r_f)$, where $r$ and $r_f$ are the average and final-stage target-language ratios.
As we show empirically, the latter two together summarize the effect of multi-stage schedules.
The $M^3$ Scaling Law maps different training recipes into a single loss-prediction surface through effective model size $M'$, effective data size $D'$, and a dual power-law factor in $(r, r_f)$.
This makes monolingual single-stage, multi-lingual single-stage, and multi-lingual multi-stage recipes directly comparable under fixed $(C, D_T)$.

We contribute in two layers.
First, as empirical modeling components:
(i) in multi-stage training, the target-language loss follows a dual power law in $(r, r_f)$, largely insensitive to intermediate-stage ratios, extending \citet{he2024scaling};
(ii) in monolingual multi-epoch training, the effective model size saturates more strongly with $k$, captured by a $k$-dependent saturation parameter $R_M^*(k)$ that recovers Chinchilla at $k{=}1$;
(iii) in multi-lingual multi-epoch training, the $(k, r)$ interaction is modeled by a language-ratio-dependent weight on high-resource tokens, extending \citet{longpre2025atlas}.

Second, using the $M^3$ Scaling Law as a surrogate objective, we then derive two recipe-selection rules:
(a) as $D_T$ decreases, the optimal approach transitions directly from monolingual single-stage to multi-lingual two-stage at a compute-budget-dependent threshold, with multi-lingual single-stage never optimal in our experimental grid;
(b) the optimal number of epochs collapses onto a single curve in $D_T/D^*(C)$, where $D^*(C)$ is the monolingual compute-optimal corpus size.

\section{Experimental Setups}

\subsection{Problem Setup}
\label{subsec:problem_setup}

In this paper, we focus on pretraining LLMs from scratch for a low-resource target language under a fixed compute budget $C$ and a fixed amount of available target-language corpus of size $D_T$.
In addition to the target-language corpus, we assume the availability of a high-resource language corpus.

To investigate the optimal training setup for each budget constraint $(C, D_T)$, we pretrained LLMs with various training setups for each $(C, D_T)$, spanning multiple training approaches, namely, multi-epoch, multi-lingual, and multi-stage training.
We adopt target-language validation loss on a held-out corpus as our primary metric, following common practice in scaling law research \cite[e.g.,][]{hoffmann2022training} and prior findings that validation loss correlates with downstream task performance \cite[e.g.,][]{chen2024scaling}; see Appendix~\ref{section:downstream_task_performance} for a preliminary correlation analysis on Japanese benchmarks.\footnote{Due to compute constraints, we ran a single training run per setup.}

In our experiments, we cover combinations of the following three well-known approaches for LLM pretraining for low-resource language:
\textbf{Multi-epoch training} \cite{muennighoff2024scaling} repeats the target-language corpus $k$ times during training, assuming abundant high-resource data (no repetition).
\textbf{Multi-lingual training} mixes the target-language corpus with a high-resource subset, with target language ratio $r=kD_T/D$, where $D$ is the total number of training tokens.
\textbf{Two-stage training} changes target language ratio from $r_1$ (first stage) to $r_2$ (second stage).

\subsection{Languages and Training Configuration}
\label{subsec:training_configuration}

In our experiments, we used three language pairs with English as the high-resource language: Japanese, Indonesian, and Swahili, all from different language families.
Japanese and Indonesian, though not typically low-resource, enable systematic analysis across data regimes through downsampling their corpora.
Swahili complements these with verification on actually lower-resource language, although its limited availability restricts experiments to smaller data settings.
We used C4 \cite{raffel2020exploring} for English and mC4 \cite{xue-etal-2021-mt5} for the target languages, both from Common Crawl with consistent preprocessing except for language-specific ones, minimizing the impact of data quality variations on our analysis.
We respectively used their original train/validation splits with the BLOOM tokenizer \cite{le2023bloom}.

We followed the training configuration of \citet{bi2024deepseek}: a decoder-only LLaMA architecture \cite{touvron2023llama} with context length $S=4096$, and their compute-dependent power-law schedules for learning rate and batch size.
We measured model scale by non-embedding FLOPs per token $M=72 n{d_\text{model}}^2+12n{d_\text{model}}S$, where $n$, $d_\text{model}$, and $S$ are the number of layers, the size of hidden states, and the sequence length, respectively.
For two-stage training, we re-warmed the learning rate at the start of the second stage to the first-stage maximum, as \citet{ibrahim2024simple} found that re-warming is effective for adaptation to new data in the later stage.
See Appendix \ref{subsec_appendix_pretrain} for detailed training configurations.

\subsection{Hyperparameter Search Space}
\label{subsec_hyperparameters}

To systematically explore the search space across multiple orders of magnitude, we parametrized each training setup by integer factors $(f_r,f_M,f_k,f_C)$ defining the language ratio $r=\frac{1}{2^{f_r}}$, model scale $M=\frac{M_0}{2^{f_M}}$, number of epochs $k=2^{f_k}$, and compute budget $C=2^{f_C}\cdot C_0$.
The resulting target language corpus size is $D_T = 2^{-f_r+f_M-f_k+f_C}\cdot D_{T,0}$, so setups sharing the same $(f_C,f_D)$ with $f_D=-f_r+f_M-f_k+f_C$ have identical $(C,D_T)$.
The reference values $(C_0, D_{T,0}, M_0)$ follow the monolingual compute-optimal scaling law of \citet{bi2024deepseek}.
See Appendix \ref{subsec_hyperparam_one_stage} for the exact values and the grid of factors explored (Table \ref{tab:search_space_factors}).
A setup involves multi-epoch training when $f_k>0$ and multi-lingual training when $f_r>0$.

For two-stage training, we additionally specified the stage-wise language ratios $(r_1,r_2)$ with $r_1<r_2$, and used the average ratio $r=s_1r_1+s_2r_2$ (which coincides with $kD_T/D$ from the single-stage definition) to compute the stage proportions via $s_1=\frac{r_2-r}{r_2-r_1}$ and $s_2=1-s_1$.
We grid-searched $r_1\in \{ 1/2, 1/4, 1/8, 1/16, 1/32, 0 \}$ and $r_2\in\{ 1/4, 1/2, 3/4, 1 \}$, combined with the same $(f_r, f_M, f_k, f_C)$ grid as in single-stage training.

\section{Scaling Law Analysis}
\label{sec:scaling_law}

In this section, we integrate and extend existing scaling laws to propose the $M^3$ Scaling Law, a unified scaling law that jointly covers multi-epoch, multi-lingual, and multi-stage training.

\paragraph{Notation.}
Throughout §\ref{sec:scaling_law}, $D'$ and $M'$ denote the
effective data and model sizes, whose definitions are
progressively extended across subsections
(Muennighoff in Eqs.~\ref{eq:effective_D_muennighoff}--\ref{eq:effective_M_muennighoff};
ATLAS in Eq.~\ref{eq:effective_D_longpre};
the $M^3$ Scaling Law in Eqs.~\ref{eq:m3_Dprime}--\ref{eq:m3_Mprime}).
Within each subsection, $D'$ and $M'$ refer to the most recent
definition unless otherwise noted.
See Appendix~\ref{appendix:notation} for a full summary of notation.

\subsection{Background}

Our approach builds on two existing scaling laws.
First, \citet{he2024scaling} showed that in single-stage single-epoch multi-lingual training, the target language validation loss follows a power law with respect to the target language ratio $r$:
\begin{align}
  &L_\text{He}(M, D, r) = L_\text{base}(M, D) \cdot \mathcal{R}(r), \label{eq:he_model_main} \\
  &L_\text{base}(M, D) = \frac{A}{M^\alpha} + \frac{B}{D^\beta} + E, \label{eq:chinchilla}
\end{align}
where $D = kD_T/r$ is the total number of training tokens, $L_\text{base}$ is the Chinchilla scaling law \cite{hoffmann2022training} with coefficients $A$, $B$, exponents $\alpha$, $\beta$, and irreducible loss $E$, $\mathcal{R}(r) = r^{-\gamma}$ is the language ratio factor, and $\gamma$ is the power-law exponent.

Second, \citet{muennighoff2024scaling} proposed the following scaling law for monolingual single-stage multi-epoch training, replacing $M$ and $D$ in $L_\text{base}$ with effective sizes $M'$ and $D'$:
\begin{align}
  &L_\text{Mu}(M', D') = L_\text{base}(M', D'), \label{eq:muennighoff_main} \\
  &D' = D_T \cdot h(R_D;\, R_D^*), \label{eq:effective_D_muennighoff}\\
  &M' = U_M \cdot h(R_M;\, R_M^*), \label{eq:effective_M_muennighoff}
\end{align}
where $h(R;\, R^*) = 1 + R^* \cdot (1 - \exp(-R/R^*))$ is a saturation function with $R^*$ controlling the rate of saturation.
Here, $R_D = k - 1$ is the data repetition count, $U_M = \min(G^{(\alpha+\beta)/\alpha}\, {D_T}^{\beta/\alpha},\, M)$ with $G = (\alpha A / \beta B)^{1/(\alpha+\beta)}$ is the compute-optimal model size for $D_T$ under single-epoch training, and $R_M = M / U_M - 1$ is the excess model size.
\citet{longpre2025atlas} extended this scaling law to the multi-lingual setting by adding the effective data size of other languages to $D'$ with a constant weight for each language.
In our setup, where the high-resource language is not repeated, this reduces to:
\begin{equation}
  D'=D_T\cdot h(R_D;R_D^*) + g \cdot D_\text{high},
  \label{eq:effective_D_longpre}
\end{equation}
where $D_\text{high}=kD_T(1-r)/r$ denotes the number of high-resource language tokens used during training, and $g$ is a constant weight.

\subsection{Extending Effective Model Size of Muennighoff Model}
\label{subsec:extending_effective_sizes}

\begin{figure}[t]
  \centering
  \includegraphics[width=\columnwidth]{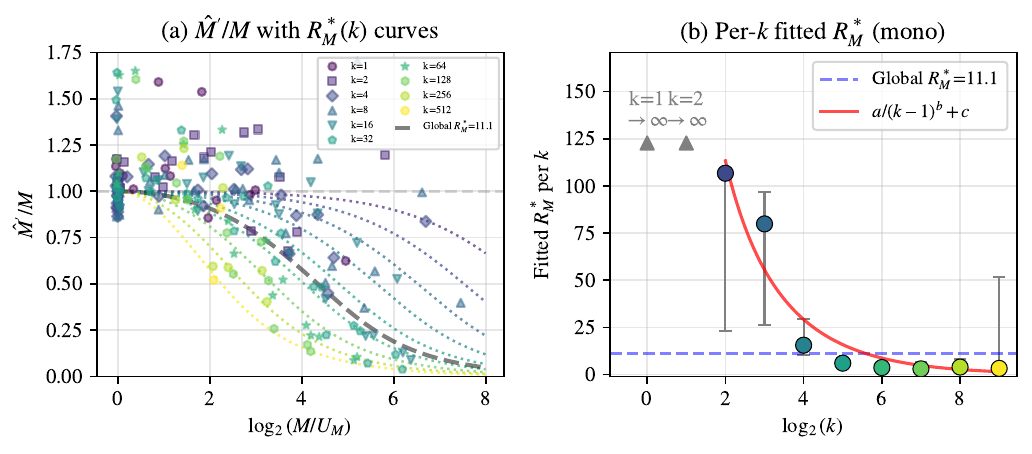}
  \caption{Empirical analysis of effective model size $M'$ on Japanese monolingual data.
  \textbf{(a)} $\hat{M}'/M$ vs.\ $\log_2(M/U_M)$ with per-$k$ $R_M^*(k)$ curves (dotted, colored by $k$) and the global constant $R_M^*$ (black dashed).
  \textbf{(b)} Per-$k$ fitted $R_M^*$, showing a decreasing trend from $\infty$ at $k=1$ to a finite floor $c$; the red curve shows the power-law fit $a/(k{-}1)^b + c$. Error bars in (b) show 95\% bootstrap confidence intervals (200 resamples).}
  \label{fig:effective_sizes}
\end{figure}

The Muennighoff form $M' = U_M \cdot h(R_M; R_M^*)$ uses a single $R_M^*$ shared across all cases, but we find that the strength of the saturation in fact depends on number of epochs $k$.
To isolate the behavior of $M'$, we invert $L_\text{Mu}$ for each data point and compute estimated effective model size $\hat{M}'$ from the observed loss $L_\text{actual}$ as
$\hat{M}'=(A/(L_\text{actual} - B/{D'}^\beta - E))^{1/\alpha}$
with $(A, B, \alpha, \beta, E)$ fit and fixed from the monolingual $k \leq 4$ subset and $D'$ computed via $h$ (Eq.~\ref{eq:effective_D_muennighoff}) with $R_D^*$ obtained by fitting $L_\text{Mu}$ to all monolingual data while keeping the above parameters fixed.
As shown in Figure \ref{fig:effective_sizes}(a), while $\hat{M}'/M$ has a decreasing trend with $M/U_M$ as the Muennighoff form predicts, the trend also depends on $k$.
For small $k$, e.g., $k=1,2$, $\hat{M}'/M$ shows a small decay, while for larger $k$, $\hat{M}'/M$ decreases faster at smaller $M/U_M$.
This suggests saturation of $M'$ also depends on $k$ in addition to $M/U_M$.

Refitting $R_M^*$ independently for each $k$ on the monolingual data confirms this trend (Figure \ref{fig:effective_sizes}(b)).
The estimated $R_M^*$ has a trend to decrease and saturate to a finite floor as $k$ increases.
We propose to model this behavior with
\begin{equation}
  R_M^*(k) = \frac{a}{(k-1)^b} + c,
  \label{eq:rm_star_mono}
\end{equation}
which diverges as $k \to 1$ and converges to $c$ as $k \to \infty$.

This form has desirable limiting behavior: as $k \to 1$, $R_M^* \to \infty$ and $h(R_M; R_M^*) \to 1 + R_M = M/U_M$, yielding $M' \to M$ and recovering Chinchilla at $k=1$.
This fixes an issue of the original Muennighoff form, where a constant $R_M^*$ predicts $M' < M$ whenever $M > U_M$, deviating from Chinchilla even without multi-epoch effects.

\subsection{Dual Power Law for Multi-Stage Training}
\label{subsubsec:dual_power_law_two_stage}

Here, we extend the language ratio factor $\mathcal{R}$ of the He model (Eq.~\ref{eq:he_model_main}) to multi-stage training.
We first analyze two-stage as a pilot case.

\subsubsection{Pilot Analysis for Two-Stage Training}
\label{subsubsec:pilot_analysis_two_stage}

\begin{figure*}[t]
  \centering
  \includegraphics[width=\textwidth]{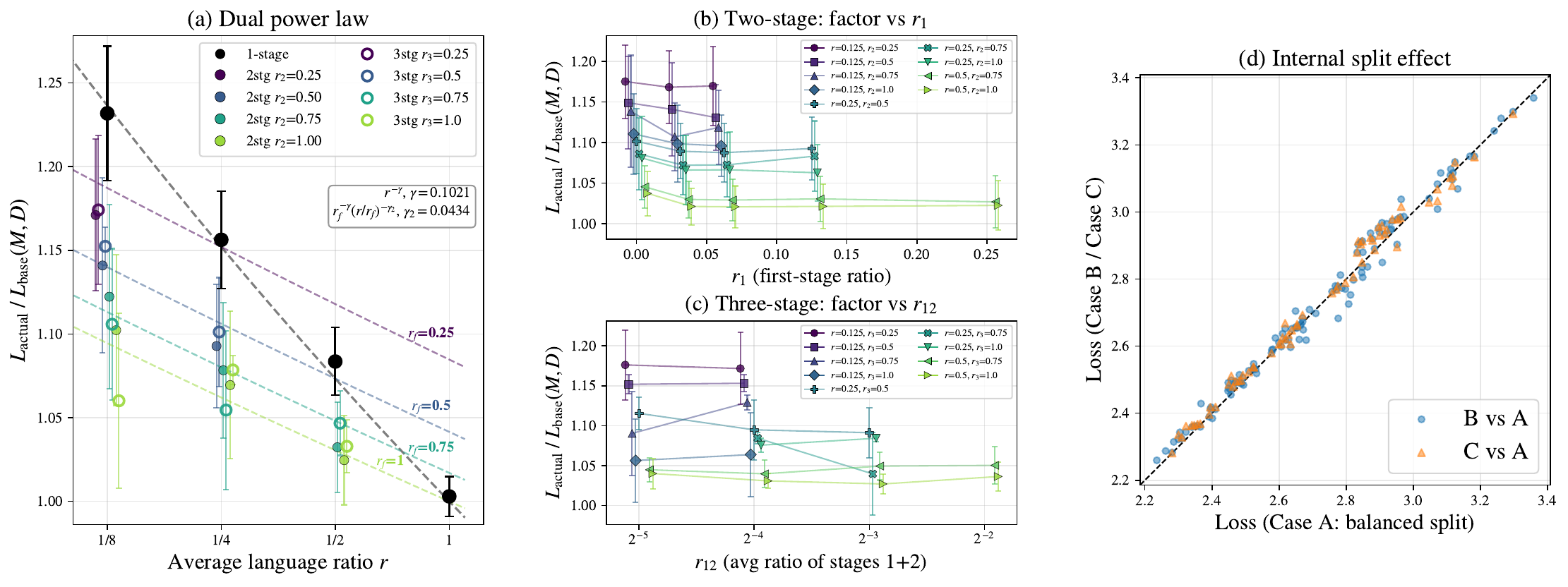}
  \caption{Verification of the dual power law for multi-stage training ((Japanese, English) pair, $k=1$ only).
  \textbf{(a)} Ratio $L_\text{actual} / L_\text{base}(M,D)$ vs.\ average language ratio $r$. Filled and hollow symbols denote two-stage and three-stage setups, respectively, colored by final-stage ratio $r_f$ ($r_2$ or $r_3$). Dashed lines show $r_f^{-\gamma}(r/r_f)^{-\gamma_2}$ fitted on \textbf{two-stage data}.
  \textbf{(b, c)} $L_\text{actual} / L_\text{base}(M,D)$ vs.\ first-stage ratio $r_1$ (two-stage) and vs.\ stages 1--2 average ratio $r_{12}$ (three-stage), showing near-flat trends.
  \textbf{(d)} Loss comparison across three $(r_1, r_2)$ patterns (A, B, C) at matched $(r_3, r_{12}, r, M, D_T)$, confirming negligible effect of the internal split.
  Error bars show standard deviations across different $(M, D_T)$ configurations sharing the same ratio settings; points are jittered horizontally for visibility.}
  \label{fig:Lr_factor_k1}
\end{figure*}

We fitted $L_\text{base}$ (Eq.~\ref{eq:chinchilla}) to monolingual single-epoch data and computed $L_\text{actual} / L_\text{base}(M,D)$ as an empirical proxy for $\mathcal{R}$ on $k=1$ two-stage setting.
Figure \ref{fig:Lr_factor_k1}(a) plots this ratio against $r$ for both single-stage (black circles) and two-stage (colored symbols) setups, averaged over $(M,D_T)$.

As shown in the figure, the single-stage data follow the power law $r^{-\gamma}$ ($\gamma = 0.1021$), consistent with \citet{he2024scaling}.
On the other hand, two-stage settings systematically exhibit lower ratios at matched $r$, and those sharing the same $r_2$ tend to trace a common slope differing from $-\gamma$.
On the other hand, Figure \ref{fig:Lr_factor_k1}(b) shows that $r_1$ has a much weaker effect than $r$ and $r_2$, with only a slight increase at lower $r_1$\footnote{See Appendix \ref{subsec:r1_effect} for a quantitative analysis}.

\subsubsection{Proposed Dual Power Law}

Based on observations in \S\ref{subsubsec:pilot_analysis_two_stage}, we propose the following dual power law model for the language ratio factor that unifies single-stage and multi-stage training:
\begin{equation}
  \mathcal{R}_\text{dual}(r, r_f) = \begin{cases}
    r^{-\gamma} & \text{(single-stage)} \\
    r_f^{-\gamma} \cdot \left(\dfrac{r}{r_f}\right)^{-\gamma_2} & \text{(multi-stage)}
  \end{cases}
  \label{eq:dual_power_law}
\end{equation}
where $r_f$ is the final-stage language ratio, $\gamma$ is shared with single-stage training, and $\gamma_2$ is an additional exponent corresponding to the common slope of multi-stage training.
Note that when $r = r_f$ (i.e., effectively single-stage training), the multi-stage formula reduces to $r_f^{-\gamma} = r^{-\gamma}$, so the two cases are consistent.

As shown in Figure \ref{fig:Lr_factor_k1}(a), fixing $\gamma$ to the value obtained from single-stage data and fitting $\gamma_2$ yields $\gamma_2 = 0.0434$, which provides a good fit to the two-stage data.
The fact that $\gamma_2 < \gamma$ indicates that two-stage training mitigates the loss increase associated with decreasing $r$ compared to single-stage training, suggesting the effectiveness of concentrating target language data in the second stage.

\subsubsection{Verification of Dual Power Law for Multi-Stage Training}

The dual power law (Eq.~\ref{eq:dual_power_law}) suggests that $\mathcal{R}$ depends only on the average language ratio $r$ and the final-stage language ratio $r_f$, independent of intermediate stage ratios.
To verify that this behavior generalizes to multi-stage training, we conducted three-stage training experiments.

More specifically, we trained three-stage models on (Japanese, English) at $k=1$, sweeping $r$, $M$, $D_T$, and the final-stage ratio $r_3$, together with the stages 1--2 average $r_{12}$.
At each matched $(r_3,r,r_{12},M,D_T)$, we additionally compared three internal split patterns: \textbf{(A)} $(r_1,r_2)$ bracketing $r_{12}$, and \textbf{(B,C)} $r_1=0$ with $r_2$ set to the smallest or largest valid value above $r_{12}$.
Since exhaustive enumeration of all valid $(r_1, r_2)$
combinations is infeasible, these three patterns were chosen as
representative configurations spanning the range of feasible
splits.
See Appendix~\ref{subsec:3stage_details} for implementation details.

Figure \ref{fig:Lr_factor_k1}(a) shows that three-stage setups also follow the dual power law with the same $\gamma$ and $\gamma_2$ fitted on two-stage data ($r_3$ taking the role of $r_f$), and panels (c,d) confirm that the ratio is insensitive to $r_{12}$ and the internal split $(r_1,r_2)$.
These results suggest that the same dual power law $\mathcal{R}_\text{dual}(r,r_f)$ governs multi-stage training in general, depending only on the average and final-stage ratios.

\subsection{Modeling Interaction of Multi-Epoch and Multi-Lingual Training}
\label{subsec:interaction_multi_epoch_multi_lingual}

For multi-epoch multi-lingual training, we consider $L = L_\text{base}(M', D') \cdot \mathcal{R}_\text{dual}(r, r_f)$ with $M'$ from Eq.~(\ref{eq:effective_M_muennighoff}) and $D'$ extending the multi-lingual form of \citet{longpre2025atlas} (Eq.~\ref{eq:effective_D_longpre}).
The $(k, r)$ dependence can be carried by either $R_M^*$ (through $M'$) or the weight $g$ on $D_\text{high}$; preliminary experiments indicated that placing the dependence solely on $g$ yields the best $M$-axis extrapolation at a minor cost in $k$-axis accuracy, while involving $R_M^*$ degrades $M$-axis extrapolation. We therefore adopt the $g$-only route, keeping $R_M^*$ as a single global constant.
As epochs increase, the effective weight of high-resource tokens is expected to decay due to target-language data repetition; however, when $r$ is small, repeated target-language tokens constitute a smaller fraction of the total training data, weakening this decay.
At $r = 0$ no data is repeated and $g$ should remain~$1$.
Based on this hypothesis, we promote $g$ to a function of $(r, R_D)$:
\begin{align}
  &g(r, R_D; R_{D,\text{high}}^*, \psi) = \notag \\
  &\quad (1-r)^\psi + \bigl[1-(1-r)^\psi\bigr]
    \exp\!\left(-\tfrac{R_D}{R_{D,\text{high}}^*}\right),
  \label{eq:g_func}
\end{align}
which gives $g = 1$ at $R_D = 0$ (recovering \citet{he2024scaling}) and $g \to (1-r)^\psi$ as $R_D \to \infty$.
Although this means the monolingual $k$-dependence $R_M^*(k)$ obtained in \S\ref{subsec:extending_effective_sizes} is not used here, it is incorporated in a separate monolingual variant introduced in \S\ref{subsec:proposed_model}.
See Appendix~\ref{appendix:gfit_details} for a more detailed analysis of the interaction between multi-epoch and multi-lingual training.

\subsection{Proposed Model: $M^3$ Scaling Law}
\label{subsec:proposed_model}

Combining the extensions in \S\ref{subsec:extending_effective_sizes}, \S\ref{subsubsec:dual_power_law_two_stage}, and \S\ref{subsec:interaction_multi_epoch_multi_lingual}, we propose the \textbf{$M^3$ Scaling Law}, a unified scaling law that jointly captures multi-epoch, multi-lingual, and multi-stage training:
\begin{equation}
\begin{aligned}
  &L_{M^3}(M, D_T, k, r, r_f) \\
  &\quad = \left(\frac{A}{{M'}^\alpha} + \frac{B}{{D'}^\beta} + E\right) \cdot \mathcal{R}_\text{dual}(r, r_f),
\end{aligned}
\label{eq:m3_main}
\end{equation}
where $\mathcal{R}_\text{dual}$ is the dual power law defined in Eq.~(\ref{eq:dual_power_law}).
The effective data and model sizes are
\begin{align}
  D' &= D_T \cdot h(R_D;\, R_D^*) \;+\; g(r,R_D) \cdot D_\text{high},
        \label{eq:m3_Dprime}\\
  M' &= U_M \cdot h(R_M;\, R_M^*),
        \label{eq:m3_Mprime}
\end{align}
where $R_D$, $R_M$, $U_M$, $G$, and $D_\text{high}$ are as defined in Eqs.~(\ref{eq:effective_D_muennighoff})--(\ref{eq:effective_D_longpre}), $h$ is the saturation function of Eq.~(\ref{eq:effective_M_muennighoff}), $g$ follows Eq.~(\ref{eq:g_func}), and $R_M^*$ is a single global constant shared across all $(k, r)$.
This form has 11 free parameters: $A, B, \alpha, \beta, E, R_D^*, R_{D,\text{high}}^*, \psi, R_M^*, \gamma$, and $\gamma_2$.

We also consider a variant that incorporates the $k$-dependent $R_M^*(k)$ from Eq.~(\ref{eq:rm_star_mono}) for monolingual multi-epoch settings ($r=1$).
In this case, $D_\text{high}=0$ and $g$ does not participate, so the interaction between multi-epoch and multi-lingual training need not be considered and $R_M^*(k)$ can be incorporated without degrading $M$-axis extrapolation.
We denote this variant $M^3 + R_M^*(k)$ and evaluate it in the monolingual multi-epoch setting.
To summarize the role of $R_M^*$ across the paper: the per-$k$ form $R_M^*(k)$ from Eq.~(\ref{eq:rm_star_mono}) is used only in this monolingual variant; in the main $M^3$ Scaling Law, $R_M^*$ is a single constant.

\subsection{Evaluation of $M^3$ Scaling Law}
\label{subsec:eval_m3}

We evaluate the proposed $M^3$ Scaling Law against existing scaling laws---Chinchilla \cite{hoffmann2022training}, He \cite{he2024scaling}, Muennighoff \cite{muennighoff2024scaling}, ATLAS \cite{longpre2025atlas}, and Sedova \cite{sedova2026scalinglawsmixturepretraining}---under four settings (Table~\ref{tab:baselines_extended}):
(a) all setups combined;
(b) single-stage setups only, the regime targeted by most prior work;
(c) monolingual ($r{=}1$) multi-epoch setups, providing a focused comparison against methods that target data repetition \cite{muennighoff2024scaling,sedova2026scalinglawsmixturepretraining}; and
(d) multi-lingual single-epoch ($k{=}1$) setups, where models are fit on the combined 1-stage and 2-stage training data and evaluated separately on 1-stage (d1) and 2-stage (d2) test sets, as well as their macro-average (d3). For reference, we additionally include methods focusing on continual pretraining, the D-CPT Law \cite{que2024d}, PTPP \cite{goffinet2025ptpp}, and \citet{zhang2024scaling}, fit on 2-stage training data only and evaluated on the 2-stage test set (d2).

Extrapolation accuracy is measured by the coefficient of determination ($R^2$) on a held-out test set.
To avoid sensitivity to any single train-test split, we construct multiple splits along each of the training setup variables $(C, M, D_T, D, r, k)$, average $R^2$ over splits within each language, and report the unweighted mean across languages (see Appendix~\ref{subsec:scaling_law_evaluation} for details).

The $M^3$ Scaling Law achieves the best average $R^2$ across settings~(a)--(c).
In setting~(d), it achieves the highest accuracy on the combined 1-stage and 2-stage evaluation (d3), outperforming all baselines, which demonstrates that the dual power law enables strong extrapolation on both single-stage and two-stage data simultaneously.
On the 2-stage test set alone (d2), the $M^3$ Scaling Law achieves comparable performance to methods specifically designed for continual pretraining, despite being trained on the combined 1-stage and 2-stage data rather than 2-stage data alone.

Ablations clarify the contribution of each extension.
The dual power law is essential for jointly achieving good extrapolation on both single-stage and two-stage data: in setting~(d), models that incorporate it (He~+~Dual~PL and $M^3$) consistently outperform their counterparts without it (He and $M^3$~$-$Dual~PL).
In the monolingual setting~(c), the $M^3 + R_M^*(k)$ variant---which incorporates the $k$-dependent $R_M^*(k)$ into $M'$---achieves the best accuracy, exceeding all baselines.
Finally, removing the high-resource weight $g$ ($-g$ in part~(a)) degrades extrapolation along the $D$ and $k$ axes, while slightly improving $M$-axis extrapolation.

\begin{table*}[t]
  \centering
  \begin{minipage}[t]{0.48\textwidth}
  \centering
  \scalebox{0.72}{
  \setlength{\tabcolsep}{3pt}
  \begin{tabular}{@{}l@{\hskip 4pt}lc*{6}{r}r@{}}
    \hline
    & Model & \#P & $C$ & $M$ & $D_T$ & $D$ & $r$ & $k$ & Avg \\
    \hline
    \hline
    \multicolumn{10}{@{}l}{\textbf{(a)} \textit{All data (1-stage $+$ 2-stage)}} \\
    \hline
    \multicolumn{10}{@{}l}{\textit{Baseline models}} \\
    \hline
    & Chinchilla & 5 & .43 & $-$.26 & .58 & .09 & .38 & $-$.05 & .19 \\
    & He & 6 & .39 & .09 & .79 & $-$1.12 & .40 & $-$.73 & $-$.03 \\
    & Muennighoff & 7 & .55 & .16 & .78 & .13 & $-$.30 & .22 & .25 \\
    & ATLAS & 7 & .54 & .61 & .71 & $-$.88 & .58 & .52 & .35 \\
    & Sedova & 9 & .77 & .52 & .80 & .44 & .75 & .42 & .62 \\
    \hline
    \multicolumn{10}{@{}l}{\textit{Proposed model and ablations}} \\
    \hline
    & $M^3$ Scaling Law & 11 & \textbf{.79} & .60 & .85 & \textbf{.52} & \textbf{.75} & .52 & \textbf{.67} \\
    & \quad $-$Dual PL & 10 & .74 & .57 & .78 & .34 & .63 & \textbf{.53} & .60 \\
    & \quad $-g$ & 9 & .70 & \textbf{.63} & \textbf{.88} & $-$.41 & .75 & .34 & .48 \\
    \hline
    \hline
    \multicolumn{10}{@{}l}{\textbf{(b)} \textit{1-stage data only}} \\
    \hline
    \multicolumn{10}{@{}l}{\textit{Baseline models}} \\
    \hline
    & Chinchilla & 5 & .31 & $-$.01 & .35 & .32 & .01 & $-$.18 & .13 \\
    & He & 6 & .21 & .19 & .90 & $-$1.05 & .42 & $-$1.24 & $-$.10 \\
    & Muennighoff & 7 & .55 & .40 & .75 & .18 & $-$.05 & .27 & .35 \\
    & ATLAS & 7 & .61 & .59 & .70 & $-$.20 & .40 & .46 & .43 \\
    & Sedova & 9 & .71 & .49 & .86 & \textbf{.68} & .72 & .21 & .61 \\
    \hline
    \multicolumn{10}{@{}l}{\textit{Proposed model}} \\
    \hline
    & $M^3$ Scaling Law & 11 & \textbf{.79} & \textbf{.70} & \textbf{.91} & .63 & \textbf{.80} & \textbf{.48} & \textbf{.72} \\
    \hline
    \hline
    \multicolumn{10}{@{}l}{\textbf{(c)} \textit{Multi-epoch only (monolingual, 1-stage)}} \\
    \hline
    \multicolumn{10}{@{}l}{\textit{Baseline models}} \\
    \hline
    & Muennighoff & 7 & .74 & .71 & .90 & .90 & --- & .47 & .75 \\
    & ATLAS & 7 & .70 & .46 & \textbf{.91} & .86 & --- & .47 & .68 \\
    & Sedova & 9 & .79 & .10 & .82 & .90 & --- & .43 & .61 \\
    \hline
    \multicolumn{10}{@{}l}{\textit{Proposed model and ablation}} \\
    \hline
    & $M^3{+}R_M^*(k)$ & 9 & \textbf{.88} & \textbf{.79} & .89 & \textbf{.92} & --- & \textbf{.77} & \textbf{.85} \\
    & $M^3$ Scaling Law & 7 & .74 & .71 & .90 & .90 & --- & .47 & .75 \\
    \hline
  \end{tabular}
  }
  \end{minipage}%
  \hfill
  \begin{minipage}[t]{0.50\textwidth}
  \centering
  \scalebox{0.72}{
  \setlength{\tabcolsep}{3pt}
  \begin{tabular}{@{}l@{\hskip 4pt}lc*{5}{r}r@{}}
    \hline
    & Model & \#P & $C$ & $M$ & $D_T$ & $D$ & $r$ & Avg \\
    \hline
    \hline
    \multicolumn{9}{@{}l}{\textbf{(d1)} \textit{Multi-lingual only (single-epoch): 1-stage test}} \\
    \hline
    \multicolumn{9}{@{}l}{\textit{Baseline models}} \\
    \hline
    & He & 6 & .51 & .06 & .70 & .28 & $-$.27 & .26 \\
    & He $+$ Dual PL & 7 & .95 & .72 & \textbf{.89} & \textbf{.87} & \textbf{.82} & \textbf{.85} \\
    & Sedova & 9 & .62 & .47 & .79 & .51 & $-$.19 & .44 \\
    \hline
    \multicolumn{9}{@{}l}{\textit{Proposed model and ablation}} \\
    \hline
    & $M^3$ Scaling Law & 8 & \textbf{.95} & \textbf{.85} & .85 & .76 & .82 & .85 \\
    & \quad $-$Dual PL & 7 & .63 & .42 & .63 & .12 & $-$.16 & .33 \\
    \hline
    \hline
    \multicolumn{9}{@{}l}{\textbf{(d2)} \textit{Multi-lingual only (single-epoch): 2-stage test}} \\
    \hline
    \multicolumn{9}{@{}l}{\textit{Baseline models}} \\
    \hline
    & He & 6 & .67 & $-$.11 & .85 & .04 & .65 & .42 \\
    & He $+$ Dual PL & 7 & .75 & .01 & .88 & .41 & .73 & .56 \\
    & Sedova & 9 & \textbf{.87} & \textbf{.60} & \textbf{.93} & \textbf{.50} & \textbf{.86} & \textbf{.75} \\
    \hline
    \multicolumn{9}{@{}l}{\textit{2-stage-only baselines$^\dagger$}} \\
    \hline
    & D-CPT Law & 8 & (.67) & ($-$.03) & (.86) & (.70) & (\textbf{.82}) & (.60) \\
    & PTPP-F1 & 10 & (.58) & (.42) & (.90) & (\textbf{.88}) & (.70) & (\textbf{.70}) \\
    & PTPP-F2 & 10 & (.55) & ($-$.23) & (.86) & (.68) & (.78) & (.53) \\
    & PTPP-F3 & 12 & (\textbf{.68}) & (\textbf{.46}) & (\textbf{.91}) & (.86) & ($-$.15) & (.55) \\
    & Zhang & 6 & (.18) & ($-$1.45) & (.61) & ($-$1.82) & (.60) & ($-$.38) \\
    \hline
    \multicolumn{9}{@{}l}{\textit{Proposed model and ablation}} \\
    \hline
    & $M^3$ Scaling Law & 8 & .84 & .36 & .86 & .28 & .78 & .63 \\
    & \quad $-$Dual PL & 7 & .84 & .40 & .82 & $-$.23 & .75 & .52 \\
    \hline
    \hline
    \multicolumn{9}{@{}l}{\textbf{(d3)} \textit{Multi-lingual only (single-epoch): both stages}} \\
    \hline
    \multicolumn{9}{@{}l}{\textit{Baseline models}} \\
    \hline
    & He & 6 & .59 & $-$.03 & .78 & .16 & .19 & .34 \\
    & He $+$ Dual PL & 7 & .85 & .36 & \textbf{.89} & \textbf{.64} & .78 & .70 \\
    & Sedova & 9 & .74 & .53 & .86 & .50 & .33 & .59 \\
    \hline
    \multicolumn{9}{@{}l}{\textit{Proposed model and ablation}} \\
    \hline
    & $M^3$ Scaling Law & 8 & \textbf{.90} & \textbf{.61} & .86 & .52 & \textbf{.80} & \textbf{.74} \\
    & \quad $-$Dual PL & 7 & .73 & .41 & .73 & $-$.05 & .29 & .42 \\
    \hline
  \end{tabular}
  }
  \end{minipage}
  \caption{Extrapolation accuracy (Test $R^2$) of baseline and proposed scaling law models across Japanese, Indonesian, and Swahili.
  \textit{Left:}
  \textbf{(a)} All (1-stage $+$ 2-stage) data.
  \textbf{(b)} 1-stage data only.
  \textbf{(c)} Multi-epoch only: monolingual ($r{=}1$) data; the $r$ axis is not applicable.
  \textit{Right:}
  \textbf{(d1, d2)} Multi-lingual only: single-epoch ($k{=}1$) data evaluated on 1-stage and 2-stage test sets separately; the $k$ axis is not applicable. \textbf{(d3)} Both stages (macro-average of (d1) and (d2)).
  $^\dagger$2-stage-only models are trained on 2-stage data only; scores in parentheses are not directly comparable to other models trained on the combined data.
  Each column shows the $R^2$ macro-averaged over held-out splits and languages along the corresponding axis (see Appendix~\ref{subsubsec:aggregation} for details). \textit{Avg} is the unweighted mean across applicable axes. \#P denotes the number of free parameters.}
  \label{tab:baselines_extended}
\end{table*}

\section{Predicting Optimal Training Setups with $M^3$ Scaling Law}

\subsection{Optimal Training Approach}
\label{subsec:optimal_training_approach}

Since empirically $\gamma_2 < \gamma$, the dual power law (Eq.~\ref{eq:dual_power_law}) at fixed $r$ is minimized at $r_f = 1$, giving $r^{-\gamma_2}$ for two-stage training, which is lower than the single-stage value $r^{-\gamma}$ at any $r < 1$ under matched $(M, D_T, k)$.

To examine which category is optimal under realistic constraints, we numerically minimize the $M^3$ Scaling Law over training setups within each of three categories---\textbf{mono 1-stage}, \textbf{multi 1-stage}, and \textbf{multi 2-stage}---for each $(C, D_T)$.
As shown in Figure~\ref{fig:main_total_graph}, mono 1-stage is optimal at large $D_T$ and multi 2-stage at small $D_T$, while multi 1-stage is never optimal in either regime.
The same pattern holds for the empirically optimal setups across all languages and $(C, D_T)$ in our grid search.

\subsection{Optimal Number of Epochs}

\begin{figure}[t]
  \centering
  \includegraphics[width=\columnwidth]{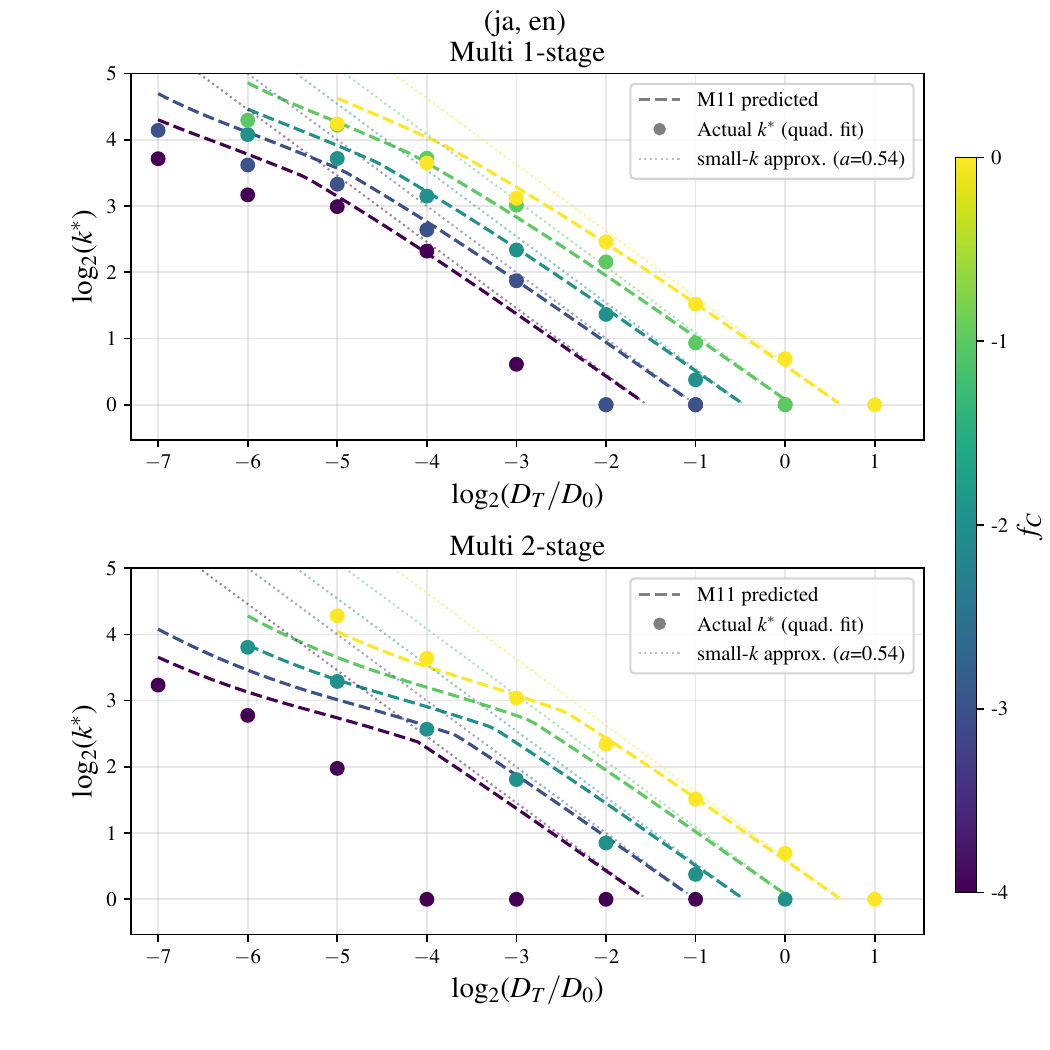}
  \caption{Optimal number of epochs $k^*$ vs.\ target language corpus size $D_T$ for (Japanese, English), shown for \textbf{multi 1-stage} (top) and \textbf{multi 2-stage} (bottom). Dashed lines show $k^*$ predicted by numerically minimizing the $M^3$ Scaling Law at each $(C, D_T)$; filled circles show empirical $k^*$ estimated by quadratic fitting of $L^*(C, D_T, k)$ in $\log_2 k$. Dotted lines show the small-$k$ approximation $k^* \approx D^*(C)/D_T$. Color encodes compute budget $C$ via $f_C = \log_2(C/C_0)$.}
  \label{fig:main_enja_D_T_optimal_k_merge}
\end{figure}

The $M^3$ Scaling Law predicts that the optimal number of epochs depends on $(C, D_T)$ only through the scarcity variable $D_T/D^*(C)$, where $D^*(C) \propto C^{\alpha/(\alpha+\beta)}$ is the monolingual compute-optimal corpus size. Specifically, under the compute constraint $C = MD$ with $D = kD_T/r$ and for a fixed ratio schedule $\mathbf{r} = (r, r_1, r_2)$ specifying the average and per-stage target-language ratios, one obtains
\begin{equation}
  k_{\mathbf{r}}^*(C, D_T) = K_{\mathbf{r}}\!\left(\frac{D_T}{D^*(C)}\right)
  \label{eq:k_star_collapse}
\end{equation}
for some function $K_{\mathbf{r}}$\footnote{The proof is given in Appendix~\ref{appendix:optimal_epochs_derivation}}. The relation is exact for fixed $\mathbf{r}$ (covering \textbf{mono 1-stage}) and approximate under joint optimization of $\mathbf{r}$ (\textbf{multi 2-stage}). In the small-$k$ regime where the saturations of $M'$ and $D'$ are negligible, $K_{\mathbf{r}}$ further reduces to $k_{\mathbf{r}}^*(C, D_T) \approx D^*(C)/D_T$, predicting a unit-slope line passing through $k^* = 1$ at $D_T = D^*(C)$ in the log-log plot.

Numerically minimizing the $M^3$ Scaling Law over $k$ separately for \textbf{multi 1-stage} and \textbf{multi 2-stage}, the resulting log-log curves of $k^*$ vs.\ $D_T$ at different $C$ are nearly parallel with uniform log-scale spacing (Figure~\ref{fig:main_enja_D_T_optimal_k_merge}), consistent with the predicted collapse along $D_T/D^*(C)$ and matching $k^* = D^*(C)/D_T$ in the small-$k$ regime.
Empirical estimates from quadratic fits of $L^*(C, D_T, k)$ in $\log_2 k$ broadly follow these predictions, with non-negligible scatter likely reflecting the limited $k$-grid resolution and residual fitting error\footnote{See Appendix~\ref{appendix:optimal_epochs} for the derivation and fit details.}.

\section{Related Work}

Continual pretraining, initially proposed for encoder-only models
\cite{chau-etal-2020-parsing,muller-etal-2021-unseen}, has been
increasingly adopted for low-resource language LLM training
\cite{luukkonen-etal-2023-fingpt,lin2024mala,fujii2024continual};
our findings corroborate its benefit while identifying when it becomes optimal.
As summarized in Table~\ref{table:scope_of_hyperparameters}, prior scaling law studies each cover a subset of the axes relevant to low-resource LLM pretraining, but none jointly addresses multi-epoch, multi-lingual, and multi-stage training.
Studies on complementary hyperparameters such as batch size, learning rate, and vocabulary size \cite{bi2024deepseek,hu2024minicpm,tao2024scaling} are orthogonal to our work.

\section{Conclusion}

We proposed the \textbf{$M^3$ Scaling Law}, a unified scaling law jointly covering multi-epoch, multi-lingual, and multi-stage training for low-resource language LLM pretraining.
It extrapolates more accurately than existing scaling laws across three language pairs, and reveals that the optimal training approach is always either monolingual single-stage or multi-lingual two-stage---never multi-lingual single-stage---and that the optimal number of epochs is governed by the scarcity variable $D_T/D^*(C)$.

\newpage

\section{Limitations}

\textbf{Evaluation Metrics:} Our evaluation of LLM performance is limited to the validation loss of the language modeling task on the low-resource target language (\S\ref{subsec:problem_setup}), and we do not directly evaluate downstream task performance.
Our training setups are relatively small-scale, and many downstream tasks exhibit near-zero performance at this scale due to the emergent nature of LLM capabilities \cite{wei2022emergent}, making direct analysis on downstream metrics infeasible.
Nevertheless, validation loss has been shown to correlate with downstream task performance in prior work \cite{brandfonbrener2024loss,chen2024scaling,du2024understanding,gadre2024language,thrush2024improving}, and our preliminary analysis (Appendix~\ref{section:downstream_task_performance}) is consistent with this finding, supporting validation loss as a reasonable proxy.

Beyond predicting aggregate target-language validation loss, a natural extension is to ask how training recipes affect different capabilities of LLMs.
Validation loss aggregates two qualitatively different capabilities: language-general abilities such as reasoning, which are expected to transfer well across languages, and language-specific knowledge, which transfers less.
Depending on which capability a downstream application emphasizes, the effect of multi-lingual or multi-stage training on transfer may differ from what aggregate validation loss suggests.
Disentangling how training recipes differentially affect these capabilities, and constructing scaling laws at the capability level, is a natural direction for future work---one for which the $M^3$ Scaling Law could serve as a starting point.

\textbf{Scope of Hyperparameters and Corpora:} While our experiments cover a broader scope of hyperparameters than prior work relevant to low-resource language LLM pretraining, we could not exhaustively explore all training hyperparameters due to compute constraints. 
Specifically, we did not search over batch size or learning rate; instead, we used the compute-dependent values optimized in \citet{bi2024deepseek}. 
We also used C4 \cite{raffel2020exploring} and mC4 \cite{xue-etal-2021-mt5} as our corpora throughout, chosen for their consistent preprocessing across languages and wide adoption in prior scaling law studies \cite{hoffmann2022training,muennighoff2024scaling}, and did not consider variations in target language corpus quality, the use of LLM-rephrased or synthetically generated data, or more recent multi-lingual corpora such as MADLAD-400 \cite{kudugunta2024madlad} and CulturaX \cite{nguyen-etal-2024-culturax}. 
This choice reflects our intent to fix reasonable defaults along these axes and concentrate experimental resources on the multi-epoch, multi-lingual, and multi-stage axes that are central to low-resource language LLM pretraining. 

A consequence of these fixed defaults is that some aspects of recipe design fall outside our analysis. 
The values from \citet{bi2024deepseek} are optimized for single-stage training, and different values may be optimal for two-stage training. 
Similarly, since C4 and mC4 share consistent preprocessing across languages, our analysis cannot characterize how asymmetries in corpus quality between high- and low-resource languages---common in practice---affect the optimal training recipe. 
A natural direction for future work is to extend scaling-law modeling to jointly cover corpus quality, data selection, and synthetic data generation, enabling the design of these components to be optimized alongside the recipe axes studied here. 
The $M^3$ Scaling Law offers a natural starting point for such investigations.

\textbf{Scope of Languages:} Our experiments used English as the high-resource language paired with three target languages from different language families: Japanese, Indonesian (relatively high-resource), and Swahili (lower-resource). 
This selection reflects three deliberate choices. 
First, we prioritized pairs from different language families, since cross-family transfer is weaker and deviates more from the monolingual setting, making such pairs more informative for studying how multi-lingual and multi-stage training depart from single-language behavior. 
Second, we fixed English as the high-resource language because English is by far the most common choice in practice for the high-resource side of low-resource LLM pretraining \cite[e.g.,][]{fujii2024continual}. 
Third, we focused on moderately low-resource languages rather than truly low-resource ones, since the latter typically lack sufficient corpus volume to support scaling-law analysis at all; building understanding from moderately low-resource settings is a natural first step. 
As a result, generalization to within-family pairs, non-English high-resource languages, or truly low-resource languages with orders of magnitude less data is not directly verified by our experiments. 

A related limitation is that our analysis models only the target-language validation loss and does not model the high-resource language loss. 
This reflects our problem setup, which focuses on improving LLM capabilities for the low-resource target language; improvements on the high-resource side are out of scope. 
However, applications such as machine translation require capabilities in both languages, and a recipe that minimizes target-language loss alone may not be Pareto-optimal across both. 
Extending the $M^3$ Scaling Law to jointly predict losses on both languages, and applying it across more diverse language pairs, is a natural direction for future work. 
The $M^3$ Scaling Law already exposes modeling primitives such as the dual power law exponent $\gamma_2$ that characterize previously uncharacterized aspects of cross-lingual transfer, and broader language coverage in such future studies would deepen the understanding of how these quantities vary across language pairs.

\begin{CJK}{UTF8}{ipxm}
\bibliography{anthology,custom}
\end{CJK}

\appendix

\section{Notation}
\label{appendix:notation}

Table~\ref{tab:notation} summarizes all symbols used in this paper.

\begin{table*}[t]
\centering
\scalebox{0.715}{
\begin{minipage}[t]{0.68\textwidth}
\begin{tabular}{@{}ll@{}}
\hline
\multicolumn{2}{@{}l@{}}{\textit{Compute, data, and model sizes}} \\
\hline
$C$ & Compute budget (FLOPs; $C{=}MD$) \\
$C_0$ & Reference compute budget ($10^{18}$) \\
$D$ & Total training tokens ($kD_T/r$) \\
$D_T$, $D_{T,0}$ & Target-language corpus size; reference \\
$D_\text{high}$ & High-resource tokens ($kD_T(1{-}r)/r$) \\
$D_1$, $D_2$ & Stage-1/2 tokens ($s_1 D$, $s_2 D$) \\
$M$, $M_0$ & Model scale (non-emb.\ FLOPs/token); ref. \\
$N$ & Model parameters (PTPP baselines only) \\
$S$ & Sequence/context length ($=4096$) \\
$n$, $d_\text{model}$, $n_\text{heads}$ & Layers, hidden dim, attention heads \\
\hline
\multicolumn{2}{@{}l@{}}{\textit{Training-recipe hyperparameters}} \\
\hline
$k$ & Number of target-corpus epochs \\
$r$ & Average target-language ratio ($kD_T/D$) \\
$r_1, r_2, r_3$ & Per-stage target-language ratios \\
$r_f$ & Final-stage target-language ratio \\
$r_{12}$ & Average ratio over stages 1--2 \\
$s_1, s_2, s_3$ & Stage proportions rel.\ to total length \\
$s_{12}$ & Combined proportion of stages 1--2 ($s_1{+}s_2$) \\
$\mathbf{r}$ & Ratio schedule $(r, r_1, r_2)$ \\
$\eta$, $\mathcal{B}$ & Learning rate, batch size \\
\hline
\multicolumn{2}{@{}l@{}}{\textit{Search-grid factors}} \\
\hline
$f_C$ & Compute factor: $C{=}2^{f_C}\!\cdot C_0$ \\
$f_M$ & Model-scale factor: $M{=}M_0/2^{f_M}$ \\
$f_k$ & Epoch factor: $k{=}2^{f_k}$ \\
$f_r$ & Language-ratio factor: $r{=}1/2^{f_r}$ \\
$f_D$ & Target-corpus factor: \\
 & $f_D{=}{-}f_r{+}f_M{-}f_k{+}f_C$, \; $D_T{=}2^{f_D}\!\cdot D_{T,0}$ \\
\hline
\multicolumn{2}{@{}l@{}}{\textit{Chinchilla base scaling law}} \\
\hline
$L_\text{base}(M,D)$ & $A/M^\alpha + B/D^\beta + E$ \\
$A$, $B$ & Coefficients of model-/data-size terms \\
$\alpha$, $\beta$ & Exponents of model/data size \\
$E$ & Irreducible loss \\
$G$ & $(\alpha A/\beta B)^{1/(\alpha+\beta)}$ \\
$D^*(C)$ & Compute-optimal corpus size, $G^{-1}C^{\alpha/(\alpha+\beta)}$ \\
$M^*(C)$ & Compute-optimal model scale, $G\cdot C^{\beta/(\alpha+\beta)}$ \\
\hline
\multicolumn{2}{@{}l@{}}{\textit{Effective sizes and saturation}} \\
\hline
$D'$, $M'$ & Effective data/model size \\
$\hat{M}'$ & Empirically estimated $M'$ (inverted from loss) \\
$U_M$ & Compute-optimal model size at $D_T$ \\
$R_D$ & Data repetition count ($k{-}1$) \\
$R_M$ & Excess model size ($M/U_M{-}1$) \\
$h(R;R^*)$ & Saturation function, $1{+}R^*(1{-}e^{-R/R^*})$ \\
$R_D^*$, $R_M^*$ & Saturation constants for $D'$, $M'$ \\
$R_M^*(k)$ & $k$-dep.\ form (mono): $a/(k{-}1)^b{+}c$ \\
$a$, $b$, $c$ & Parameters of $R_M^*(k)$ \\
$R_{D,\text{high}}^*$ & Saturation const.\ for $D_\text{high}$ in $g$ \\
$h_{Lp}$ & Sharper saturation variant (Appendix) \\
$p$ & Sharpness parameter in $h_{Lp}$ \\
\hline
\end{tabular}
\end{minipage}%
\hfill
\begin{minipage}[t]{0.68\textwidth}
\begin{tabular}{@{}ll@{}}
\hline
\multicolumn{2}{@{}l@{}}{\textit{Language-ratio factor}} \\
\hline
$\mathcal{R}(r)$ & He ratio factor, $r^{-\gamma}$ \\
$\mathcal{R}_\text{dual}(r,r_f)$ & Dual power law (single-/multi-stage) \\
$\gamma$ & Exponent for language ratio (single-stage) \\
$\gamma_2$ & Exponent for $r/r_f$ (multi-stage) \\
$\tilde\gamma_2(r_1)$ & $r_1$-dependent extension (Appendix) \\
$\dot\gamma_2$, $r_0$ & Slope and offset in $\tilde\gamma_2$ \\
\hline
\multicolumn{2}{@{}l@{}}{\textit{High-resource token weight}} \\
\hline
$g$ & Weight on $D_\text{high}$ in $D'$ \\
$g(r,R_D)$ & Functional form of $g$: \\
 & $(1{-}r)^\psi + [1{-}(1{-}r)^\psi]\exp(-R_D/R_{D,\text{high}}^*)$ \\
$\psi$ & Exponent controlling lower bound of $g$ \\
$\tau$ & Constant weight in ATLAS baseline \\
\hline
\multicolumn{2}{@{}l@{}}{\textit{Loss-prediction models}} \\
\hline
$L$ & Target-language validation loss \\
$L_\text{He}$, $L_\text{Mu}$ & He / Muennighoff scaling laws \\
$L_{M^3}$ & Proposed $M^3$ Scaling Law \\
$L_\text{actual}$ & Observed validation loss \\
$L^*(C,D_T,k)$ & Min.\ loss over remaining hyperparams \\
\hline
\multicolumn{2}{@{}l@{}}{\textit{Optimization targets}} \\
\hline
$k^*$, $k^*_\mathbf{r}$ & Optimal epochs; at fixed schedule $\mathbf{r}$ \\
$K_\mathbf{r}$ & Collapse function: $k^*_\mathbf{r}{=}K_\mathbf{r}(D_T/D^*)$ \\
$r^*$, $M^*$ & Optimal ratio; optimal model scale \\
$x$ & Scarcity variable, $D_T/C^{\alpha/(\alpha+\beta)}$ \\
$\Phi$, $Q_D$, $Q_U$, $Q_M$ & Helpers in optimal-$k$ derivation \\
\hline
\multicolumn{2}{@{}l@{}}{\textit{Model fitting and evaluation}} \\
\hline
$\theta$, $s_i$ & Scaling-law params; training setting of run $i$ \\
$L_i$, $\hat{L}_i$ & Observed / predicted loss for run $i$ \\
$\mathcal{T}$, $\bar{L}_\mathcal{T}$ & Test set; test-set mean loss \\
$R^2$ & Coefficient of determination \\
$\mathrm{Huber}_\delta$ & Huber loss ($\delta{=}0.001$) \\
$p_0, p_1, p_2$ & Quadratic-fit coefficients for $k^*$ \\
$\rho$ & Pearson correlation (downstream) \\
\hline
\multicolumn{2}{@{}l@{}}{\textit{Baseline-only symbols}} \\
\hline
\multicolumn{2}{@{}l@{}}{\footnotesize Sedova: $E_s, C_s, B_s, \alpha_s, \beta_s, \delta_s, \gamma_s, \tau_s, R_{D,s}^*, D'_\text{Se}$} \\
\multicolumn{2}{@{}l@{}}{\footnotesize D-CPT: $C_c, \nu$} \\
\multicolumn{2}{@{}l@{}}{\footnotesize PTPP: $\mathrm{PTPP}{=}D_1/N$;\; $F, \xi$ (floor coeff./exp.);\; $\lambda, \zeta, \beta_\text{eff}$} \\
\multicolumn{2}{@{}l@{}}{\footnotesize Zhang: $\phi_1, \phi_2$ (per-stage data exponents)} \\
\hline
\multicolumn{2}{@{}l@{}}{\textit{Optimizer (Appendix)}} \\
\hline
$\beta_1, \beta_2, \epsilon$ & Adam parameters \\
\hline
\end{tabular}
\end{minipage}
}
\caption{Summary of notation.}
\label{tab:notation}
\end{table*}

\section{Implementation Details of LLM Pretraining and Evaluation}
\label{subsec_appendix_pretrain}

\begin{table}[!htb]
\centering
\begin{tabular}{ccccc}
\hline
$f_M$ & $n_\text{layers}$ & $n_\text{heads}$ & $d_\text{model}$ & $M$ \\
\hline
5 & 2 & 4 & 128 & $1.49 \times 10^7$ \\
4 & 4 & 4 & 128 & $2.99 \times 10^7$ \\
3 & 4 & 7 & 224 & $5.85 \times 10^7$ \\
2 & 4 & 12 & 384 & $1.18 \times 10^8$ \\
1 & 8 & 12 & 384 & $2.36 \times 10^8$ \\
0 & 8 & 39 & 624 & $4.70 \times 10^8$ \\
-1 & 16 & 39 & 624 & $9.39 \times 10^8$ \\
\hline
\end{tabular}
\caption{The number of layers $n_\text{layers}$, the number of heads $n_\text{heads}$, and the dimension of hidden states $d_\text{model}$ for each model scale factor $f_M$.}
\label{tab:model_shape}
\end{table}

\begin{algorithm}[h]
\caption{Determine Batch Size}
\label{alg:determine_batchsize}
\begin{algorithmic}[1]\small
\Procedure{DetermineBatchSize}{$C$, $d_{model}$, $n_{layer}$}
    \State $\texttt{seq\_length} \gets 4096$
    \State $\texttt{complexity} \gets n_{layer} \cdot d_{model}^2$

    \If{$\texttt{complexity} < 0.1 \times 10^8$}
        \State $\texttt{local\_batch\_size} \gets 4$
    \ElsIf{$\texttt{complexity} < 0.5 \times 10^8$}
        \State $\texttt{local\_batch\_size} \gets 2$
    \ElsIf{$\texttt{complexity} < 1.1 \times 10^8$}
        \State $\texttt{local\_batch\_size} \gets 1$
    \EndIf

    \State $\texttt{optimal\_local\_batch\_size} \gets \lfloor 0.292 \cdot C^{0.3271} / (\texttt{seq\_length} \cdot 8) \rceil$

    \If{$\texttt{optimal\_local\_batch\_size} < \texttt{local\_batch\_size}$}
        \State $\texttt{local\_batch\_size} \gets \texttt{optimal\_local\_batch\_size}$
        \State $\texttt{accumulation\_step} \gets 1$
    \Else
        \State $\texttt{accumulation\_step} \gets \lfloor \texttt{optimal\_local\_batch\_size} / \texttt{local\_batch\_size} \rceil$
    \EndIf

    \State $\texttt{batch\_size} \gets \texttt{local\_batch\_size} \cdot 8 \cdot \texttt{accumulation\_step}$

    \Return $\texttt{batch\_size}$
\EndProcedure
\end{algorithmic}
\end{algorithm}

\subsection{Model Pretraining}

Throughout our experiments, we adopted the training configuration of \citet{bi2024deepseek}.
More specifically, we trained decoder-only language models with LLaMA architecture \cite{touvron2023llama} and set context length $S=4096$.
We measured the model scale by non-embedding FLOPs/token, $M=72 n{d_\text{model}}^2+12n{d_\text{model}}S$, where $n$, $d_\text{model}$, and $S$ are the number of layers, the size of hidden states, and the sequence length respectively.
To improve the efficiency of our experiments by reusing intermediate checkpoints, we used their multi-step learning rate scheduler, in which we decreased the learning rate to 31.6\% and 10\% after processing 80\% and 90\% of the total steps of each training stage, respectively.
For learning rate $\eta$ and batch size $\mathcal{B}$, we adopted their power-law relationships, $\eta=0.3118\cdot C^{-0.1250}$ and $\mathcal{B}=0.2920\cdot C^{0.3271}$, which were fitted through extensive experiments across different compute budgets $C$ to achieve near-optimal performance in \cite{bi2024deepseek}.\footnote{We added minor adjustment to $\mathcal{B}$ to improve training efficiency in some cases, which is detailed in Appendix \ref{subsec_appendix_pretrain}.}
Following findings by \citet{ibrahim2024simple} that the learning rate of continual pretraining should be re-warmed to that of pretraining, we set the same maximum learning rate for both the first and second stages for two-stage training.

We used the Transformers library \cite{wolf-etal-2020-transformers} (ver. 4.37.2, Apache-2.0 license) in our experiments.
We used \texttt{LlamaForCausalLM} as the model architecture, and set the standard deviation for parameter initialization to 0.006.
Configurations of the model shape is shown in Table \ref{tab:model_shape}.
Based on the analysis by \citet{kaplan2020scaling}, we only used those model shapes with aspect ratio $\frac{d_\text{model}}{n_\text{layers}}$ between 30 and 150.
We used Adam optimizer \cite{kingma2014adam} and set its parameter to $(\beta_1, \beta_2, \epsilon) = (0.9, 0.95, 10^{-8})$.
We set the number of warmup steps to 500, the weight decay to 0.1 and gradient clipping at a norm of 1.
We used \texttt{Trainer} class to train our models, with all unspecified parameters left at their default values.
To improve efficiency of training, we used the batch size adjusted by Algorithm \ref{alg:determine_batchsize}.
We used a server with eight A100 80GB GPUs to train our models.

For multi-epoch training, the order of the repeated data was randomized with different seeds for each epoch.

\subsection{Corpus and Tokenizer}

We retrieved C4 \cite{raffel2020exploring} and mC4 \cite{xue-etal-2021-mt5} from \url{https://huggingface.co/datasets/allenai/c4}.
We only used \texttt{ja}, \texttt{id}, and \texttt{sw} subsets of mC4 and did not use \texttt{ja-Latn} subsets.
To reduce the cost of the evaluation, we respectively subsampled 5\%, and 6.6\% of the validation corpora for Japanese, and Indonesian, while using the full validation corpus for Swahili.

We tokenized the corpora with BLOOM tokenizer.
BLOOM tokenizer is available at \url{https://huggingface.co/bigscience/tokenizer} under the BigScience RAIL License v1.0.
Note that English, Indonesian, and Swahili were used to train the BLOOM tokenizer, while Japanese was not used to train it.

\subsection{Training Hyperparameters}
\label{subsec_hyperparam_one_stage}

Based on the scaling law of the optimal model scale and data allocation by \citet{bi2024deepseek}, we set the reference compute budget $C_0$, target language corpus budget $D_{T,0}$, and model scale $M_0$ as follows:
\begin{equation}
  \begin{aligned}
      C_0 &= 10^{18}, \\
      D_{T,0} &= 5.8316\cdot C_0^{0.4757}, \\
      M_0 &= \frac{C_0}{D_{T,0}}.
  \end{aligned}
  \label{eq:ref_values}
\end{equation}

\begin{table}[!htb]
  \centering
  \begin{tabular}{c||c|c|c||c}
    \hline
    $f_C$  & $f_r$ & $f_M$ & $f_k$ & $f_D$ \\
    \hline
    \hline

    0  & \multirow{5}{*}{[0,4)} & [-1,5) & \multirow{5}{*}{[0,10)} & [-5,2) \\
    \cline{1-1} \cline{5-5}

    -1  &  & \multirow{2}{*}{[0,5)} & & \multirow{2}{*}{[-6,1)} \\
    \cline{1-1}

    -2  &  &  &  & \\
    \cline{1-1} \cline{5-5}

    -3  &  & \multirow{2}{*}{[1,6)} & & \multirow{2}{*}{[-7,0)} \\
    \cline{1-1}

    -4 & &  & &  \\
    \hline
  \end{tabular}
  \caption{Search space of factors $(f_r,f_M,f_k)$ for each compute factor $f_C$. We grid-searched a set of factors, $(f_r,f_M,f_k,f_C)$, within those ranges and selected it for the experiment if its corresponding $f_D=-f_r+f_M-f_k+f_C$ is within the specified range. For the (Swahili, English) pair, we limit the search to $f_D \leq -3$ due to limited corpus availability.}
  \label{tab:search_space_factors}
\end{table}

\textbf{Single-stage Training:} The hyperparameters of single-stage training are the constant language ratio $r$, the model scale $M$, and the number of target language epochs $k$.

We parametrized each single-stage training setup by four integer factors, $(f_r, f_M, f_k, f_C)$, where $f_*\in\mathbb{Z}$, and set each of its hyperparameters and compute $C$ and $D_T$ as follows:
\begin{equation}
  \begin{array}{cc}
      r=\frac{1}{2^{f_r}}, & M=\frac{M_0}{2^{f_M}}, \\
      k=2^{f_k}, & C=2^{f_C}\cdot C_0, \\
      \multicolumn{2}{c}{D_T = 2^{-f_r+f_M-f_k+f_C}\cdot D_{T,0}}.
  \end{array}
  \label{eq:parametrization_hyperparameter}
\end{equation}
The training setup involved multi-epoch training when $f_k>0$ and multi-lingual training when $f_r>0$.
Multiple training setups with the same $(C,D_T)$ were created by keeping $f_C$ and $f_D=-f_r+f_M-f_k+f_C$ constant.
We list the search spaces for those factors in Table \ref{tab:search_space_factors}.

\textbf{Two-stage Training:} For each stage of two-stage training, we used a different constant language ratio, $r_1$ (for the first stage) and $r_2$ (for the second stage), where $r_1<r_2$.
Using the proportions of the first and second stage training, $s_1$ and $s_2$, relative to the total length of the training, we define the average language ratio as $r=s_1r_1+s_2r_2$ by slightly abusing notation.
Instead of explicitly setting $s_1$ and $s_2$, we used $r$ to compute $s_1=\frac{r_2-r}{r_2-r_1}$ and $s_2=1-s_1$.

We parametrized each two-stage training setup by $(r_1, r_2, f_r, f_M, f_k, f_C)$, and used Eq.~(\ref{eq:parametrization_hyperparameter}) in \S\ref{subsec_hyperparam_one_stage} to set $r$, $M$, and $k$, and compute $C$ and $D_T$.
We grid-searched over $r_1\in \{ 1/2, 1/4, 1/8, 1/16, 1/32, 0 \}$, $r_2\in\{ 1/4, 1/2, 3/4, 1 \}$, and the same search space of other factors as shown in Table \ref{tab:search_space_factors}.

\subsection{Details of Three-Stage Training Experiments}
\label{subsec:3stage_details}

In three-stage training experiments, we trained models with randomly sampled three-stage training setups $(r_3, r, M, D_T)$ on the (Japanese, English) pair under single-epoch training ($k=1$), spanning $C\in\{C_0/16,\ldots, C_0\}$, average ratios $r \in \{1/2,1/4,1/8\}$, and final-stage ratios $r_3 \in \{0.25,0.5, 0.75,1.0\}$ ($r_3>r$) with $M$ and $D_T$ drawn from the same search space as the two-stage experiments.
To isolate the effects of the intermediate ratios, for each $(r_3, r, M, D_T)$ we varied $r_1$, $r_2$, and their average $r_{12}$ and observed the resulting performance changes.
Specifically, we sampled $r_{12}<r$ from $\{1/4,1/8,1/16,1/32\}$, and for each $(r_3,r,r_{12})$ compared three representative $(r_1,r_2)$ patterns:
\textbf{(A)} the closest valid $(r_1,r_2)$ bracketing $r_{12}$, approximating a near-constant two-stage schedule;
\textbf{(B)} $r_1=0$ with $r_2$ set to the smallest valid candidate above $r_{12}$;
and
\textbf{(C)} $r_1=0$ with $r_2$ set to the largest valid candidate between $r_{12}$ and $r_3$.

The stage proportions $s_1, s_2, s_3$ for three-stage training were determined hierarchically, extending the inverse-computation approach used for two-stage training (\S\ref{subsec:training_configuration}).
We first split the total training into a combined first-and-second stage block and the third stage using the outer constraint $r = s_{12} r_{12} + s_3 r_3$:
\begin{equation}
  s_{12} = \frac{r_3 - r}{r_3 - r_{12}}, \quad s_3 = 1 - s_{12},
\end{equation}
where $r_{12}$ is the average language ratio over stages 1 and 2.
We then split $s_{12}$ into stages 1 and 2 using the inner constraint $r_{12} = (s_1 / s_{12}) r_1 + (s_2 / s_{12}) r_2$:
\begin{equation}
  s_1 = s_{12} \cdot \frac{r_2 - r_{12}}{r_2 - r_1}, \quad s_2 = s_{12} - s_1.
\end{equation}

For each combination of $(r_3, r_{12})$, we collected the set of $(f_M, C)$ pairs that were feasible for all three patterns (A, B, and C) described in \S\ref{subsubsec:dual_power_law_two_stage}, pooled across all $C$ values, and randomly sampled up to 10 pairs (with a fixed random seed).
Pattern C was omitted when $r_2$ coincided with that of pattern B, since the two patterns become identical in that case.
In total, we conducted 307 three-stage training runs (pattern A: 117, B: 125, C: 65).

\section{Implementation Details of Scaling Law Analysis}

\subsection{Scaling Law Model Fitting}

In this paper, following \citet{hoffmann2022training}, we fit a scaling law model $L(s_i; \theta)$ parameterized by $\theta$ to the observed validation loss $L_i$ of an LLM trained under setting $s_i$ (e.g., data size $D_T$, model size $M$, language ratio $r$, etc.), by minimizing the Huber loss \cite{huber1964robust} using the L-BFGS algorithm \cite{nocedal1980updating}:
\begin{equation}
  \min_\theta \sum_i \text{Huber}_\delta (\log L(s_i; \theta) - \log L_i),
\end{equation}
where we used $\delta=0.001$.
The parameter bounds during optimization were set as shown in Table \ref{tab:param_bounds}.
To mitigate sensitivity to initial values, we randomly sampled initial values of $\theta_i$ within the ranges shown in Table \ref{tab:param_bounds} (uniformly in log scale for $A$, $B$, and $r_0$, and uniformly in linear scale for other parameters), independently ran the optimization 50 times, and adopted the best fit result.

\begin{table*}[!htb]
  \centering
  \scalebox{0.8}{
  \begin{tabular}{llcc}
    \hline
    Symbol & Description & Bounds & Initial Sampling Range \\
    \hline
    \multicolumn{4}{c}{\textit{Parameters of Chinchilla Model \cite{hoffmann2022training}}} \\
    \hline
    $A$$^\dagger$ & Coefficient for model size term & $[10^{-6}, 10^{6}]$ & $[10^{-2}, 10^{4}]$ \\
    $B$$^\dagger$ & Coefficient for data size term & $[10^{-6}, 10^{6}]$ & $[10^{-2}, 10^{4}]$ \\
    $\alpha$ & Exponent for model size & $[0.1, 2.0]$ & $[0.1, 0.8]$ \\
    $\beta$ & Exponent for data size & $[0.01, 5.0]$ & $[0.1, 0.8]$ \\
    $E$ & Irreducible loss & $[0.001, 10.0]$ & $[1.0, 5.0]$ \\
    \hline
    \multicolumn{4}{c}{\textit{Parameters of Muennighoff Model (Multi-epoch) \cite{muennighoff2024scaling}}} \\
    \hline
    $R_D^*$ & Saturation constant for effective data & $[0.1, 200.0]$ & $[1.0, 100.0]$ \\
    $R_M^*$ & Saturation constant for effective model size & $[0.1, 100.0]$ & $[0.5, 50.0]$ \\
    $R_{D,\text{high}}^*$ & Saturation constant for high-resource language tokens in $g$ (Eq.~\ref{eq:g_func}) & $[0.1, 200.0]$ & $[1.0, 100.0]$ \\
    $\psi$ & Exponent for lower bound of $g$ (Eq.~\ref{eq:g_func}) & $[0.01, 10.0]$ & $[0.1, 5.0]$ \\
    \hline
    \multicolumn{4}{c}{\textit{Parameters of He Model (Multi-lingual) \cite{he2024scaling}}} \\
    \hline
    $\gamma$ & Exponent for language ratio & $[0.001, 1.0]$ & $[0.01, 0.5]$ \\
    \hline
    \multicolumn{4}{c}{\textit{Parameters of Searched Two-stage Models}} \\
    \hline
    $\gamma_2$ & Exponent for language ratio in stage 2 & $[0.001, 1.0]$ & $[0.01, 0.5]$ \\
    $r_0$$^\dagger$ & Offset for language ratio (effective language ratio at $r=0$) & $[10^{-6}, 1.0]$ & $[10^{-4}, 0.1]$ \\
    $\dot{\gamma}_2$ & Slope of $r_1$-dependent $\gamma_2$ (Eq.~\ref{eq:gamma2_r1}) & $[-0.5, 0.5]$ & $[-0.2, 0.2]$ \\
    \hline
  \end{tabular}
  }
  \caption{Parameter bounds and initial value sampling ranges used in scaling law model fitting. Bounds define the hard constraints during L-BFGS optimization, while sampling ranges define the narrower region from which initial values are sampled. $^\dagger$Sampled uniformly in log scale while other parameters are sampled uniformly in linear scale.}
  \label{tab:param_bounds}
\end{table*}

\subsection{Scaling Law Model Evaluation}
\label{subsec:scaling_law_evaluation}

The extrapolation performance of scaling law models can vary depending on the choice of train-test split.
To avoid relying on a single arbitrary split, we evaluate extrapolation performance along multiple axes of the training setups.
Specifically, for each axis, we define a threshold that partitions the data into a training set and a held-out test set.
Each model is fit on the training set and evaluated by the coefficient of determination ($R^2$) on the test set, defined as
\begin{equation}
  R^2 = 1 - \frac{\sum_{i \in \mathcal{T}} (L_i - \hat{L}_i)^2}{\sum_{i \in \mathcal{T}} (L_i - \bar{L}_{\mathcal{T}})^2},
  \label{eq:r_squared}
\end{equation}
where $\mathcal{T}$ is the test set, $L_i$ and $\hat{L}_i$ are the observed and predicted validation losses, and $\bar{L}_{\mathcal{T}} = \frac{1}{|\mathcal{T}|}\sum_{i \in \mathcal{T}} L_i$ is the test set mean.
Note that $R^2$ is computed in the original (non-log) scale and can be negative when predictions are worse than predicting the test set mean.
We require both the training and test sets to contain at least 10 data points, and splits that do not meet this criterion are skipped.

\subsubsection{Held-Out Splits for Combined Evaluation}
\label{subsubsec:splits_all}

For the main evaluation on all single-stage and two-stage data, we define 18 train-test splits along six axes of the training setups (Table~\ref{tab:splits_all}).
The test set in each split contains data from regions that are of practical interest for extrapolation: larger compute budgets, larger models, larger corpora, lower or higher target language ratios, and higher epoch counts.
To avoid dependence on any single threshold, we use multiple thresholds per axis.
Note that for Swahili, the available target language corpus is limited to $f_D \leq -3$, so no data points fall into the test set of the three $D_T$-axis splits (DT\_ge0, DT\_ge$-$1, DT\_ge$-$2); these splits are therefore omitted for Swahili.

\begin{table}[!htb]
  \centering
  \scalebox{0.6}{
  \begin{tabular}{lllp{3.2cm}}
    \hline
    Axis & Split name & Test condition & Extrapolation direction \\
    \hline
    \multirow{2}{*}{$C$ (compute budget)}
      & C\_ge0  & $f_C \geq 0$  & \multirow{2}{3.2cm}{Larger compute} \\
      & C\_ge-1 & $f_C \geq -1$ & \\
    \hline
    \multirow{3}{*}{$M$ (model scale)}
      & M\_le-1 & $f_M \leq -1$ & \multirow{3}{3.2cm}{Larger model} \\
      & M\_le0  & $f_M \leq 0$  & \\
      & M\_le1  & $f_M \leq 1$  & \\
    \hline
    \multirow{3}{*}{$D_T$ (target corpus)}
      & DT\_ge0  & $f_D \geq 0$  & \multirow{3}{3.2cm}{Larger target\newline tokens} \\
      & DT\_ge-1 & $f_D \geq -1$ & \\
      & DT\_ge-2 & $f_D \geq -2$ & \\
    \hline
    \multirow{3}{*}{$D$ (total tokens)}
      & D\_ge34 & $\log_2 D \geq 34$ & \multirow{3}{3.2cm}{Larger total\newline tokens} \\
      & D\_ge33 & $\log_2 D \geq 33$ & \\
      & D\_ge32 & $\log_2 D \geq 32$ & \\
    \hline
    \multirow{2}{*}{}
      & r\_le0.125 & $r \leq 0.125$ & \multirow{2}{3.2cm}{Lower target\newline language ratio} \\
      & r\_le0.25  & $r \leq 0.25$  & \\
    \cline{2-4}
    \multirow{2}{*}{$r$ (language ratio)}
      & r\_ge0.5   & $r \geq 0.5$   & \multirow{2}{3.2cm}{Higher target\newline language ratio} \\
      & r\_ge1     & $r \geq 1.0$   & \\
    \hline
    \multirow{3}{*}{$k$ (epochs)}
      & k\_ge32  & $k \geq 32$  & \multirow{3}{3.2cm}{Higher epoch} \\
      & k\_ge64  & $k \geq 64$  & \\
      & k\_ge128 & $k \geq 128$ & \\
    \hline
  \end{tabular}
  }
  \caption{Train-test split definitions for the combined (1-stage + 2-stage) evaluation.
  Data points satisfying the test condition form the test set; the remainder forms the training set.}
  \label{tab:splits_all}
\end{table}

\subsubsection{Held-Out Splits for Two-Stage Evaluation}
\label{subsubsec:splits_2stage}

For multi-lingual single-epoch evaluation, of the 18 splits in Table~\ref{tab:splits_all}, we use 13 for this evaluation (Table~\ref{tab:splits_2stage}).
From the $r$-axis, only r\_le0.125 and r\_ge0.5 are retained; r\_le0.25 is excluded because the two-stage data has only $r \in \{0.125, 0.25, 0.5\}$, and training set with $r > 0.25$ would contain only a single $r$ level ($r = 0.5$), preventing continual-pretraining baselines trained only on two-stage data from identifying the effect of $r$.
Similarly, r\_ge1 is excluded because no monolingual ($r = 1$) data exists in the two-stage experiments.
All $k$-axis splits are omitted since this evaluation is restricted to $k = 1$.

\begin{table}[!htb]
  \centering
  \scalebox{0.85}{
  \begin{tabular}{ll}
    \hline
    Axis & Splits used \\
    \hline
    $C$ (compute budget) & C\_ge0, C\_ge-1 \\
    $M$ (model scale) & M\_le-1, M\_le0, M\_le1 \\
    $D_T$ (target corpus) & DT\_ge0, DT\_ge-1, DT\_ge-2 \\
    $D$ (total tokens) & D\_ge34, D\_ge33, D\_ge32 \\
    $r$ (language ratio) & r\_le0.125, r\_ge0.5 \\
    \hline
  \end{tabular}
  }
  \caption{Train-test split definitions for the multi-lingual single-epoch evaluation (13 splits).
  Two $r$-axis splits and all $k$-axis splits from Table~\ref{tab:splits_all} are excluded; see text for rationale.}
  \label{tab:splits_2stage}
\end{table}

\subsubsection{Aggregation of Test $R^2$ Across Splits and Languages}
\label{subsubsec:aggregation}

For each axis (e.g., $C$, $M$), the per-axis $R^2$ reported in Table~\ref{tab:baselines_extended} is computed in two steps.
First, for each language, we average the test $R^2$ over all held-out splits belonging to that axis (e.g., for the $C$ axis, over C\_ge0 and C\_ge$-$1).
If a split is skipped for a particular language (e.g., because the training or test set contains fewer than 10 data points) or yields an invalid $R^2$ (e.g., due to numerical issues in fitting), it is excluded from that language's average.
To ensure a fair comparison, whenever a split is excluded for any model, we exclude it for \emph{all} models in the same evaluation setting, so that all models are compared on the same set of splits.
Second, we take the unweighted mean of these per-language averages across all languages.
The \textit{Avg} column is the unweighted mean of the per-axis values.

\subsubsection{Two-Phase Fitting}
\label{subsubsec:twophase_fitting}

In preliminary experiments, we found that fitting the Chinchilla base parameters ($A$, $B$, $\alpha$, $\beta$, $E$) jointly with the remaining parameters causes the base parameters to drift, substantially degrading extrapolation performance along the $M$- and $D$-axis splits.
To address this, we adopt a two-phase fitting procedure for all models that build on the Chinchilla formulation:
\begin{enumerate}
  \item \textbf{Phase 1.} Fit the Chinchilla base parameters ($A$, $B$, $\alpha$, $\beta$, $E$) using only monolingual ($r = 1$), low-epoch ($k \leq 4$) single-stage data from the training set.
  \item \textbf{Phase 2.} Fix the Phase~1 parameters and optimize the remaining parameters (e.g., $R_D^*$, $R_M^*$, $\gamma$, $\gamma_2$) on the full training set.
\end{enumerate}

\subsection{Implementation of Baseline Models}
\label{subsec:implementation_baseline_models}

This section summarizes the scaling law models used as baselines in Table~\ref{tab:baselines_extended}.
All models share the Chinchilla base $L_\text{base}(M,D) = A/M^\alpha + B/D^\beta + E$ (Eq.~\ref{eq:chinchilla}) and the saturation function $h(R;\,R^*) = 1 + R^*\bigl(1 - \exp(-R/R^*)\bigr)$ (Eq.~\ref{eq:effective_M_muennighoff}).

\paragraph{Chinchilla \cite{hoffmann2022training} (5P).}
$L = L_\text{base}(M,D)$.

\paragraph{He \cite{he2024scaling} (6P).}
$L = L_\text{base}(M,D) \cdot r^{-\gamma}$.

\paragraph{He + Dual PL (7P).}
$L = L_\text{base}(M,D) \cdot \mathcal{R}_\text{dual}(r,r_f)$, where $\mathcal{R}_\text{dual}$ is defined in Eq.~(\ref{eq:dual_power_law}).

\paragraph{Muennighoff \cite{muennighoff2024scaling} (7P).}
$L = L_\text{base}(M',D')$ with effective sizes $D'$ and $M'$ as defined in Eqs.~(\ref{eq:effective_D_muennighoff})--(\ref{eq:effective_M_muennighoff}).

\paragraph{ATLAS \cite{longpre2025atlas} (7P).}
$L = L_\text{base}(M, D')$, where $D'$ is defined in Eq.~(\ref{eq:effective_D_longpre}) with constant weight $g = \tau$.
Unlike the Muennighoff model, ATLAS does not use the effective model size $M'$, applying $M$ directly.

\paragraph{Sedova \cite{sedova2026scalinglawsmixturepretraining} (9P).}
\begin{equation*}
  L = E_s + \frac{C_s}{M^{\beta_s}} + \frac{B_s M^{\delta_s}}{{D'_\text{Se}}^{\alpha_s}} + \gamma_s \cdot r,
\end{equation*}
where $D'_\text{Se} = D_\text{high} + \tau_s \cdot D_T \cdot h(R_D;\,R_{D,s}^*)$.
Since this model does not contain the Chinchilla model as a special case, all 9 parameters are fitted jointly in a single phase, unlike the two-phase procedure used for the other models (\S\ref{subsec:scaling_law_evaluation}).

The following four two-stage-only baselines also do not contain the Chinchilla model as a substructure and are fitted in a single phase.

\paragraph{D-CPT Law \cite{que2024d} (8P).}
\begin{equation*}
  L = E + \frac{A}{M^\alpha} + \frac{B \cdot r_f^{\nu}}{D_2^\beta} + \frac{C_c}{r_f^\gamma},
\end{equation*}
where $D_2 = s_2 D$ is the number of second-stage tokens.

\paragraph{PTPP-F1 \cite{goffinet2025ptpp} (10P).}
\begin{equation*}
  L = E + \frac{A}{M^\alpha} + \frac{B \cdot r_f^{\nu}}{D_2^\beta} + \frac{C_c}{r_f^\gamma} + \frac{F}{\mathrm{PTPP}^\xi},
\end{equation*}
where $\mathrm{PTPP} = D_1 / N$ is the pretraining tokens per parameter, $D_1 = s_1 D$ is the number of first-stage tokens, and $N$ is the number of model parameters.

\paragraph{PTPP-F2 \cite{goffinet2025ptpp} (10P).}
Same as D-CPT Law, but with data exponent modulated by PTPP:
\begin{equation*}
  \beta_\text{eff} = \beta \left(1 - \frac{\lambda \cdot \mathrm{PTPP}^\zeta}{1 + \mathrm{PTPP}^\zeta}\right),
\end{equation*}
and $D_2^\beta$ in the D-CPT formula is replaced by $D_2^{\beta_\text{eff}}$.

\paragraph{PTPP-F3 \cite{goffinet2025ptpp} (12P).}
Same as PTPP-F1, but with data exponent modulated by PTPP as in PTPP-F2:
\begin{equation*}
  L = E + \frac{A}{M^\alpha} + \frac{B \cdot r_f^{\nu}}{D_2^{\beta_\text{eff}}} + \frac{C_c}{r_f^\gamma} + \frac{F}{\mathrm{PTPP}^\xi},
\end{equation*}
where $\beta_\text{eff} = \beta \bigl(1 - \frac{\lambda \cdot \mathrm{PTPP}^\zeta}{1 + \mathrm{PTPP}^\zeta}\bigr)$, combining the gated data exponent of PTPP-F2 with the additive floor of PTPP-F1.

\paragraph{Zhang \cite{zhang2024scaling} (6P).}
\citet{zhang2024scaling} propose a multiplicative joint scaling law over $D_2$ and one other variable ($D_1$ or $M$).
Here we evaluate a na\"ive extension to four variables ($M$, $D_1$, $D_2$, $r_f$):
\begin{equation*}
  L = \frac{A}{M^\alpha \cdot D_1^{\phi_1} \cdot D_2^{\phi_2} \cdot r_f^\gamma} + E.
\end{equation*}

\section{Additional Scaling Law Analysis}
\label{subsec:scaling_law_full_results}

\subsection{Per-Language, Per-Split Evaluation Results}
\label{subsec:subset_evaluation}

Tables~\ref{tab:full_splits_a}--\ref{tab:full_splits_d2} provide the per-language, per-split test $R^2$ values underlying the aggregated results in Table~\ref{tab:baselines_extended}.
Splits where the training or test set contains fewer than 10 data points are marked ``---''.
For Swahili, the $D_T$-axis splits (DT\_ge0, DT\_ge$-$1, DT\_ge$-$2) are omitted due to limited corpus availability.

\begin{table*}[!htb]
  \centering
  \scalebox{0.65}{
  \setlength{\tabcolsep}{2.5pt}
  \begin{tabular}{@{}ll|rr|rrr|rrr|rrr|rrrr|rrr@{}}
    \hline
    & & \multicolumn{2}{c|}{$C$} & \multicolumn{3}{c|}{$M$} & \multicolumn{3}{c|}{$D_T$} & \multicolumn{3}{c|}{$D$} & \multicolumn{4}{c|}{$r$} & \multicolumn{3}{c}{$k$} \\
    Model & Lang & \rotatebox{70}{C\_ge0} & \rotatebox{70}{C\_ge$-$1} & \rotatebox{70}{M\_le$-$1} & \rotatebox{70}{M\_le0} & \rotatebox{70}{M\_le1} & \rotatebox{70}{DT\_ge0} & \rotatebox{70}{DT\_ge$-$1} & \rotatebox{70}{DT\_ge$-$2} & \rotatebox{70}{D\_ge34} & \rotatebox{70}{D\_ge33} & \rotatebox{70}{D\_ge32} & \rotatebox{70}{r\_le.125} & \rotatebox{70}{r\_le.25} & \rotatebox{70}{r\_ge.5} & \rotatebox{70}{r\_ge1} & \rotatebox{70}{k\_ge32} & \rotatebox{70}{k\_ge64} & \rotatebox{70}{k\_ge128} \\
    \hline
    \hline
    Chinchilla & ja & .39 & .46 & $-$.03 & $-$.33 & $-$.08 & .55 & .58 & .54 & $-$.38 & .39 & .31 & .24 & .30 & .44$^*$ & .41$^*$ & .20 & .03 & $-$.10 \\
    Chinchilla & id & .40 & .34 & $-$.26 & $-$.76 & $-$.29 & .62 & .62 & .55 & $-$.33 & .31 & .05 & .39 & .40 & .33$^*$ & .10$^*$ & $-$.17 & $-$.27 & $-$.30 \\
    Chinchilla & sw & .51 & .48 & .01 & $-$.37 & $-$.12$^*$ & --- & --- & --- & $-$.36 & .43 & .41 & .45 & .47 & .37$^*$ & .10$^*$ & .18 & .02 & $-$.06 \\
    \hline
    He & ja & .60 & .43 & .06 & $-$.17 & $-$.24 & .90 & .81 & .50 & $-$.03 & .49 & .65 & .54 & .58 & --- & --- & $-$.03 & $-$.31 & $-$.50 \\
    He & id & .47 & .39 & .08 & $-$.14 & $-$.45 & .85 & .85 & .81 & $-$.10 & .43 & .56 & .68 & .69 & --- & --- & $-$.30 & $-$.44 & $-$.48 \\
    He & sw & .35 & .10 & .64 & .45 & --- & --- & --- & --- & $-$9.45 & $-$2.23 & $-$.44 & .59 & $-$.67 & --- & --- & $-$.96 & $-$1.52 & $-$2.05 \\
    \hline
    Muennighoff & ja & .63 & .60 & .59 & .53 & .43 & .85 & .86 & .55 & .17 & .59 & .69 & $-$.75 & $-$.26 & --- & --- & .15 & .19 & .42 \\
    Muennighoff & id & .49 & .54 & $-$.13 & $-$.02 & .39 & .88 & .78 & .74 & .44 & .67 & .71 & $-$1.30 & $-$.62 & --- & --- & $-$.08 & $-$.10 & .14 \\
    Muennighoff & sw & .48 & .53 & $-$.23 & $-$.01 & --- & --- & --- & --- & $-$2.56 & $-$.06 & .50 & .51 & .60 & --- & --- & .42 & .41 & .46 \\
    \hline
    ATLAS & ja & .57 & .48 & .75 & .55 & .38 & .73 & .83 & .44 & $-$.99 & .18 & .53 & .45 & .56 & --- & --- & .60 & .63 & .74 \\
    ATLAS & id & .62 & .61 & .79 & .60 & .18 & .81 & .70 & .71 & $-$.52 & .35 & .64 & .59 & .64 & --- & --- & .42 & .45 & .55 \\
    ATLAS & sw & .49 & .46 & .75 & .73 & --- & --- & --- & --- & $-$7.11 & $-$1.15 & .11 & .60 & .66 & --- & --- & .41 & .40 & .51 \\
    \hline
    Sedova & ja & .84 & .79 & .79 & .44 & .50 & .84 & .85 & .83 & .25 & .71 & .79 & .75 & .76 & .63$^*$ & .58$^*$ & .63 & .52 & .43 \\
    Sedova & id & .76 & .69 & .77 & .22 & .28 & .79 & .79 & .68 & .09 & .64 & .73 & .75 & .71 & .57$^*$ & .34$^*$ & .39 & .30 & .27 \\
    Sedova & sw & .83 & .73 & .46 & .64 & .45$^*$ & --- & --- & --- & $-$.52 & .60 & .69 & .77 & .78 & .46$^*$ & .29$^*$ & .49 & .37 & .38 \\
    \hline
    $M^3$ & ja & .80 & .78 & .64 & .36 & .28 & .99 & .93 & .41 & .89 & .89 & .89 & .75 & .77 & --- & --- & .64 & .59 & .66 \\
    $M^3$ & id & .78 & .76 & .71 & .63 & .48 & .98 & .91 & .87 & .47 & .84 & .87 & .85 & .71 & --- & --- & .48 & .50 & .55 \\
    $M^3$ & sw & .83 & .80 & .73 & .80 & --- & --- & --- & --- & $-$1.25 & .39 & .73 & .84 & .61 & --- & --- & .54 & .38 & .33 \\
    \hline
    $M^3$ $-$DPL & ja & .76 & .66 & .61 & .32 & .13 & .91 & .87 & .35 & .51 & .72 & .81 & .63 & .61 & --- & --- & .61 & .63 & .66 \\
    $M^3$ $-$DPL & id & .76 & .71 & .69 & .61 & .46 & .90 & .85 & .81 & .11 & .70 & .82 & .69 & .51 & --- & --- & .51 & .52 & .56 \\
    $M^3$ $-$DPL & sw & .81 & .76 & .73 & .80 & --- & --- & --- & --- & $-$1.57 & .27 & .68 & .75 & .59 & --- & --- & .54 & .40 & .35 \\
    \hline
    $M^3$ $-g$ & ja & .79 & .77 & .69 & .46 & .30 & .99 & .92 & .56 & .67 & .80 & .83 & .73 & .76 & --- & --- & .59 & .58 & .66 \\
    $M^3$ $-g$ & id & .75 & .73 & .72 & .69 & .52 & .98 & .93 & .90 & .69 & .79 & .82 & .85 & .78 & --- & --- & .41 & .41 & .50 \\
    $M^3$ $-g$ & sw & .60 & .58 & .68 & .81 & --- & --- & --- & --- & $-$7.32 & $-$1.15 & .15 & .81 & .55 & --- & --- & .10 & $-$.08 & $-$.07 \\
    \hline
  \end{tabular}
  }
  \caption{Per-language, per-split test $R^2$ for \textbf{(a)}~All data (1-stage $+$ 2-stage), corresponding to Table~\ref{tab:baselines_extended}(a). Splits marked with $^*$ are excluded from the aggregation in Table~\ref{tab:baselines_extended} because at least one model failed to fit on that split.}
  \label{tab:full_splits_a}
\end{table*}

\begin{table*}[!htb]
  \centering
  \scalebox{0.65}{
  \setlength{\tabcolsep}{2.5pt}
  \begin{tabular}{@{}ll|rr|rrr|rrr|rrr|rrrr|rrr@{}}
    \hline
    & & \multicolumn{2}{c|}{$C$} & \multicolumn{3}{c|}{$M$} & \multicolumn{3}{c|}{$D_T$} & \multicolumn{3}{c|}{$D$} & \multicolumn{4}{c|}{$r$} & \multicolumn{3}{c}{$k$} \\
    Model & Lang & \rotatebox{70}{C\_ge0} & \rotatebox{70}{C\_ge$-$1} & \rotatebox{70}{M\_le$-$1} & \rotatebox{70}{M\_le0} & \rotatebox{70}{M\_le1} & \rotatebox{70}{DT\_ge0} & \rotatebox{70}{DT\_ge$-$1} & \rotatebox{70}{DT\_ge$-$2} & \rotatebox{70}{D\_ge34} & \rotatebox{70}{D\_ge33} & \rotatebox{70}{D\_ge32} & \rotatebox{70}{r\_le.125} & \rotatebox{70}{r\_le.25} & \rotatebox{70}{r\_ge.5} & \rotatebox{70}{r\_ge1} & \rotatebox{70}{k\_ge32} & \rotatebox{70}{k\_ge64} & \rotatebox{70}{k\_ge128} \\
    \hline
    \hline
    Chinchilla & ja & .27 & .40 & .23 & .03 & .26 & .36 & .43 & .39 & $-$.03 & .60 & .67 & $-$.17 & .01 & .25$^*$ & .37$^*$ & .20 & .01 & $-$.20 \\
    Chinchilla & id & .30 & .25 & $-$.04 & $-$.31 & .01 & .30 & .33 & .30 & $-$.00 & .32 & .38 & $-$.10 & $-$.01 & .18$^*$ & .16$^*$ & $-$.26 & $-$.38 & $-$.50 \\
    Chinchilla & sw & .35 & .31 & .03 & $-$.18 & $-$.08$^*$ & --- & --- & --- & $-$.01 & .40 & .52 & .16 & .16 & .18$^*$ & .16$^*$ & $-$.04 & $-$.19 & $-$.27 \\
    \hline
    He & ja & .69 & .55 & .74 & .29 & .09 & .98 & .94 & .67 & .19 & .69 & .65 & .64 & .60 & --- & --- & $-$.40 & $-$.79 & $-$1.16 \\
    He & id & .39 & .17 & .37 & $-$.05 & $-$.26 & .97 & .93 & .91 & .37 & .44 & .35 & .83 & .78 & --- & --- & $-$.66 & $-$.83 & $-$.00 \\
    He & sw & $-$.14 & $-$.40 & .41 & $-$.08 & --- & --- & --- & --- & $-$8.33 & $-$2.89 & $-$.89 & .55 & $-$.88 & --- & --- & $-$1.51 & $-$2.02 & $-$2.80 \\
    \hline
    Muennighoff & ja & .64 & .64 & .65 & .68 & .56 & .81 & .89 & .64 & $-$.16 & .71 & .80 & $-$.00 & .31 & --- & --- & .50 & .45 & .49 \\
    Muennighoff & id & .49 & .47 & .05 & .24 & .43 & .79 & .68 & .68 & $-$.03 & .53 & .64 & $-$1.05 & $-$.58 & --- & --- & $-$.02 & .06 & .11 \\
    Muennighoff & sw & .53 & .52 & .32 & .33 & --- & --- & --- & --- & $-$1.42 & .06 & .53 & .44 & .59 & --- & --- & .33 & .25 & .29 \\
    \hline
    ATLAS & ja & .70 & .65 & .85 & .68 & .45 & .77 & .86 & .61 & $-$.55 & .59 & .74 & .40 & .60 & --- & --- & .61 & .58 & .65 \\
    ATLAS & id & .66 & .54 & .65 & .45 & .14 & .72 & .62 & .63 & $-$.78 & .25 & .63 & .19 & .46 & --- & --- & .37 & .39 & .48 \\
    ATLAS & sw & .59 & .52 & .79 & .60 & --- & --- & --- & --- & $-$2.71 & $-$.45 & .43 & .26 & .50 & --- & --- & .36 & .29 & .35 \\
    \hline
    Sedova & ja & .86 & .85 & .71 & .55 & .72 & .93 & .92 & .90 & .47 & .83 & .86 & .80 & .72 & $-$.11$^*$ & .43$^*$ & .42 & .32 & .25 \\
    Sedova & id & .73 & .57 & .14 & .13 & .30 & .83 & .79 & .78 & .81 & .54 & .65 & .56 & .59 & $-$.20$^*$ & .15$^*$ & .07 & .01 & .39 \\
    Sedova & sw & .69 & .54 & .66 & .56 & .09$^*$ & --- & --- & --- & .84 & .46 & .68 & .81 & .83 & $-$.93$^*$ & .00$^*$ & .14 & .11 & .14 \\
    \hline
    $M^3$ & ja & .88 & .89 & .90 & .68 & .57 & .99 & .95 & .73 & .85 & .97 & .93 & .76 & .82 & --- & --- & .65 & .63 & .72 \\
    $M^3$ & id & .81 & .69 & .75 & .64 & .54 & .98 & .92 & .88 & .72 & .91 & .81 & .87 & .82 & --- & --- & .46 & .43 & .50 \\
    $M^3$ & sw & .78 & .72 & .71 & .75 & --- & --- & --- & --- & $-$.58 & .37 & .73 & .84 & .71 & --- & --- & .34 & .24 & .30 \\
    \hline
  \end{tabular}
  }
  \caption{Per-language, per-split test $R^2$ for \textbf{(b)}~1-stage data only, corresponding to Table~\ref{tab:baselines_extended}(b). Splits marked with $^*$ are excluded from the aggregation in Table~\ref{tab:baselines_extended} because at least one model failed to fit on that split.}
  \label{tab:full_splits_b}
\end{table*}

\begin{table*}[!htb]
  \centering
  \scalebox{0.74}{
  \setlength{\tabcolsep}{2.5pt}
  \begin{tabular}{@{}ll|rr|rrr|rrr|rrr|rrr@{}}
    \hline
    & & \multicolumn{2}{c|}{$C$} & \multicolumn{3}{c|}{$M$} & \multicolumn{3}{c|}{$D_T$} & \multicolumn{3}{c|}{$D$} & \multicolumn{3}{c}{$k$} \\
    Model & Lang & \rotatebox{70}{C\_ge0} & \rotatebox{70}{C\_ge$-$1} & \rotatebox{70}{M\_le$-$1} & \rotatebox{70}{M\_le0} & \rotatebox{70}{M\_le1} & \rotatebox{70}{DT\_ge0} & \rotatebox{70}{DT\_ge$-$1} & \rotatebox{70}{DT\_ge$-$2} & \rotatebox{70}{D\_ge34} & \rotatebox{70}{D\_ge33} & \rotatebox{70}{D\_ge32} & \rotatebox{70}{k\_ge32} & \rotatebox{70}{k\_ge64} & \rotatebox{70}{k\_ge128} \\
    \hline
    \hline
    Muennighoff & ja & .79 & .85 & .33 & .83 & .72 & .97 & .93 & .76 & --- & .96 & .94 & .62 & .64 & .71 \\
    Muennighoff & id & .76 & .61 & --- & .69 & .79 & .98 & .91 & .87 & --- & .92 & .81 & .40 & .36 & .52 \\
    Muennighoff & sw & .78 & .67 & --- & .78 & --- & --- & --- & --- & --- & .92 & .84 & .28 & .30 & .40 \\
    \hline
    ATLAS & ja & .78 & .77 & .71 & .49 & .38 & .98 & .93 & .76 & --- & .93 & .92 & .62 & .64 & .69 \\
    ATLAS & id & .70 & .56 & --- & .43 & .11 & .98 & .93 & .89 & --- & .79 & .78 & .39 & .38 & .48 \\
    ATLAS & sw & .77 & .63 & --- & .58 & --- & --- & --- & --- & --- & .92 & .84 & .27 & .31 & .40 \\
    \hline
    Sedova & ja & .83 & .80 & .39 & .64 & .63 & .95 & .58 & .92 & --- & .95 & .92 & .61 & .58 & .55 \\
    Sedova & id & .76 & .79 & --- & .80 & $-$2.60 & .88 & .91 & .67 & --- & .92 & .76 & .10 & .13 & .48 \\
    Sedova & sw & .77 & .77 & --- & .64 & .38$^*$ & --- & --- & --- & --- & .95 & .89 & .42 & .54 & .50 \\
    \hline
    $M^3{+}R_M^*(k)$ & ja & .85 & .88 & .83 & .76 & .68 & .97 & .94 & .68 & --- & .97 & .96 & .90 & .91 & .73 \\
    $M^3{+}R_M^*(k)$ & id & .90 & .85 & --- & .82 & .80 & .98 & .89 & .86 & --- & .92 & .93 & .59 & .72 & .85 \\
    $M^3{+}R_M^*(k)$ & sw & .93 & .89 & --- & .82 & --- & --- & --- & --- & --- & .82 & .91 & .81 & .69 & .68 \\
    \hline
    $M^3$ & ja & .79 & .85 & .33 & .83 & .72 & .97 & .93 & .76 & --- & .96 & .94 & .62 & .64 & .71 \\
    $M^3$ & id & .76 & .61 & --- & .69 & .79 & .98 & .91 & .87 & --- & .92 & .81 & .40 & .36 & .52 \\
    $M^3$ & sw & .78 & .67 & --- & .78 & --- & --- & --- & --- & --- & .92 & .84 & .28 & .30 & .40 \\
    \hline
  \end{tabular}
  }
  \caption{Per-language, per-split test $R^2$ for \textbf{(c)}~1-stage monolingual ($r{=}1$) data only, corresponding to Table~\ref{tab:baselines_extended}(c). The $r$ axis is not applicable. Splits marked with $^*$ are excluded from the aggregation in Table~\ref{tab:baselines_extended} because at least one model failed to fit on that split.}
  \label{tab:full_splits_c}
\end{table*}

\begin{table*}[!htb]
  \centering
  \scalebox{0.74}{
  \setlength{\tabcolsep}{2.5pt}
  \begin{tabular}{@{}ll|rr|rrr|rrr|rrr|rr@{}}
    \hline
    & & \multicolumn{2}{c|}{$C$} & \multicolumn{3}{c|}{$M$} & \multicolumn{3}{c|}{$D_T$} & \multicolumn{3}{c|}{$D$} & \multicolumn{2}{c}{$r$} \\
    Model & Lang & \rotatebox{70}{C\_ge0} & \rotatebox{70}{C\_ge$-$1} & \rotatebox{70}{M\_le$-$1} & \rotatebox{70}{M\_le0} & \rotatebox{70}{M\_le1} & \rotatebox{70}{DT\_ge0} & \rotatebox{70}{DT\_ge$-$1} & \rotatebox{70}{DT\_ge$-$2} & \rotatebox{70}{D\_ge34} & \rotatebox{70}{D\_ge33} & \rotatebox{70}{D\_ge32} & \rotatebox{70}{r\_le.125} & \rotatebox{70}{r\_ge.5} \\
    \hline
    \hline
    \multicolumn{15}{@{}l}{\textit{Full models (trained on 1-stage $+$ 2-stage, evaluated on 1-stage test)}} \\
    \hline
    He & ja & .59 & .53 & .37 & .01 & --- & .84 & .82 & --- & --- & --- & .52 & $-$.10 & --- \\
    He & id & .50 & .41 & --- & $-$.07 & --- & .71 & .45 & --- & --- & --- & .04 & $-$.45 & --- \\
    He & sw & --- & --- & --- & --- & --- & --- & --- & --- & --- & --- & --- & --- & --- \\
    \hline
    He $+$ Dual PL & ja & .96 & .95 & .89 & .64 & --- & .99 & .89 & --- & --- & --- & .92 & .82 & --- \\
    He $+$ Dual PL & id & .95 & .94 & --- & .67 & --- & .97 & .73 & --- & --- & --- & .82 & .82 & --- \\
    He $+$ Dual PL & sw & --- & --- & --- & --- & --- & --- & --- & --- & --- & --- & --- & --- & --- \\
    \hline
    Sedova & ja & .74 & .61 & .37 & .48 & .12$^*$ & .79 & .84 & .88$^*$ & --- & --- & .69 & $-$.00 & $-$.63$^*$ \\
    Sedova & id & .65 & .46 & --- & .51 & $-$.21$^*$ & .73 & .81 & .66$^*$ & --- & --- & .33 & $-$.39 & $-$.96$^*$ \\
    Sedova & sw & --- & --- & --- & .74$^*$ & .59$^*$ & --- & --- & --- & --- & --- & --- & .13$^*$ & --- \\
    \hline
    $M^3$ & ja & .95 & .95 & .96 & .79 & --- & .98 & .89 & --- & --- & --- & .85 & .85 & --- \\
    $M^3$ & id & .97 & .95 & --- & .83 & --- & .94 & .62 & --- & --- & --- & .68 & .78 & --- \\
    $M^3$ & sw & --- & --- & --- & --- & --- & --- & --- & --- & --- & --- & --- & --- & --- \\
    \hline
    $M^3$ $-$DPL & ja & .72 & .65 & .89 & .30 & --- & .76 & .79 & --- & --- & --- & .37 & .12 & --- \\
    $M^3$ $-$DPL & id & .65 & .50 & --- & .25 & --- & .63 & .33 & --- & --- & --- & $-$.14 & $-$.45 & --- \\
    $M^3$ $-$DPL & sw & --- & --- & --- & --- & --- & --- & --- & --- & --- & --- & --- & --- & --- \\
    \hline
    \hline
    \multicolumn{15}{@{}l}{\textit{Full models (evaluated on 2-stage test) and 2-stage-only models$^\dagger$}} \\
    \hline
    $M^3$ & ja & .86 & .86 & .45 & .28 & --- & .98 & .86 & --- & .95 & .93 & .72 & .74 & --- \\
    $M^3$ & id & .82 & .83 & .39 & .35 & --- & .96 & .62 & --- & $-$2.04 & .37 & .77 & .83 & --- \\
    $M^3$ & sw & --- & --- & --- & --- & --- & --- & --- & --- & --- & --- & --- & --- & --- \\
    \hline
    D-CPT$^\dagger$ & ja & .71 & .71 & .16 & $-$.30 & $-$.04 & .78 & .84 & .91 & .57 & .71 & .79$^*$ & .79 & .89 \\
    D-CPT$^\dagger$ & id & .60 & .60 & $-$.17 & $-$.85 & $-$1.26 & .83$^*$ & .90 & .85 & .66 & .82 & .79 & .80 & .85 \\
    D-CPT$^\dagger$ & sw & .70 & .70 & .75 & .70 & .71 & --- & --- & --- & --- & --- & --- & .86 & .72 \\
    \hline
    PTPP-F1$^\dagger$ & ja & .74 & .74 & .45 & .27 & .53 & .91 & .91 & .92 & .73 & .90 & .89$^*$ & .80 & .64 \\
    PTPP-F1$^\dagger$ & id & .41 & .41 & .05 & .34 & .02 & .88$^*$ & .91 & .86 & .97 & .95 & .91 & .71 & .63 \\
    PTPP-F1$^\dagger$ & sw & .61 & .61 & .64 & .70 & .75 & --- & --- & --- & --- & --- & --- & .76 & .65 \\
    \hline
    PTPP-F2$^\dagger$ & ja & .74 & .74 & .11 & $-$.28 & $-$.04 & .83 & .89 & .82 & .53 & .69 & ---$^*$ & .79 & .89 \\
    PTPP-F2$^\dagger$ & id & .49 & .49 & $-$.34 & $-$1.41 & $-$1.73 & ---$^*$ & .89 & .84 & .66 & .75 & .84 & .80 & .66 \\
    PTPP-F2$^\dagger$ & sw & .42 & .42 & .71 & .51 & .42 & --- & --- & --- & --- & --- & --- & .84 & .70 \\
    \hline
    PTPP-F3$^\dagger$ & ja & .74 & .74 & .42 & .27 & .54 & .89 & .90 & .92 & .74 & .87 & .91$^*$ & .72 & .62 \\
    PTPP-F3$^\dagger$ & id & .72 & .72 & .51 & .40 & $-$.17 & .92$^*$ & .92 & .92 & .88 & .96 & .91 & .72 & .76 \\
    PTPP-F3$^\dagger$ & sw & .59 & .59 & .68 & .71 & .79 & --- & --- & --- & --- & --- & --- & .78 & $-$4.47 \\
    \hline
    Zhang$^\dagger$ & ja & .43 & .43 & $-$2.26 & $-$1.28 & $-$.31 & .64 & .73 & .56 & $-$4.33 & $-$.25 & .67$^*$ & .58 & .67 \\
    Zhang$^\dagger$ & id & $-$.02 & $-$.02 & $-$4.37 & $-$2.68 & $-$1.35 & .72$^*$ & .72 & .43 & $-$4.73 & $-$.03 & .72 & .60 & .49 \\
    Zhang$^\dagger$ & sw & .13 & .13 & .20 & $-$.57 & $-$.41 & --- & --- & --- & --- & --- & --- & .46 & .78 \\
    \hline
  \end{tabular}
  }
  \caption{Per-language, per-split test $R^2$ for \textbf{(d)}~Multi-lingual single-epoch ($k{=}1$) evaluation, corresponding to Table~\ref{tab:baselines_extended}(d1, d2). The $k$ axis is not applicable. $^\dagger$2-stage-only models trained and tested on 2-stage data only. Splits marked with $^*$ are excluded from the aggregation in Table~\ref{tab:baselines_extended} because at least one model failed to fit on that split.}
  \label{tab:full_splits_d2}
\end{table*}

\subsection{$M^3$ Scaling Law Parameters Fitted on All Data}
\label{appendix:full_data_fits}

Table~\ref{tab:m3_fitted_params} shows the $M^3$ Scaling Law parameters fitted on the full dataset (1-stage $+$ 2-stage) for each language pair.

\begin{table}[!htb]
  \centering
  \scalebox{0.85}{
  \begin{tabular}{lrrr}
    \hline
        Parameter & Japanese & Indonesian & Swahili \\
    \hline
    \hline
    $A$ & 5598.7 & 6732.1 & 48958.3 \\
    $B$ & 3988.8 & 15625.9 & 3134.6 \\
    $\alpha$ & 0.504 & 0.532 & 0.646 \\
    $\beta$ & 0.426 & 0.515 & 0.418 \\
    $E$ & 1.548 & 1.760 & 2.428 \\
    $R_D^*$ & 10.18 & 7.91 & 5.14 \\
    $R_{D,\text{high}}^*$ & 51.89 & 35.14 & 19.34 \\
    $\psi$ & 3.232 & 4.273 & 4.107 \\
    $R_M^*$ & 23.80 & 25.67 & 53.62 \\
    $\gamma$ & 0.0834 & 0.0571 & 0.0815 \\
    $\gamma_2$ & 0.0343 & 0.0184 & 0.0544 \\
    \hline
  \end{tabular}
  }
  \caption{$M^3$ Scaling Law parameters fitted on the full dataset for each language pair.}
  \label{tab:m3_fitted_params}
\end{table}

\subsection{Analysis of Extending Effective Data Multiplier}
\label{appendix:flp_analysis}

\begin{figure}[t]
  \centering
  \includegraphics[width=\columnwidth]{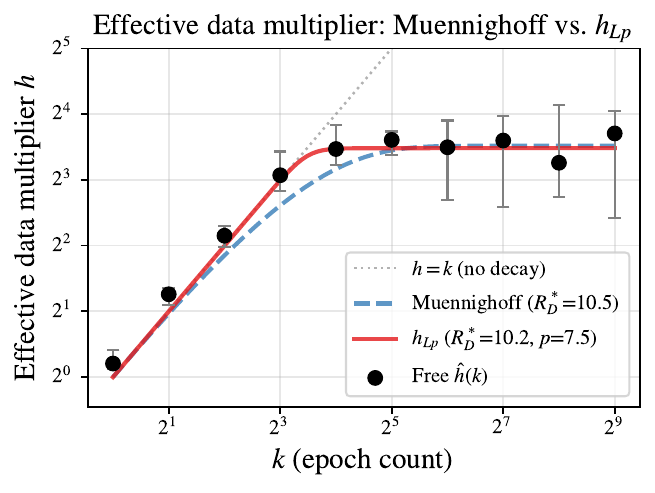}
  \caption{Effective data multiplier comparison on Japanese monolingual data.
  Black points show independently fitted $\hat{h}(k)$ for each epoch count $k$;
  the Muennighoff form $h$ (blue dashed) saturates too early,
  while the $h_{Lp}$ extension (red solid) with sharpness parameter $p$ more closely tracks the free estimates around the transition region. Error bars show 95\% bootstrap confidence intervals (200 resamples).}
  \label{fig:flp_comparison}
\end{figure}

The saturation function $h(R_D; R_D^*)$ in the Muennighoff model (Eq.~\ref{eq:effective_D_muennighoff}) may not fully capture how the effective data size $D'$ transitions from the linear regime to the saturation regime.
To see this, we focus on the Japanese monolingual single-stage setting ($r=1$).
We first fit and fix the Chinchilla base parameters $(A, B, \alpha, \beta, E)$ of $L_\text{base}$ (Eq.~\ref{eq:chinchilla}) using the subset with $k \leq 4$.
We then model the effective data multiplier in $D' = D_T \cdot \hat{h}$ in three ways, each fit using the Muennighoff model $L_\text{Mu}$ (Eq.~\ref{eq:muennighoff_main}) on all monolingual data:
(i) a \textit{discrete} parameterization that fits an independent $\hat{h}_k$ for each $k$, jointly with a shared $R_M^*$ for $M'$;
(ii) the parametric \textit{Muennighoff} form $\hat{h}=h(R_D; R_D^*)$ (Eq.~\ref{eq:effective_D_muennighoff}), fitting $R_D^*$ and $R_M^*$; and
(iii) the parametric form $\hat{h}=h_{Lp}(R_D; R_D^*, p)$ defined below, fitting $R_D^*$, $p$, and $R_M^*$.
As shown in Figure~\ref{fig:flp_comparison}, the Muennighoff form $h(R_D; R_D^*)$ captures the convergence of the free $\hat{h}(k)$ to a constant in the large-$k$ regime, but does not capture the sharpness of the transition from the linear ($\hat{h}\approx k$) to the saturation regime.
A sharper transition can be modeled by introducing a sharpness parameter $p$:
\begin{equation}
h_{Lp}(R_D;\, R_D^{*},\, p) = 1 + \frac{R_D}{\left(1 + \left(\dfrac{R_D}{R_D^{*}}\right)^{p}\right)^{1/p}},
\label{eq:effective_D_lp}
\end{equation}
which preserves the same asymptotic limit $1+R^*$ as the original $h$.
As shown in Table~\ref{tab:flp_ablation}, replacing $h$ with $h_{Lp}$ in the $M^3$ Scaling Law improves test $R^2$ on the full data (Avg $.69$ vs.\ $.67$), primarily through improved $k$-axis extrapolation.

\begin{figure}[t]
  \centering
  \includegraphics[width=\columnwidth]{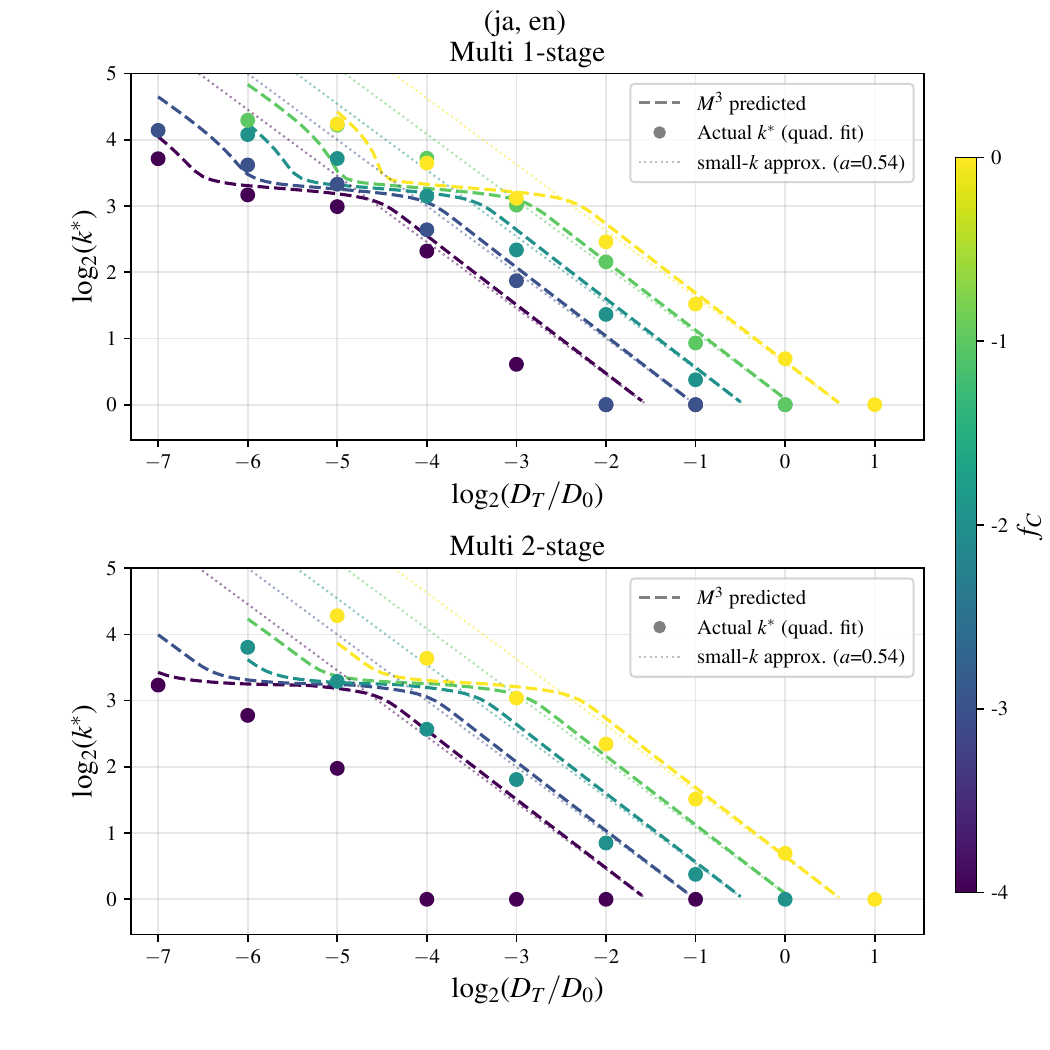}
  \caption{Same as Figure~\ref{fig:main_enja_D_T_optimal_k_merge} but with $h_{Lp}$ replacing $h$ in the $M^3$ Scaling Law. The predicted $k^*$ curves (dashed) exhibit non-smooth behavior near the saturation knee of $h_{Lp}$, unlike the more smooth curves obtained with the original $h$ (Figure~\ref{fig:main_enja_D_T_optimal_k_merge}).}
  \label{fig:appendix_flp_optimal_k}
\end{figure}

However, we do not adopt $h_{Lp}$ in the proposed model for two reasons.
First, the sharp saturation of $h_{Lp}$ introduces non-smooth behavior in the predicted optimal number of epochs $k^*$ (Figure~\ref{fig:appendix_flp_optimal_k}): as $D_T$ decreases, $k^*$ rises smoothly until it reaches the saturation knee of $h_{Lp}$, at which point $k^*$ plateaus abruptly before resuming its increase---a pattern not observed in the empirical estimates.
Second, the non-smooth transition of $h_{Lp}$ seems intuitively unnatural for LLM training, where data repetition effects would be expected to degrade smoothly rather than switching abruptly between a linear and a saturated regime.
The original Muennighoff form $h$, while fitting the per-$k$ data less tightly, produces smoother $k^*$ predictions that better match empirical trends (Figure~\ref{fig:main_enja_D_T_optimal_k_merge}).

\begin{table}[!htb]
  \centering
  \scalebox{0.82}{
  \setlength{\tabcolsep}{3pt}
  \begin{tabular}{@{}llc*{6}{r}r@{}}
    \hline
    Data & $D'$ form & \#P & $C$ & $M$ & $D_T$ & $D$ & $r$ & $k$ & Avg \\
    \hline
    \hline
    \multicolumn{10}{@{}l}{\textit{All data (1-stage $+$ 2-stage)}} \\
    \hline
    & $h_{Lp}$ & 12 & \textbf{.80} & \textbf{.62} & .85 & .50 & \textbf{.78} & \textbf{.56} & \textbf{.69} \\
    & $h$ (adopted) & 11 & .79 & .60 & \textbf{.85} & \textbf{.52} & .75 & .52 & .67 \\
    \hline
    \hline
    \multicolumn{10}{@{}l}{\textit{1-stage monolingual ($r{=}1$), with $R_M^*(k)$}} \\
    \hline
    & $h_{Lp}$ & 10 & \textbf{.88} & .78 & .88 & .92 & --- & \textbf{.77} & .85 \\
    & $h$ (adopted) & 9 & .88 & \textbf{.79} & \textbf{.89} & \textbf{.92} & --- & .77 & \textbf{.85} \\
    \hline
    \multicolumn{10}{@{}l}{\textit{1-stage monolingual ($r{=}1$), constant $R_M^*$}} \\
    \hline
    & $h_{Lp}$ & 8 & \textbf{.77} & \textbf{.75} & .88 & \textbf{.90} & --- & \textbf{.53} & \textbf{.77} \\
    & $h$ (adopted) & 7 & .74 & .71 & \textbf{.90} & .90 & --- & .47 & .75 \\
    \hline
  \end{tabular}
  }
  \caption{Effect of replacing the Muennighoff form $h$ with $h_{Lp}$ (Eq.~\ref{eq:effective_D_lp}).
  On the full data, $h_{Lp}$ improves Avg $R^2$ (+.05), but we adopt $h$ for smoother optimal-$k^*$ predictions; see text.
  Evaluation follows the same protocol as Table~\ref{tab:baselines_extended}.}
  \label{tab:flp_ablation}
\end{table}

\subsection{Effect of First-Stage Language Ratio $r_1$}
\label{subsec:r1_effect}

The dual power law (Eq.~\ref{eq:dual_power_law}) formulates the language ratio factor for two-stage training as
$\mathcal{R}_\text{dual} = r_f^{-\gamma} \cdot (r/r_f)^{-\gamma_2}$,
treating the exponent $\gamma_2$ as a constant independent of the first-stage language ratio $r_1$.
As noted in \S\ref{subsubsec:dual_power_law_two_stage}, Figure~\ref{fig:Lr_factor_k1}(b) suggests that $r_1$ has only a minor effect: lower $r_1$ (i.e., more high-resource language in the first stage) tends to slightly increase the loss.
Here we quantitatively assess whether explicitly modeling this $r_1$ dependence improves extrapolation accuracy.

We extend $\gamma_2$ to depend on $r_1$ as follows:
\begin{equation}
  \tilde{\gamma}_2(r_1) = \gamma_2 + \dot{\gamma}_2 \cdot \ln(r_0 + r_1),
  \label{eq:gamma2_r1}
\end{equation}
where $\dot{\gamma}_2$ controls the strength of $r_1$ dependence and $r_0$ is an offset to prevent divergence at $r_1 = 0$.
When $\dot{\gamma}_2 = 0$, this reduces to the constant-$\gamma_2$ model.
The extended dual power law becomes:
\begin{equation}
  \mathcal{R}_\text{dual}(r, r_1, r_f) = r_f^{-\gamma} \cdot \left(\frac{r}{r_f}\right)^{-\tilde{\gamma}_2(r_1)}.
  \label{eq:dual_power_law_r1}
\end{equation}

We incorporate this extension into the $M^3$ Scaling Law (Eq.~\ref{eq:m3_main}), adding two parameters ($\dot{\gamma}_2$ and $r_0$) for a total of 13.
Table~\ref{tab:r1_effect} compares the original 11-parameter model against this 13-parameter variant using the same multi-split evaluation protocol described in \S\ref{subsec:eval_m3}.

\begin{table}[!htb]
  \centering
  \begin{tabular}{lcrr}
    \hline
    Model & \#P & ja & id \\
    \hline
    $M^3$ ($\dot{\gamma}_2 = 0$) & 11 & 0.704 & 0.711 \\
    $M^3$ + $r_1$-dep.\ $\gamma_2$ & 13 & 0.705 & 0.707 \\
    \hline
  \end{tabular}
  \caption{Average test $R^2$ across multi-split evaluation (setting~(a) of Table~\ref{tab:baselines_extended}) with and without $r_1$-dependent $\gamma_2$, evaluated on Japanese (ja) and Indonesian (id).
  The extended model adds $\dot{\gamma}_2$ and $r_0$ to the $M^3$ Scaling Law.}
  \label{tab:r1_effect}
\end{table}

The $r_1$-dependent extension yields no notable improvement in average test $R^2$ for either language.
Based on the result, we conclude that the effect of $r_1$ on the scaling behavior is negligible at the current experimental scale, and adopt the constant-$\gamma_2$ formulation in the $M^3$ Scaling Law.

\subsection{Trade-off Between Modeling Routes for $(k, r)$ Dependence}
\label{appendix:gfit_details}

As discussed in \S\ref{subsec:interaction_multi_epoch_multi_lingual}, the $(k, r)$ dependence of the loss can be carried either through $R_M^*$ (in $M'$) or through the weight $g$ on $D_\text{high}$ (in $D'$).
We first inspect the per-$(k,r)$ trends of each parameter to motivate the parametric forms, and then report the full fitting experiments behind the trade-off observed between these two routes.

\subsubsection{Pilot Analysis: Per-\texorpdfstring{$(k, r)$}{(k, r)} Fitting of \texorpdfstring{$g$}{g} and \texorpdfstring{$R_M^*$}{R\_M*}}
\label{subsubsec:per_kr_pilot}

\begin{figure*}[t]
  \centering
  \includegraphics[width=\textwidth]{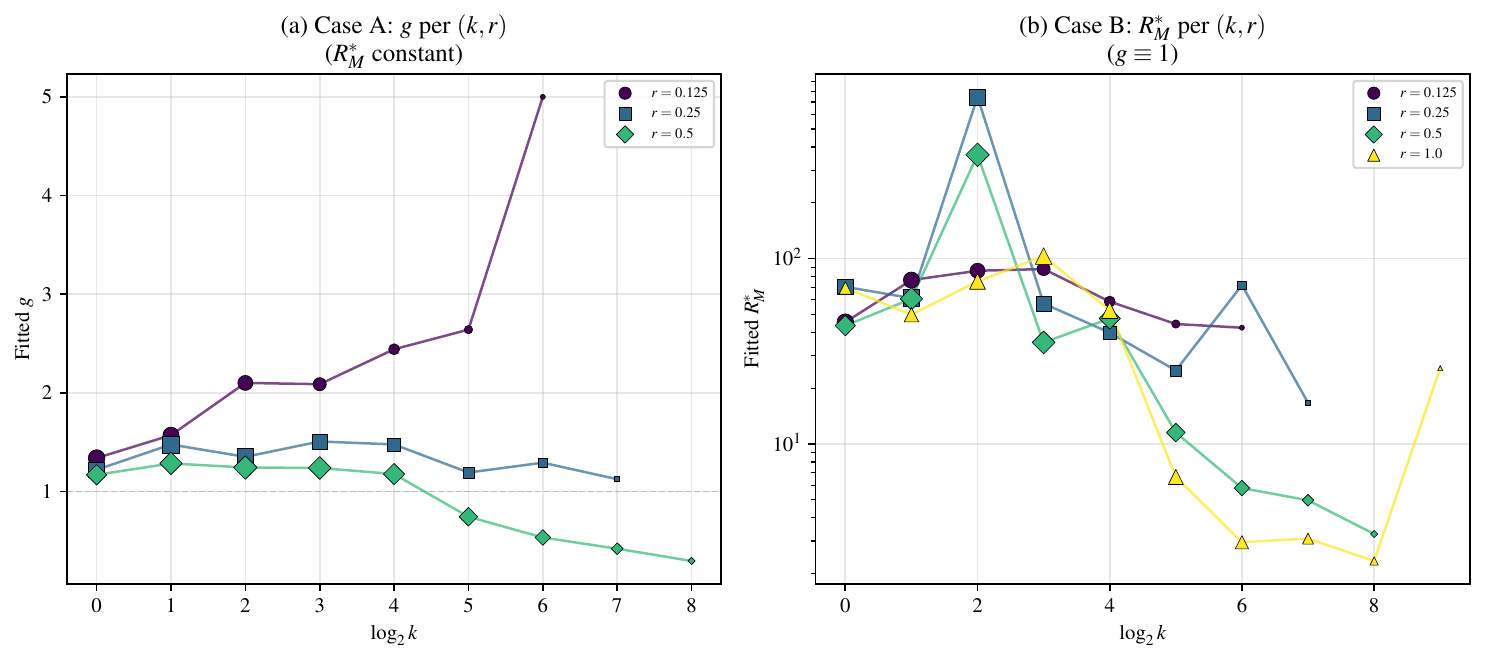}
  \caption{Per-$(k,r)$ fitting of $g$ and $R_M^*$ on (Japanese, English) single-stage data.
  Marker sizes are proportional to the number of data points for each $(k,r)$ combination.
  \textbf{(a)} Case~A: $R_M^*$ is a shared constant; an independent $g_{k,r}$ is fitted for each $(k,r)$ with $r<1$. At low $k$, $g$ is close to~1 for all $r$; at higher $k$, $g$ decreases for large $r$ but increases for small $r$.
  \textbf{(b)} Case~B: $g\equiv 1$; an independent $R_{M,k,r}^*$ is fitted for each $(k,r)$. $R_M^*$ is large at small $k$ across all $r$, and decreases with $k$ more rapidly at larger~$r$.}
  \label{fig:per_kr_gfit}
\end{figure*}

Starting from the multiplicative form of $L = L_\text{base}(M', D') \cdot \mathcal{R}_\text{dual}(r, r_f)$ with effective sizes $M' = U_M \cdot h(R_M;\, R_M^*)$ and $D' = D_T \cdot h(R_D;\, R_D^*) + g \cdot D_\text{high}$, we seek to determine how $g$ and $R_M^*$ each depend on $(k, r)$.
We fix the Chinchilla base parameters $(A, B, \alpha, \beta, E)$ from the monolingual low-epoch ($k \leq 4$) subset and then simultaneously fit the remaining global parameters together with per-$(k,r)$ free parameters on all single-stage data:

\textbf{Case~A.} We fix $R_M^*$ as a single shared constant and fit an independent $g_{k,r}$ for each $(k,r)$ with $r<1$ (at $r=1$, $D_\text{high}=0$ makes $g$ unidentifiable), jointly with the global parameters $R_D^*$, $R_M^*$, and $\gamma$.

\textbf{Case~B.} We fix $g\equiv 1$ and fit an independent $R_{M,k,r}^*$ for each $(k,r)$, jointly with the global parameters $R_D^*$ and $\gamma$.

Figure~\ref{fig:per_kr_gfit} shows the results.
In Case~A (panel (a)), $g$ is close to~1 at $k{=}1$ across all $r$, but develops $k$-dependent variation: for large $r$ ($r{=}0.5$), $g$ decreases below~1, while for small $r$ ($r{=}0.125$), $g$ increases well above~1.
However, $g$ growing above its $k{=}1$ baseline of approximately~1 as $k$ increases would counterintuitively imply that repeating data increases the effective data size beyond the non-repeated case.
To avoid this, we sought a parametric form that satisfies $g \leq 1$ while allowing the floor to depend on $r$.
This led to the form adopted in Eq.~(\ref{eq:g_func}), which decays from $g{=}1$ at $R_D{=}0$ to an $r$-dependent floor $(1{-}r)^\psi < 1$.

In Case~B (panel~b), $R_M^*$ is large at small $k$ for all $r$, indicating weak epoch saturation of the effective model size, but decreases with $k$ more rapidly at larger $r$.
This is a natural consequence: when the fraction of repeated target-language tokens is higher (larger $r$), multi-epoch repetition has a stronger saturation effect on $M'$.
Based on this trend, we model the $r$-dependence as
\begin{equation}
  R_M^*(k, r) = R_M^*(k) \cdot r^{-d},
  \label{eq:rm_star_kr}
\end{equation}
where $R_M^*(k)$ is the monolingual form from Eq.~(\ref{eq:rm_star_mono}) and $d > 0$ is a new parameter.
This form yields larger $R_M^*$ (weaker saturation) at smaller $r$, consistent with the observed trend.

\subsubsection{Experimental Setup}

We compare four combinations of $R_M^*$ and $g$ forms:
\begin{itemize}
  \item \textbf{$R_M^*(k, r) + g \equiv 1$}: $(k, r)$ dependence is carried entirely by $R_M^*$, with $g$ fixed to constant. Here, $R_M^*(k, r)$ follows Eq.~(\ref{eq:rm_star_kr}).
  \item \textbf{$R_M^*(k, r) + g(r, R_D)$}: $(k, r)$ dependence is carried jointly by both routes, with $R_M^*(k, r)$ as above and $g$ following Eq.~(\ref{eq:g_func}).
  \item \textbf{$R_M^*(k) + g(r, R_D)$}: $R_M^*$ retains only $k$-dependence (Eq.~\ref{eq:rm_star_mono}), with the $(k, r)$ dependence otherwise carried by $g$.
  \item \textbf{$R_M^* = \text{const} + g(r, R_D)$ (adopted)}: $R_M^*$ is a single global constant, with $(k, r)$ dependence carried entirely by $g$. This is the form adopted in the $M^3$ Scaling Law (\S\ref{subsec:proposed_model}).
\end{itemize}
Each variant is fit and evaluated under the same protocol as the main scaling law evaluation (\S\ref{subsec:eval_m3}).

\subsubsection{Results}

Table~\ref{tab:gfit_tradeoff} summarizes the extrapolation accuracy of each variant.
Configurations that place the $(k, r)$ or $k$ dependence on $R_M^*$ (the first three variants) achieve better $k$-axis extrapolation than the adopted form, but degrade $M$-axis extrapolation; the adopted form ($R_M^* = \text{const}$ with $g(r, R_D)$) shows the opposite tendency.

\begin{table}[!htb]
  \centering
  \scalebox{0.66}{
  \setlength{\tabcolsep}{3pt}
  \begin{tabular}{@{}l@{\hskip 4pt}lc*{6}{r}r@{}}
    \hline
    & Variant & \#P & $C$ & $M$ & $D_T$ & $D$ & $r$ & $k$ & Avg \\
    \hline
    \hline
    & $R_M^*(k, r)$ $+$ $g \equiv 1$ & 12 & .81 & .45 & .87 & .22 & \textbf{.79} & \textbf{.73} & .64 \\
    & $R_M^*(k, r)$ $+$ $g(r, R_D)$ & 14 & \textbf{.83} & .44 & .86 & .44 & .75 & .72 & .67 \\
    & $R_M^*(k)$ $+$ $g(r, R_D)$ & 13 & .81 & .44 & .86 & .51 & .78 & .65 & \textbf{.68} \\
    & $R_M^* {=} \text{const}$ $+$ $g(r, R_D)$ (adopted) & 11 & .79 & .60 & .85 & \textbf{.52} & .75 & .52 & .67 \\
    & $R_M^* {=} \text{const}$ $+$ $g \equiv 1$ & 9 & .70 & \textbf{.63} & \textbf{.88} & $-$.41 & .75 & .34 & .48 \\
    \hline
  \end{tabular}
  }
  \caption{Trade-off between modeling routes for $(k, r)$ dependence.
  Each row shows the test $R^2$ (macro-averaged over held-out splits and languages)
  for a different attribution of the $(k, r)$ interaction to $R_M^*$ (in $M'$) vs.\ $g$ (in $D'$).
  Evaluation follows the same protocol as Table~\ref{tab:baselines_extended}(a).}
  \label{tab:gfit_tradeoff}
\end{table}

\subsubsection{Hypothesis for the Trade-off}

We hypothesize that this trade-off stems from an asymmetry between the two routes: $R_M^*$ enters the loss in both monolingual ($r=1$) and multi-lingual ($r<1$) settings, whereas $g$ contributes only through $D_\text{high}$ and is inactive at $r=1$.
Absorbing $(k, r)$ into $R_M^*$ thus captures the $k$-dependence from both settings, but couples the $D_\text{high}$ contribution to the $M$-axis, which may not transfer to unseen $M$ at $r<1$ and degrade $M$-axis extrapolation.
Absorbing $(k, r)$ into $g$ instead decouples the $(k, r)$ contribution from the $M$-axis, yielding more conservative $M$-axis behavior, at the cost of capturing the $k$-dependence only in multi-lingual settings and weakening $k$-axis extrapolation.

\subsection{Details of Analysis of Optimal Number of Epochs}
\label{appendix:optimal_epochs}

\subsubsection{Empirical Estimation of $k^*$ via Quadratic Fitting}

To estimate the optimal number of epochs $k^*(C, D_T)$ from experimental data, we first compute the minimum validation loss $L^*(C, D_T, k)$ achieved across all training setups sharing the same $(C, D_T, k)$.
For \textbf{multi 1-stage}, the minimum is taken over model scale $M$ and $r$ for single-stage setups; for \textbf{multi 2-stage}, it is taken over $M$, $r$, $r_1$, and $r_2$ including both single-stage and two-stage setups.

\begin{figure*}[t]
  \includegraphics[width=\textwidth]{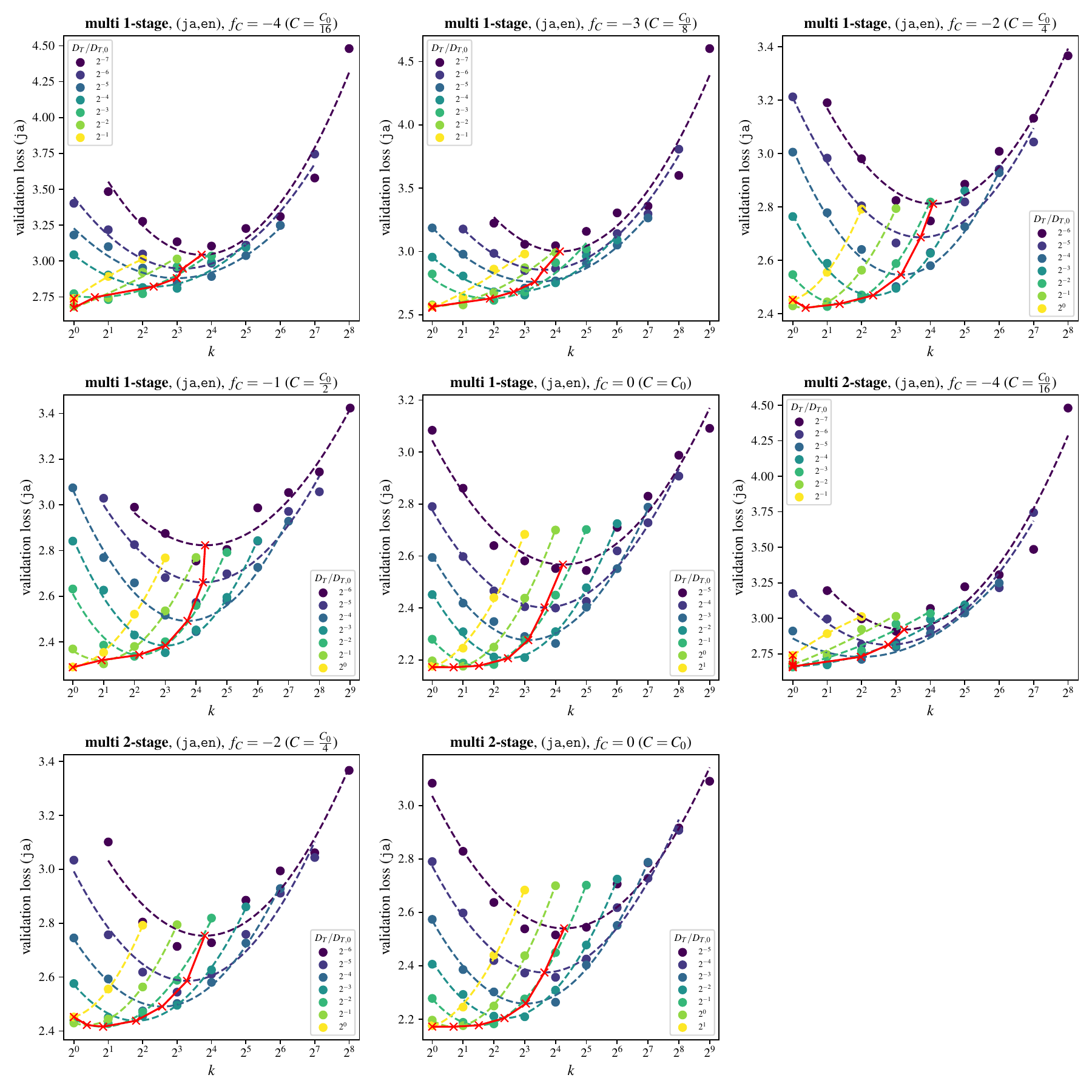}
  \caption{Change of minimum target language validation loss $L^*(C, D_T,k)$ by number of epochs $k$ (x-axis) for each target language corpus size $D_T$ (color) ((Japanese, English) pair). Each dashed line represents a quadratic function of $f_k=\log_2 k$ fitted to estimate $L^*(C,D_T,k)$. The red solid line connects minimum points of the fitted curves.}
  \label{fig:supplementary_all_enja_optimal_k_quad_fitting_merge}
\end{figure*}

\begin{figure*}[t]
  \includegraphics[width=\textwidth]{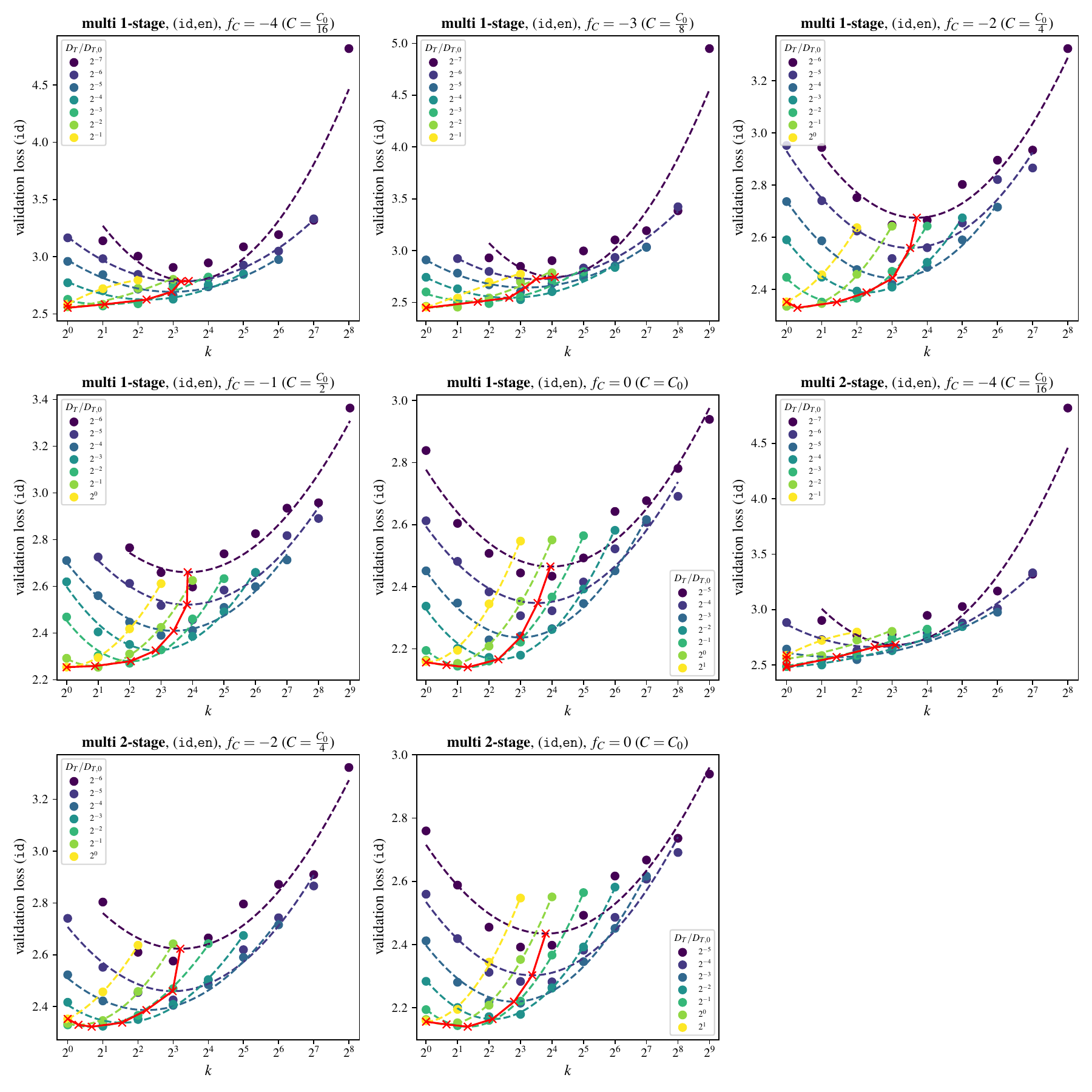}
  \caption{Change of minimum target language validation loss $L^*(C, D_T,k)$ by number of epochs $k$ (x-axis) for each target language corpus size $D_T$ (color) ((Indonesian, English) pair). Each dashed line represents a quadratic function of $f_k=\log_2 k$ fitted to estimate $L^*(C,D_T,k)$. The red solid line connects minimum points of the fitted curves.}
  \label{fig:supplementary_all_enja_optimal_k_quad_fitting_merge_id}
\end{figure*}

We then fit a quadratic function $L^* = p_0 f_k^2 + p_1 f_k + p_2$ in $f_k = \log_2 k$ to the resulting $(f_k, L^*)$ pairs for each $(C, D_T)$, and estimate $k^*$ as the vertex of the fitted parabola, i.e., $f_k^* = -p_1/(2p_0)$ and $k^* = 2^{f_k^*}$.
This smoothing mitigates the impact of noise and the limited resolution of our $k$ grid ($k \in \{1, 2, 4, \ldots\}$).
We exclude estimates where (i)~the fitted parabola has a maximum rather than a minimum ($p_0 \leq 0$) within the observed range, indicating a noisy fit, or (ii)~the vertex lies below the smallest observed $f_k$ and $k{=}1$ was not tested, as the true optimum cannot be reliably determined.
When the vertex falls at or below $f_k{=}0$ and $k{=}1$ is included in the search grid, we clamp $k^*$ to~1.
Figure~\ref{fig:supplementary_all_enja_optimal_k_quad_fitting_merge} and Figure~\ref{fig:supplementary_all_enja_optimal_k_quad_fitting_merge_id} illustrate the quadratic fits for the (Japanese, English) and (Indonesian, English) pairs respectively, showing the fitted curves together with the observed $L^*(C, D_T, k)$ values.

\subsubsection{Results for Indonesian}

\begin{figure}[t]
  \centering
  \includegraphics[width=\columnwidth]{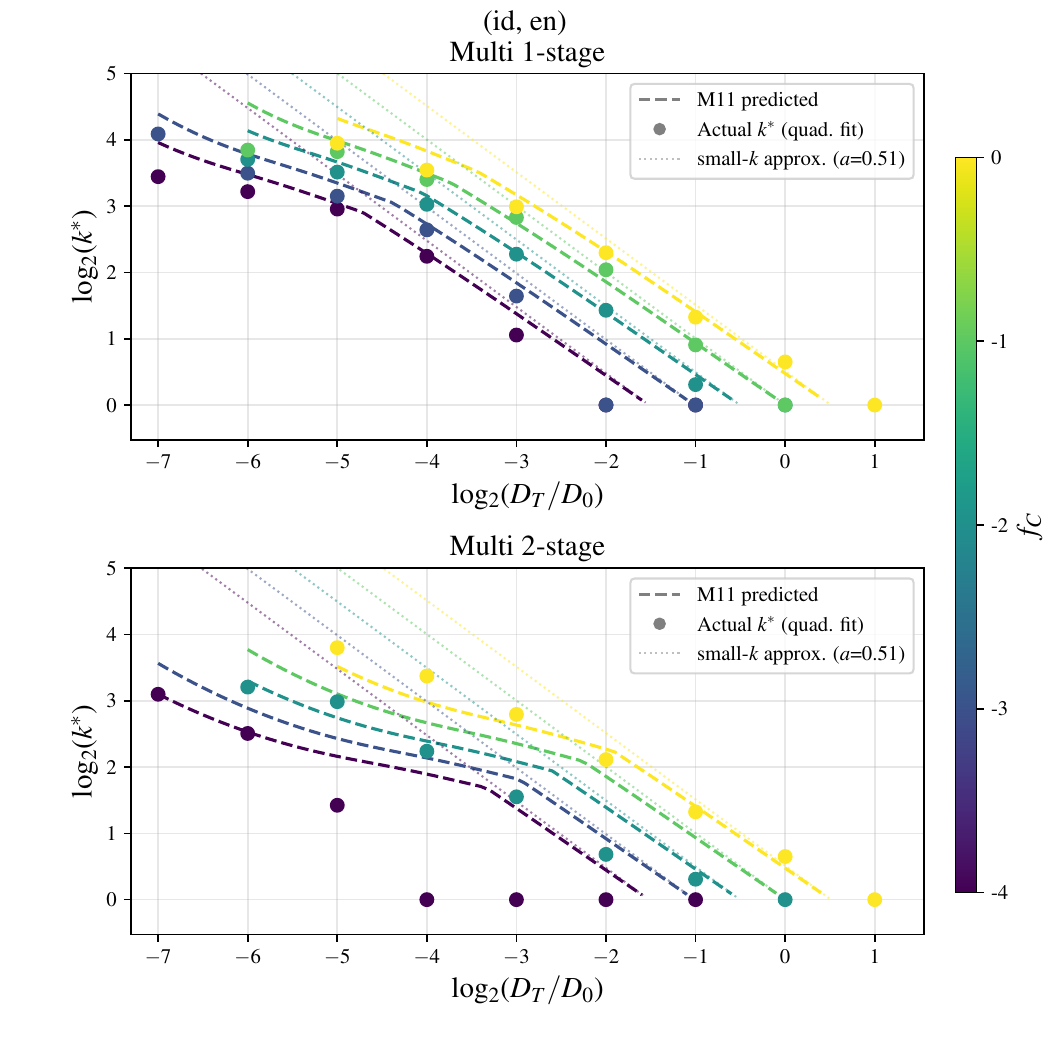}
  \caption{Same as Figure~\ref{fig:main_enja_D_T_optimal_k_merge} but for the (Indonesian, English) pair.}
  \label{fig:supplementary_optimal_k_id}
\end{figure}

Figure~\ref{fig:supplementary_optimal_k_id} shows the optimal number of epochs analysis for the (Indonesian, English) pair.
We do not present optimal epoch analysis for the (Swahili, English) pair, as the available $D_T$ range is too narrow to investigate the dependence of $k^*$ on $D_T$ across compute budgets.

\subsubsection{Derivation of Optimal Number of Epochs}
\label{appendix:optimal_epochs_derivation}

We now prove Eq.~(\ref{eq:k_star_collapse}) for the $M^3$ Scaling Law in Eqs.~(\ref{eq:m3_main})--(\ref{eq:m3_Mprime}).
Since $M$ is defined as the non-embedding FLOPs per token, the total
compute is simply
\begin{equation}
\begin{aligned}
C = MD,
\qquad
D = \frac{kD_T}{r}.
\end{aligned}
\end{equation}

In the derivation below, we restrict attention to the fixed-$r$
setting and optimize only over $k$ under the compute constraint.
This directly covers single-stage training, for which $r_f=r$; in
particular, the monolingual regime corresponds to $r=1$.
We state the argument for an arbitrary fixed $r$ because the same
homogeneity proof goes through without modification, but we do not
claim here that the globally optimal $r$ remains fixed as $D_T$
changes.
Thus, the proof is exact for fixed-$r$ settings (especially mono
1-stage), whereas the collapse observed for jointly optimized
multi 2-stage training should be interpreted as empirical.

\paragraph{Proposition.}
Fix a ratio schedule $\mathbf{r}$.
Then, for the $M^3$ Scaling Law, the argmin over $k$ depends on $(C,D_T)$
only through the scarcity variable $D_T/D^*(C)$, where $D^*(C)$ is the
monolingual single-epoch Chinchilla-optimal data size.
When the minimizer is unique, this is written as
\begin{equation}
\begin{aligned}
k_{\mathbf{r}}^*(C,D_T)
=
K_{\mathbf{r}}\!\left(
\frac{D_T}{D^*(C)}
\right).
\end{aligned}
\end{equation}

\paragraph{Proof.}
Let
\begin{equation}
\begin{aligned}
x = \frac{D_T}{C^{\alpha/(\alpha+\beta)}}.
\end{aligned}
\end{equation}
Then
\begin{equation}
\begin{aligned}
D_T &= C^{\alpha/(\alpha+\beta)} x, \\
M
&=
\frac{Cr}{kD_T}
=
C^{\beta/(\alpha+\beta)}
\frac{r}{kx}.
\end{aligned}
\end{equation}

For brevity, write
\begin{equation}
\begin{aligned}
g_k(r)
=
g\!\left(
r,\,
k-1;\,
R_{D,\text{high}}^*,\,
\psi
\right).
\end{aligned}
\end{equation}
Using Eq.~(\ref{eq:m3_Dprime}) and
$D_\text{high} = kD_T(1-r)/r$, the effective data size becomes
\begin{equation}
\begin{aligned}
D'
&=
D_T \cdot h(k-1;R_D^*)
+
g_k(r)\cdot \frac{kD_T(1-r)}{r}
\\
&=
D_T
\Biggl[
h(k-1;R_D^*)
+
\frac{k(1-r)}{r} g_k(r)
\Biggr]
\\
&\equiv
D_T\,Q_D(k;r)
\\
&=
C^{\alpha/(\alpha+\beta)}
x\,Q_D(k;r),
\end{aligned}
\end{equation}
where
\begin{equation}
\begin{aligned}
Q_D(k;r)
=
h(k-1;R_D^*)
+
\frac{k(1-r)}{r} g_k(r).
\end{aligned}
\end{equation}
Hence
\begin{equation}
\begin{aligned}
\frac{1}{D'^\beta}
=
C^{-\alpha\beta/(\alpha+\beta)}
\frac{1}{x^\beta Q_D(k;r)^\beta}.
\end{aligned}
\end{equation}

Next, recall that
\begin{equation}
\begin{aligned}
U_M
=
\min\!\left(
G^{(\alpha+\beta)/\alpha} D_T^{\beta/\alpha},
\, M
\right),
\end{aligned}
\end{equation}
where
\begin{equation}
\begin{aligned}
G
=
\left(
\frac{\alpha A}{\beta B}
\right)^{1/(\alpha+\beta)}.
\end{aligned}
\end{equation}
Define
\begin{equation}
\begin{aligned}
Q_U(x,k;r)
=
\min\!\left(
G^{(\alpha+\beta)/\alpha} x^{\beta/\alpha},
\,
\frac{r}{kx}
\right).
\end{aligned}
\end{equation}
Then
\begin{equation}
\begin{aligned}
U_M
&=
\min\!\Bigl(
G^{(\alpha+\beta)/\alpha}
C^{\beta/(\alpha+\beta)} x^{\beta/\alpha},
\\
&\qquad\qquad
C^{\beta/(\alpha+\beta)}\frac{r}{kx}
\Bigr)
\\
&=
C^{\beta/(\alpha+\beta)} Q_U(x,k;r).
\end{aligned}
\end{equation}
Therefore,
\begin{equation}
\begin{aligned}
R_M
=
\frac{M}{U_M}-1
=
\frac{r}{kx\,Q_U(x,k;r)} - 1,
\end{aligned}
\end{equation}
which depends only on $(x,k,r)$.

Using Eq.~(\ref{eq:m3_Mprime}),
\begin{equation}
\begin{aligned}
M'
&=
U_M \cdot h(R_M;R_M^*)
\\
&=
C^{\beta/(\alpha+\beta)}
Q_U(x,k;r)\,
h(R_M;R_M^*)
\\
&\equiv
C^{\beta/(\alpha+\beta)}
Q_M(x,k;r),
\end{aligned}
\end{equation}
where $Q_M(x,k;r)$ is independent of $C$ and $D_T$.
Hence
\begin{equation}
\begin{aligned}
\frac{1}{M'^\alpha}
=
C^{-\alpha\beta/(\alpha+\beta)}
\frac{1}{Q_M(x,k;r)^\alpha}.
\end{aligned}
\end{equation}

Substituting the above expressions into Eq.~(\ref{eq:m3_main})
yields
\begin{equation}
\begin{aligned}
L_{M^3}
&=
\mathcal{R}_\text{dual}(r,r_f)
\\
&\quad\times
\Bigl[
C^{-\alpha\beta/(\alpha+\beta)}
\Phi(x,k;r)
+
E
\Bigr],
\end{aligned}
\label{eq:kstar_factorized_loss}
\end{equation}
where
\begin{equation}
\begin{aligned}
\Phi(x,k;r)
=
\frac{A}{Q_M(x,k;r)^\alpha}
+
\frac{B}{x^\beta Q_D(k;r)^\beta}.
\end{aligned}
\end{equation}
Because $\mathcal{R}_\text{dual}(r,r_f)>0$,
$C^{-\alpha\beta/(\alpha+\beta)}$, and $E$ are all independent of $k$,
minimizing $L_{M^3}$ over $k$ is equivalent to minimizing
$\Phi(x,k;r)$ over $k$.
Therefore, the argmin over $k$ depends on $(C,D_T)$ only through
\begin{equation}
\begin{aligned}
x
=
\frac{D_T}{C^{\alpha/(\alpha+\beta)}}.
\end{aligned}
\end{equation}

It remains to connect this variable to $D^*(C)$.
Under monolingual single-epoch Chinchilla training,
\begin{equation}
\begin{aligned}
L_\text{base}\!\left(
\frac{C}{D},\, D
\right)
=
A\left(
\frac{D}{C}
\right)^\alpha
+
\frac{B}{D^\beta}
+
E.
\end{aligned}
\end{equation}
Its first-order condition is
\begin{equation}
\begin{aligned}
\alpha A C^{-\alpha} D^{\alpha-1}
=
\beta B D^{-\beta-1},
\end{aligned}
\end{equation}
which gives
\begin{equation}
\begin{aligned}
D^*(C)
&=
\left(
\frac{\beta B}{\alpha A}
\right)^{1/(\alpha+\beta)}
C^{\alpha/(\alpha+\beta)}
\\
&=
G^{-1} C^{\alpha/(\alpha+\beta)}.
\end{aligned}
\end{equation}
Thus,
\begin{equation}
\begin{aligned}
x
=
\frac{D_T}{C^{\alpha/(\alpha+\beta)}}
=
G^{-1} \cdot \frac{D_T}{D^*(C)}.
\end{aligned}
\end{equation}
Since the multiplicative constant $G^{-1}$ can be absorbed into the
definition of the function, the argmin over $k$ depends on $(C,D_T)$
only through $D_T/D^*(C)$.
When the minimizer is unique, this proves
Eq.~(\ref{eq:k_star_collapse}).
\hfill$\square$

The same derivation also applies to the monolingual variant
$M^3 + R_M^*(k)$, because replacing the constant $R_M^*$ in
Eq.~(\ref{eq:m3_Mprime}) by the $k$-dependent
$R_M^*(k)$ in Eq.~(\ref{eq:rm_star_mono}) still does not introduce any
direct dependence on $C$ or $D_T$.

\paragraph{Remark.}
The proposition above is exact only for fixed ratio schedule $\mathbf{r}$.
If $\mathbf{r}$ is jointly optimized together with $k$, the same collapse
does not follow from this argument alone.
Therefore, for \textbf{multi 1-stage} and \textbf{multi 2-stage}, Eq.~(\ref{eq:k_star_collapse})
should be interpreted as an empirical approximation rather than a
formal invariance result.

\paragraph{Corollary (small-$k$ regime).}
Assume that the saturation effects in both $D'$ and $M'$ are negligible.
Then, for fixed $\mathbf{r}$, the $M^3$ Scaling Law reduces to the Chinchilla form with total
token count $D = kD_T/r$ multiplied by a constant $\mathcal{R}_\text{dual}$ factor, and the optimal number of epochs is
approximately proportional to $D^*(C)/D_T$.

\paragraph{Proof.}
When $k-1 \ll R_D^*$, the saturation in the target-language effective
data is negligible, and
\begin{equation}
\begin{aligned}
h(k-1;R_D^*) \approx k.
\end{aligned}
\end{equation}
Similarly, when $k-1 \ll R_{D,\text{high}}^*$, the high-resource
weight in Eq.~(\ref{eq:g_func}) satisfies
\begin{equation}
\begin{aligned}
g\!\left(
r,\,
k-1;\,
R_{D,\text{high}}^*,\,
\psi
\right)
\approx 1.
\end{aligned}
\end{equation}
Therefore,
\begin{equation}
\begin{aligned}
D'
&=
D_T \cdot h(k-1;R_D^*)
+
g \cdot D_\text{high}
\\
&\approx
kD_T + D_\text{high}
\\
&=
kD_T + \frac{kD_T(1-r)}{r}
\\
&=
\frac{kD_T}{r}
=
D.
\end{aligned}
\end{equation}
If the model-side saturation is also negligible, then
\begin{equation}
\begin{aligned}
M' \approx M.
\end{aligned}
\end{equation}
Hence Eq.~(\ref{eq:m3_main}) reduces to
\begin{equation}
\begin{aligned}
L_{M^3}
\approx
\left(
\frac{A}{M^\alpha}
+
\frac{B}{D^\beta}
+
E
\right)
\mathcal{R}_\text{dual}(r,r_f).
\end{aligned}
\end{equation}
For fixed $\mathbf{r}$, the factor
$\mathcal{R}_\text{dual}(r,r_f)$ is constant with respect to $k$.
Therefore, minimizing $L_{M^3}$ over $k$ is asymptotically equivalent
to minimizing the Chinchilla base loss over the total number of tokens
$D$, whose optimum is $D^*(C)$.
Using $D = kD_T/r$, we obtain
\begin{equation}
\begin{aligned}
k_{\mathbf{r}}^*(C,D_T)
\approx
\frac{r\,D^*(C)}{D_T}.
\end{aligned}
\label{eq:k_star_smallk_general}
\end{equation}
In particular, for monolingual single-stage training ($r=1$),
\begin{equation}
\begin{aligned}
k^*(C,D_T)
\approx
\frac{D^*(C)}{D_T}.
\end{aligned}
\label{eq:k_star_smallk_mono}
\end{equation}
This proves the unit-slope asymptote in the log-log plot.
\hfill$\square$

\paragraph{Remark.}
Equation~(\ref{eq:k_star_smallk_mono}) is the monolingual special case
of Eq.~(\ref{eq:k_star_smallk_general}).
Thus, the statement
$k^*(C,D_T) \approx D^*(C)/D_T$ is exact only in the monolingual
small-$k$ regime.
For fixed multi-lingual $r<1$, the corresponding approximation is
$k_{\mathbf{r}}^*(C,D_T) \approx rD^*(C)/D_T$.
If $r$ is also jointly optimized, the small-$k$ approximation becomes
\begin{equation}
\begin{aligned}
k^*(C,D_T)
\approx
\frac{r^*(C,D_T)\,D^*(C)}{D_T}.
\end{aligned}
\end{equation}

\subsection{Additional Predictions of the $M^3$ Scaling Law}

\subsubsection{Optimal Model Scale}

\begin{figure*}[t]
  \centering
  \includegraphics[width=\textwidth]{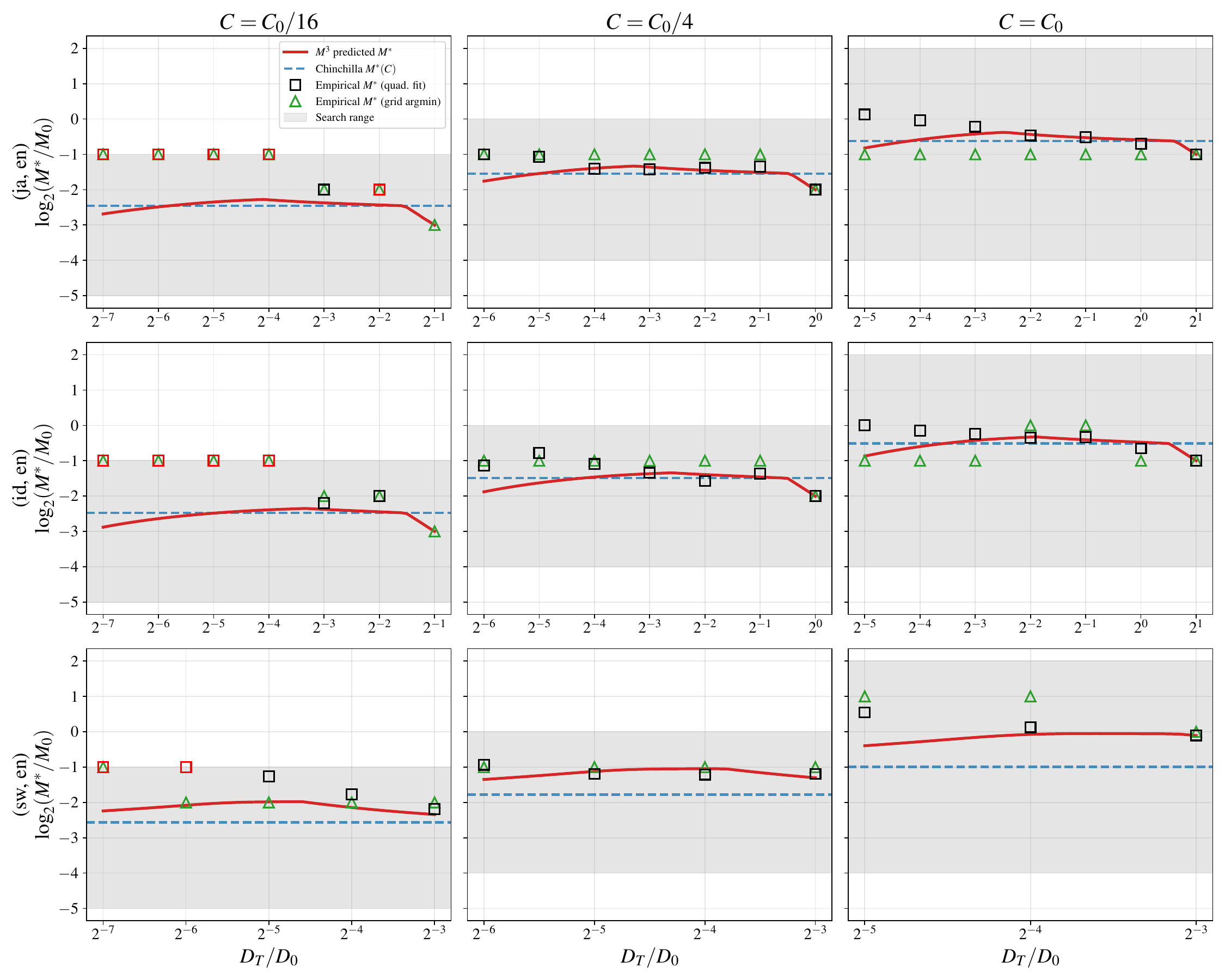}
  \caption{Optimal model scale $M^*$ predicted by the $M^3$ Scaling Law (red solid) compared with the Chinchilla compute-optimal scale $M^*(C)$ (blue dashed), empirical $M^*$ estimated by quadratic fitting in $f_M$ (squares) and discrete grid argmin (triangles), across three language pairs and three compute budgets. The gray band indicates the range of model scales explored in our experiments. Red-bordered squares indicate points where the quadratic fit optimum lies at or beyond the search boundary.}
  \label{fig:optimal_model_scale}
\end{figure*}

Figure~\ref{fig:optimal_model_scale} shows the optimal model scale $M^*$ predicted by the $M^3$ Scaling Law as a function of $D_T$ for each compute budget $C$.
We numerically minimized the predicted validation loss over all training approaches (mono 1-stage, multi 1-stage, and multi 2-stage) and their hyperparameters $(r, k, r_2)$ for each $(C, D_T)$, and report the model scale $M$ that achieved the minimum predicted loss.
For comparison, we show the Chinchilla compute-optimal model scale $M^*(C) = G \cdot C^{\beta/(\alpha+\beta)}$, as well as two empirical estimates of $M^*$ from the experimental data.
The first estimate (squares) fits a quadratic function in $f_M$ to the minimum validation loss at each model scale: for each $(C, D_T)$, we compute $\min_{r, k, r_2} L$ at each $f_M$ across all training approaches and setups sharing the same $(C, D_T)$, fit a quadratic function $L_{\min}(f_M) = p_0\,f_M^2 + p_1\,f_M + p_2$ to these per-$f_M$ minima, and take the vertex $f_M^* = -p_1/(2p_0)$ as the estimated optimal (clamped to the observed range of $f_M$).
The second estimate (triangles) simply takes the $f_M$ that achieves the lowest minimum validation loss on the discrete experimental grid.

Across all three language pairs and compute budgets, the $M^*$ predicted by the $M^3$ Scaling Law remains relatively stable as $D_T$ decreases, and the prediction error remains within the experimental grid resolution ($\Delta f_M = 1$) for most $(C, D_T)$ conditions.
In the regime where the optimal approach is monolingual single-stage (large $D_T$), the predicted $M^*$ increases gradually as $D_T$ decreases; after the optimal approach switches to multi-lingual two-stage training, $M^*$ tends to decrease slightly.
However, these predicted variations are smaller than the grid resolution of model scales in our experiments, so we cannot empirically confirm this trend.

Overall, these results suggest that, in practice, the compute-optimal model scale computed from the standard monolingual data-rich setting serves as a good starting point for low-resource language LLM pretraining, even when $D_T$ is substantially smaller than the compute-optimal data size $D^*(C)$.

\subsubsection{Optimal Language Ratio}

\begin{figure*}[t]
  \centering
  \includegraphics[width=\textwidth]{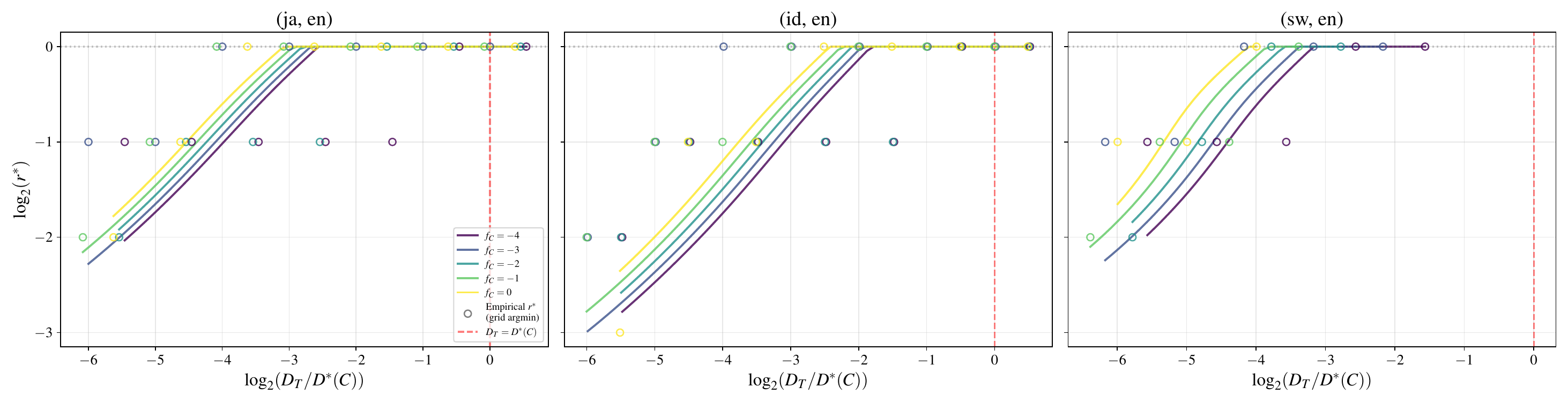}
  \caption{Optimal language ratio $r^*$ predicted by the $M^3$ Scaling Law (solid lines) and empirical $r^*$ from discrete grid argmin (circles) as a function of $\log_2(D_T/D^*(C))$ for each compute budget $C$ (colored by $f_C$). The red dashed line marks $D_T = D^*(C)$.}
  \label{fig:optimal_language_ratio}
\end{figure*}

Figure~\ref{fig:optimal_language_ratio} shows the optimal language ratio $r^*$ predicted by the $M^3$ Scaling Law for each $(C, D_T)$.
For each condition, we numerically minimized the predicted loss over all training approaches and their hyperparameters $(M, k, r_2)$, and report the $r$ that achieved the minimum.
The horizontal axis is normalized by the compute-optimal data size $D^*(C)$, which is estimated by minimizing the $M^3$ Scaling Law over $M$ under monolingual single-epoch training ($r=1$, $k=1$) at each $C$.
For comparison, we also plot the empirical $r^*$ obtained by taking the $r$ value that achieves the minimum validation loss on the discrete experimental grid for each $(C, D_T)$.

Across all three language pairs, the predicted $r^*$ remains at $r^* = 1$ (monolingual training is optimal) as $D_T$ decreases from $D^*(C)$, until a threshold is reached at which $r^*$ begins to decrease.
Below this threshold, the optimal approach is multi-lingual two-stage training with $r_f = 1$ (\S\ref{subsec:optimal_training_approach}), and $\log_2 r^*$ decreases approximately linearly in $\log_2(D_T / D^*(C))$.
The threshold at which this transition occurs approximately aligns across different $C$ when expressed in terms of $D_T / D^*(C)$, but does not collapse perfectly: a residual shift remains that is approximately equally spaced in $\log_2 C$.

The predictions of the $M^3$ Scaling Law broadly capture the trend of the empirical optimal $r^*$.
However, due to the coarse resolution of the language ratio grid ($r \in \{1, 1/2, 1/4, 1/8\}$) and the limited range of $C$ explored in our experiments, we cannot confirm from the experimental data alone whether the $C$-dependence of the residual shift of the threshold is a genuine trend or whether the linearity in log-log space holds exactly.

\section{Downstream Task Performance}
\label{section:downstream_task_performance}

\subsection{Justification of Using Validation Loss as the Primary Metric}

While the most important aspect of LLM applications is downstream task performance, our choice of validation loss on language modeling tasks as our primary evaluation metric is motivated by three key factors: (1) due to the emergent nature of LLM capabilities \cite{wei2022emergent}, analyzing scaling behaviors of downstream task performance with smaller-scale experiments (10M parameters, 200M tokens) is challenging and still under research \cite{hu2024predicting}, (2) validation loss has been shown to correlate well with downstream task performance in multiple studies \cite{brandfonbrener2024loss,chen2024scaling,du2024understanding,gadre2024language,thrush2024improving}, making it a reliable proxy for downstream capabilities despite some variance observed in models with similar validation losses \cite{xu2025unveiling}, and (3) it has been widely adopted in existing works that analyze scaling behaviors of LLMs \cite{kaplan2020scaling,hoffmann2022training,bi2024deepseek}.
More complete analysis including comprehensive downstream task analysis remains an important avenue for future work.
See \S\ref{section:correlation_analysis} for our preliminary empirical analysis.

\subsection{Preliminary Attempts for Evaluating Downstream Task Performance}
\label{section:correlation_analysis}

While analyzing downstream model performance for our smaller-scale models is challenging due to the emergent nature of LLM capabilities \cite{wei2022emergent}, we conducted a preliminary analysis to investigate relationship between validation loss and downstream task performance.

We evaluated Japanese LLMs trained in our experiments on the following Japanese downstream task benchmarks\footnote{JAQKET, JEMHopQA, NIILC, and JSQuAD are licensed under the CC BY-SA 4.0 license.}:
\begin{itemize}
  \item Question Answering: JAQKET \cite{suzuki2020jaqket}, JEMHopQA \cite{ishii-etal-2024-jemhopqa}, and NIILC \cite{sekine2003encyclopedia}
  \item Reading Comprehension: JSQuAD \cite{kurihara-etal-2022-jglue}
\end{itemize}
We also evaluated the models on other natural language inference, entity linking, and commonsense reasoning tasks in \texttt{llm-jp-eval}, but exact match scores of these tasks were almost consistently 0\% due to the small scale of our models, showcasing difficulties of downstream task evaluation and making these tasks unsuitable for our analysis.
We used the implementation of \texttt{llm-jp-eval} (v1.4.1, Apache-2.0 license) \cite{aizawa2024llm} and set the maximum sequence length to 4096, the number of few shot examples to 16, and the maximum number of samples to 100.
We report character F1 scores for these tasks.

\begin{table}[!htb]
  \centering
  \begin{tabular}{cc}
  \hline
  Task & $\rho$ \\
  \hline
  JAQKET & -0.378 \\
  JEMHopQA & -0.555 \\
  JSQuAD & -0.611 \\
  NIILC & -0.117 \\
  \hline
  Average & -0.696 \\
  \hline
  \end{tabular}
  \caption{Pearson correlation coefficients ($\rho$) between validation loss and downstream task performance (character F1 scores). All tasks show negative correlation, indicating that lower validation loss tends to correspond to better task performance. Note that ``Average'' represents the correlation with the mean F1 score across all tasks, not the mean of individual correlation coefficients.}
  \label{tab:correlation}
  \end{table}

As shown in Table \ref{tab:correlation}, we observed moderate negative correlations between validation loss and task performance for most tasks, with a correlation coefficient of -0.696 for the average score across all tasks.
While NIILC showed relatively weak correlation (-0.117), other tasks demonstrated moderate negative correlations, ranging from -0.378 to -0.611.
This result suggests that improvements in validation loss generally correspond to better downstream task performance.

\begin{figure*}[t]
  \centering
  \includegraphics[width=\textwidth]{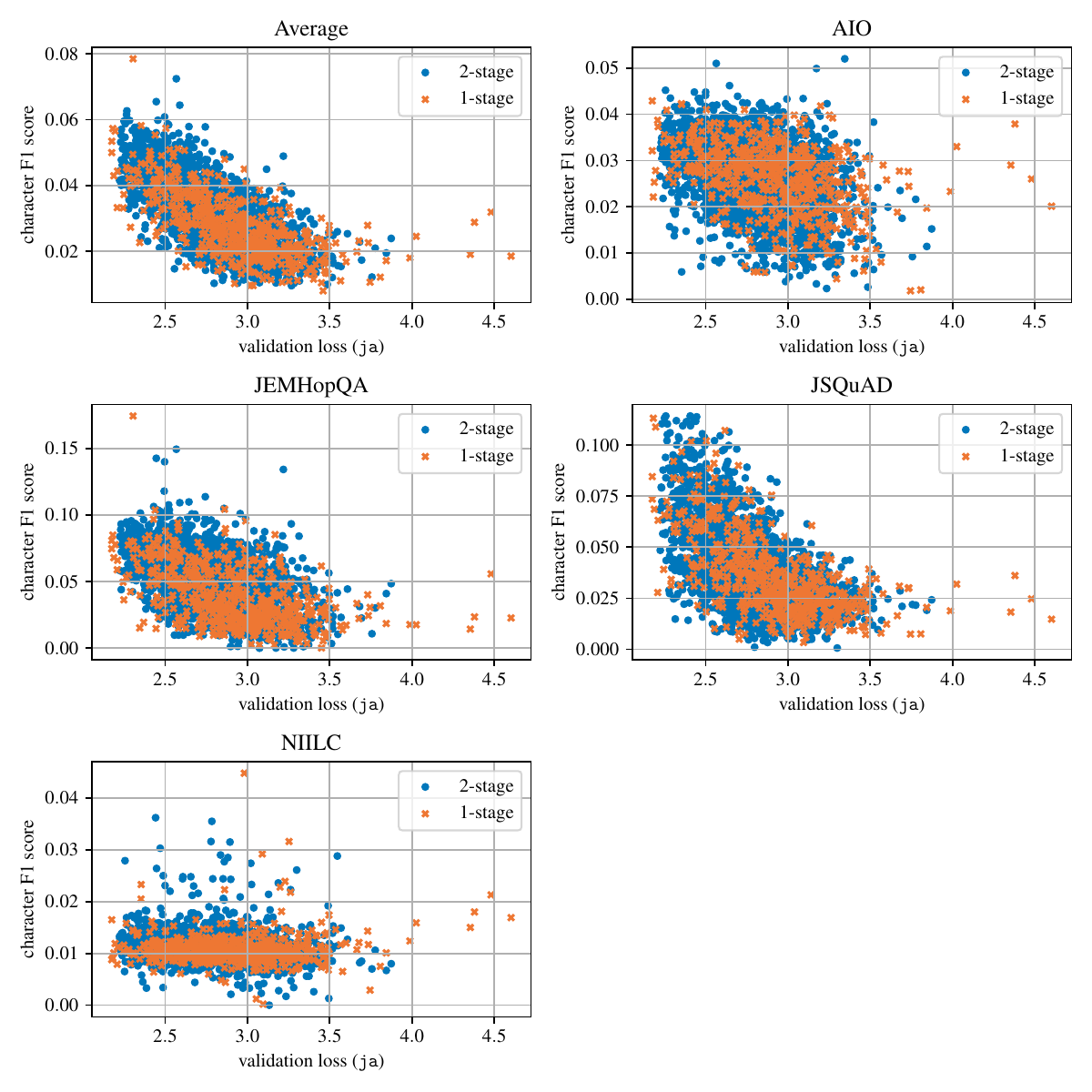}
  \caption{Relationship between validation loss and downstream task performance (character F1 scores). Different colors represent single-stage (orange) and two-stage (blue) training approaches, showing similar correlation patterns regardless of the training strategy.}
  \label{fig:supplementary_jatask}
\end{figure*}

Furthermore, as shown in Figure \ref{fig:supplementary_jatask}, both single-stage and two-stage training approaches exhibit similar patterns in the relationship between validation loss and downstream task performance.
This similarity supports our comparative analysis of different training approaches based on validation loss.

It is important to note that this correlation analysis was conducted using relatively small-scale models with limited absolute performance on these tasks.
While our results suggest the validity of using validation loss as a proxy metric for model comparison, future work with larger-scale models would be necessary for a more precise analysis of this relationship.

\section{Full Experimental Results}
\label{section:full_results}

\begin{figure*}[t]
  \includegraphics[width=\textwidth]{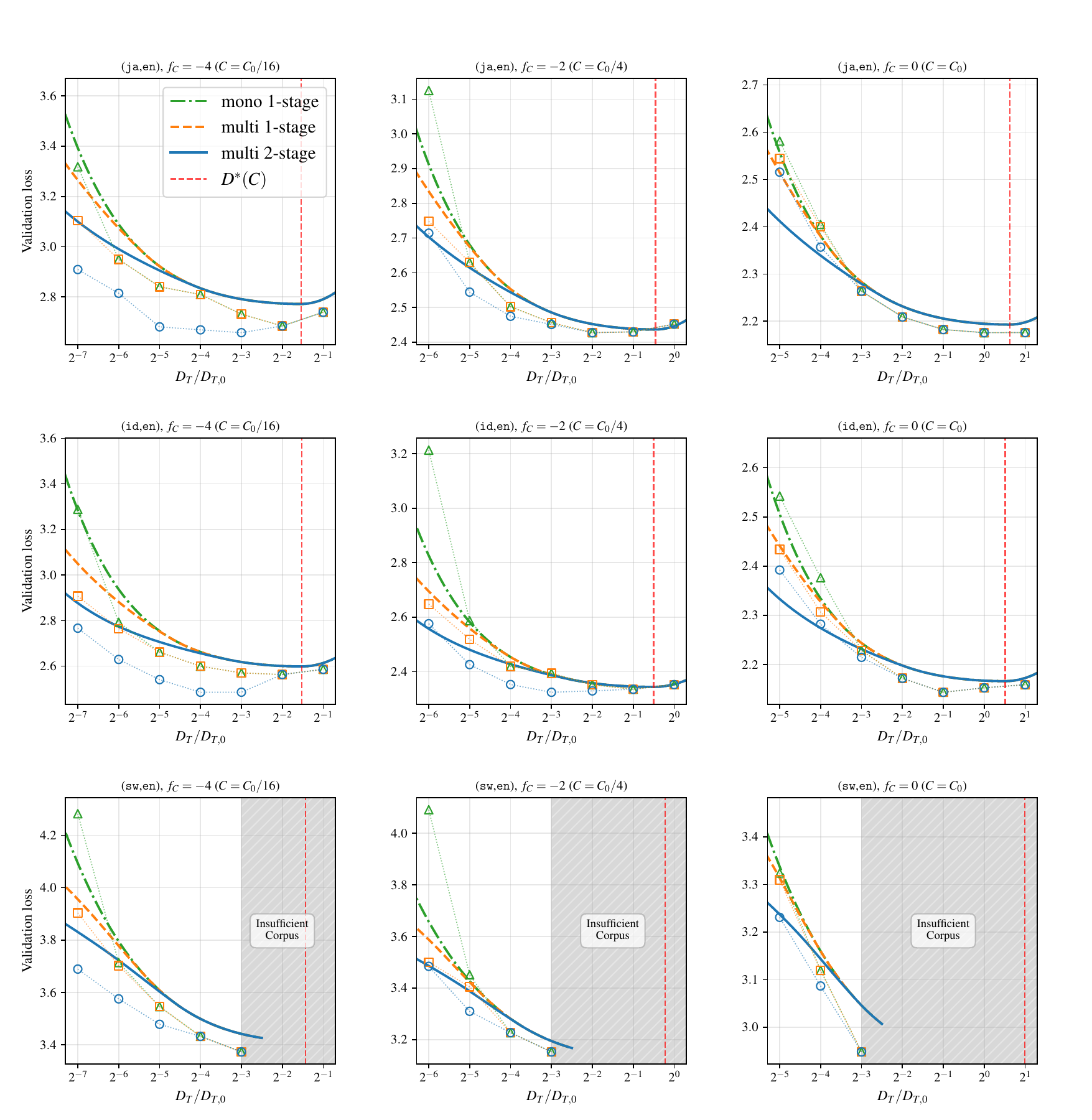}
  \caption{Minimum validation loss for each training approach across various target language corpus sizes $D_T$ and compute budgets $C$, for all language pairs. Smooth curves show $M^3$ Scaling Law predictions; dotted lines with hollow markers show corresponding observed values. The red dashed vertical lines represent the compute-optimal data size $D^*(C)$.}
  \label{fig:supplementary_all_total}
\end{figure*}

Figure \ref{fig:supplementary_all_total} shows the minimum validation loss on the target language achieved by different LLM training approaches across various budget constraints $(C,D_T)$.

\section{License Notice}

This paper contains information from C4 and mC4 corpus\footnote{Available at \url{https://huggingface.co/datasets/allenai/c4}.} which is made available under the ODC Attribution License.

\section{Use of AI Assistants}

We used AI assistants (including large language models) during this research in the following ways: (i) grammar and style polishing and translation assistance for the manuscript text; (ii) drafting prose from author-provided outlines or bullet points, which the authors then reviewed and revised; (iii) assisting the implementation of experimental and visualization code; and (iv) suggesting draft outlines for mathematical proofs of propositions formulated by the authors. All AI-generated content---including manuscript text, code, and proof drafts---was reviewed and verified by the authors, who take full responsibility for the final content of this paper.

\end{document}